\PassOptionsToPackage{unicode}{hyperref}
\PassOptionsToPackage{hyphens}{url}
\documentclass[11pt]{article}

\usepackage[utf8]{inputenc}
\usepackage[T1]{fontenc}
\usepackage{lmodern}

\usepackage[left=1in,right=1in,top=0.75in,bottom=1in]{geometry}
\usepackage{setspace}
\usepackage{etoolbox}
\usepackage{needspace}
\usepackage{amsmath,amssymb,amsfonts}
\usepackage{mathtools}

\usepackage{booktabs}
\usepackage{graphicx}
\usepackage[font=small,labelfont=bf,margin=1em]{caption}
\usepackage{float}
\usepackage[section]{placeins}
\usepackage{longtable}
\usepackage{array}
\usepackage{ragged2e}

\usepackage{xcolor}
\IfFileExists{orcidlink.sty}{\usepackage{orcidlink}}{}
\newcommand{\orcidmark}[1]{%
  \IfFileExists{orcidlink.sty}{\orcidlink{#1}}{\href{https://orcid.org/#1}{ORCID: #1}}%
}

\usepackage{iftex}
\ifPDFTeX
  \usepackage[T1]{fontenc}
  \usepackage[utf8]{inputenc}
  \usepackage{textcomp} % provide euro and other symbols
\else % if luatex or xetex
  \usepackage{unicode-math} % this also loads fontspec
  \defaultfontfeatures{Scale=MatchLowercase}
  \defaultfontfeatures[\rmfamily]{Ligatures=TeX,Scale=1}
\fi
\usepackage{lmodern}
\ifPDFTeX\else
\fi
\IfFileExists{upquote.sty}{\usepackage{upquote}}{}
\IfFileExists{microtype.sty}{% use microtype if available
  \usepackage[]{microtype}
  \UseMicrotypeSet[protrusion]{basicmath} % disable protrusion for tt fonts
}{}
\makeatletter
\@ifundefined{KOMAClassName}{% if non-KOMA class
  \IfFileExists{parskip.sty}{%
    \usepackage{parskip}
  }{% else
    \setlength{\parindent}{0pt}
    \setlength{\parskip}{6pt plus 2pt minus 1pt}}
}{% if KOMA class
  \KOMAoptions{parskip=half}}
\makeatother
\usepackage{color}
\usepackage{fancyvrb}

\DefineVerbatimEnvironment{Highlighting}{Verbatim}{commandchars=\\\{\}}
\newenvironment{Shaded}{}{}

\newcommand{\CommentTok}[1]{\textcolor[rgb]{0.38,0.63,0.69}{\textit{#1}}}

\newcommand{\FunctionTok}[1]{\textcolor[rgb]{0.02,0.16,0.49}{#1}}

\newcommand{\InformationTok}[1]{\textcolor[rgb]{0.38,0.63,0.69}{\textbf{\textit{#1}}}}

\newcommand{\NormalTok}[1]{#1}

\newcommand{\OtherTok}[1]{\textcolor[rgb]{0.00,0.44,0.13}{#1}}

\newcommand{\SpecialStringTok}[1]{\textcolor[rgb]{0.73,0.40,0.53}{#1}}

\usepackage{longtable,booktabs,array}
\usepackage{calc} % for calculating minipage widths
\usepackage{etoolbox}
\makeatletter
\patchcmd\longtable{\par}{\if@noskipsec\mbox{}\fi\par}{}{}
\makeatother
\IfFileExists{footnotehyper.sty}{\usepackage{footnotehyper}}{\usepackage{footnote}}
\makesavenoteenv{longtable}
\usepackage{graphicx}
\makeatletter
\newsavebox\pandoc@box
\newcommand*\pandocbounded[1]{% scales image to fit in text height/width
  \sbox\pandoc@box{#1}%
  \Gscale@div\@tempa{\textheight}{\dimexpr\ht\pandoc@box+\dp\pandoc@box\relax}%
  \Gscale@div\@tempb{\linewidth}{\wd\pandoc@box}%
  \ifdim\@tempb\p@<\@tempa\p@\let\@tempa\@tempb\fi% select the smaller of both
  \ifdim\@tempa\p@<\p@\scalebox{\@tempa}{\usebox\pandoc@box}%
  \else\usebox{\pandoc@box}%
  \fi%
}
\def\fps@figure{htbp}
\makeatother
\providecommand{\tightlist}{%
  \setlength{\itemsep}{0pt}\setlength{\parskip}{0pt}}
\usepackage[numbers]{natbib}
\usepackage{fvextra}
\fvset{
  breaklines=true,
  breakanywhere=true,
  fontsize=\footnotesize
}

\newcolumntype{L}[1]{>{\RaggedRight\arraybackslash}p{#1}}
\AtBeginEnvironment{longtable}{\footnotesize\setlength{\tabcolsep}{3pt}}

\makeatletter
\def\fps@figure{!htbp}
\makeatother

\usepackage{bookmark}
\IfFileExists{xurl.sty}{\usepackage{xurl}}{} % add URL line breaks if available
\hypersetup{
  pdftitle={Measuring the sensitivity of LLM-based structured extraction to prompt{,} model{,} and schema choices in clinical discharge summaries},
  hidelinks,
  pdfcreator={LaTeX via pandoc}}
\hypersetup{hidelinks}

\title{Measuring the sensitivity of LLM-based structured extraction to
prompt, model, and schema choices in clinical discharge summaries}

\makeatletter
\patchcmd{\@maketitle}{\null\vskip 2em}{\null\vskip -0.25em}{}{}
\patchcmd{\@maketitle}{\par\vskip 1.5em}{\par\vskip 0.5em}{}{}
\makeatother

\author{Martin Murin \orcidmark{0000-0002-2388-1969}\\DryLabz GmbH,
Basel, Switzerland\\martin.murin@proton.me}
\date{June 2026}

\begin{document}
\pagestyle{plain}
\maketitle
\begin{abstract}
Large language models are increasingly used for structured extraction
from clinical free-text notes, but the sensitivity of their output to
upstream configuration choices is less understood than their accuracy on
fixed benchmarks. This work measures that sensitivity without
human-annotated ground truth, by holding the extraction task fixed and
varying one choice at a time. The fixed schema comprises 17 clinical
documentation flags on a three-way yes/no/not\_documented value set and
a 47-tag vocabulary for the primary admission reason. Three prompt
variants expressing this schema were each run at two model sizes on
MIMIC-IV v3.1 discharge summaries. Cross-prompt agreement was measured
by Cohen's kappa on ICD-stratified subsets. A paired same-note
comparison isolated the effect of model choice, and a post-hoc collapse
of the three-way flags to binary tested the schema's contribution to
disagreement. On the three-way flags, the two models reach the same
pooled cross-prompt agreement (median kappa 0.69 and 0.68); the larger
model raises agreement on some fields and lowers it on others, a
redistribution rather than the absence of an effect. Collapsing the
schema to binary dissolves most of the cross-prompt disagreement,
locating it on the absence-versus-silence distinction rather than on
whether the finding is present. On the multi-class admission
categorization, changing the model reassigns the dominant tag on close
to half of all notes while changing the prompt phrasing reassigns it on
roughly one in eight, and the larger model places far less mass on
residual catch-all categories (44\% to 26\%). These patterns indicate a
schema-imposed source of disagreement concentrated on the
absence-versus-silence axis and a dominance of model over prompt
phrasing on multi-class categorization, identified by a reusable
methodology for auditing extraction reproducibility on a
population-scale deployment.
\end{abstract}

\section{Introduction}\label{introduction}

Large language models (LLMs) have become a practical tool for structured
information extraction from clinical free-text notes. Several recent
studies illustrate the range of what is now feasible. Zero-shot and
few-shot prompting extracts pathology classifications
\cite{huang2024chatgpt}; iterative refinement pipelines reach macro-F1
above 0.97 on kidney-tumor pathology reports \cite{hein2025iterative};
retrieval-augmented methods extract multiple variables across 20,000
notes \cite{lopez2025clear}; instruction-tuned LLaMA models perform
clinical named-entity recognition and relation extraction
\cite{hu2024improving,hu2026information}; and recent systems extract
structured oncology data \cite{zhang2025mcodegpt} and structured
radiology-report fields using both proprietary and open-weight models
\cite{leguellec2024radiology}. Across these settings, LLM-based
extraction reaches high accuracy on narrow schemas where expert-curated
structured data is available for validation. This literature optimizes
extraction quality given a fixed prompt and a fixed target schema,
against a reference standard derived from expert curation, billing-code
linkages, or task-specific human annotation.

Building an LLM extraction pipeline requires a set of configuration
choices that precede any measurement of quality: how the extraction
prompt is worded, how fine-grained the target schema is, and which model
tier is used to run it. Each of these choices is typically fixed once
and thereafter treated as settled. Yet they shape the extraction output
in ways that remain invisible unless the alternatives are also run for
comparison. General-domain work has begun to formalize parts of this
dependence: prompt phrasing has been established as a measurable driver
of LLM behavior \cite{errica2025quantifying,razavi2025benchmarking},
prompt-strategy choice has been shown to affect performance across
clinical NLP tasks \cite{sivarajkumar2024prompting}, and annotation
frameworks have been proposed in which disagreement between sources is
preserved rather than resolved to a single agreed label
\cite{xu2026beyond}. Within clinical NLP, assertion status, the
distinction between an explicitly negated condition and one about which
the note is simply silent, has long been recognized as a consequential
schema decision rather than a neutral one
\cite{uzuner2011i2b2,harkema2009context}. What has not been provided is
a systematic characterization of how far these configuration choices
move the extraction output for structured extraction from long,
heterogeneous discharge summaries.

This paper provides that characterization, together with a methodology
for producing it. The influence of three configuration choices is
studied by holding the extraction task fixed and varying each choice in
a controlled manner: prompt wording, across three variants that express
one identical target schema; model size, across two operating points of
a single vendor's product range; and schema granularity, through
post-hoc binary collapse of a three-way categorical schema. Applied to a
population-scale extraction from MIMIC-IV v3.1 discharge summaries
\cite{johnson2023mimiciv,johnson2024mimiciv,johnson2023mimicnote,goldberger2000physionet},
the procedure yields a field-level account of where extraction is stable
and where it is not. Three distinct patterns emerge: a schema-imposition
effect on the three-way categorical fields, a dominance of model size
over prompt phrasing on multi-class categorization, and a residual
phrasing sensitivity on the secondary tags. The methodology is reusable:
it assembles a set of agreement and disagreement diagnostics that
require no human-annotated ground truth and apply to any extraction task
with multiple prompt variants targeting one schema.

\section{Methods}\label{sec-methods}

\subsection{Cohort and splits}\label{sec-cohort-splits}

The study draws on MIMIC-IV v3.1 discharge summary notes
\cite{johnson2023mimiciv,johnson2024mimiciv,johnson2023mimicnote,goldberger2000physionet}
(331,793 unique admissions). Each admission has exactly one discharge
summary in the source data.

To enable a controlled methodology study while preserving generalization
tests, disjoint splits of admissions were constructed. Notes were
stratified by ICD-10 chapter so that clinical content was distributed
comparably across splits. Random sampling within strata used a fixed
seed. Each split is defined by a manifest of admission identifiers
committed to the project repository, and each manifest carries a SHA-256
hash for integrity verification.

The following splits were used:

\begin{itemize}
\tightlist
\item
  a 200-note smoke set, used for an initial round of prompt revision
  before three-variant extraction was launched;
\item
  a 150-note refinement set, used for the initial three-variant
  extraction and the optimization loop;
\item
  a 150-note holdout set, used for single-touch generalization
  evaluation after the optimization loop completed (Section
  \ref{sec-optimization-loop});
\item
  a 1,000-note validation set, used for cross-prompt agreement
  evaluation at intermediate scale and for selecting the production
  variant (Section \ref{sec-variant-selection-production});
\item
  a 5,000-note extended validation set, used to confirm that
  cross-prompt agreement is stable as the sample grows beyond the
  1,000-note scale;
\item
  a 500-note audit subset, drawn from the extended validation set, used
  for detailed disagreement audit;
\item
  a 1,500-note paired sample, formed by the union of the 1,000-note
  validation set and the 500-note audit subset, used for the same-note
  model-size comparison (Section \ref{sec-model-size-comparison});
\item
  the population extraction over the remaining admissions, used for
  downstream applications outside the scope of this work (Section
  \ref{sec-variant-selection-production}).
\end{itemize}

The smoke, refinement, and holdout sets were constructed together in a
single stratified split build. Split-construction scripts for the larger
samples exclude all earlier admission identifiers, and overlap between
splits is checked at construction time. Split sizes and purposes are
summarized in Table \ref{tbl:splits}.

The holdout set was treated as single-touch: it was used exactly once,
to evaluate the cross-prompt agreement of the prompt set produced after
the optimization loop (Section \ref{sec-optimization-loop}). No prompt
edits, schema changes, or hyperparameter choices were made after
consulting holdout results.

\subsection{Schema design}\label{sec-schema-design}

The extraction schema defines four feature categories: an
admission-reason vocabulary, a set of TriState clinical flags, a set of
enumerated single-choice fields, and a set of count and free-text
fields.

The first category is the admission-reason vocabulary, a controlled set
of 47 tags organized by organ system and clinical pattern. Each note
requires two outputs: a non-empty subset of the vocabulary (the
admission-reason tag set) and a single dominant tag, which must be a
member of the returned set. The vocabulary was designed before
three-variant prompt development and was not modified during the
methodology study. The full vocabulary with tag definitions is provided
in Table \ref{tbl:admission-tags}.

The second category is the TriState clinical flags: 17 fields
representing the documentation status of specific clinical events,
social determinants, and discharge planning items (Table
\ref{tbl:tristate-fields}). Each field takes one of three values. The
value is \texttt{yes} when the feature is affirmatively documented in
the note, \texttt{no} when an explicit negation is present, and
\texttt{not\_documented} when the note is silent on the feature. This
three-way distinction was adopted so that downstream consumers can
separate active absence, in which the feature was considered and
rejected, from passive absence, in which the note carries no information
about the feature. The clinical motivation is the documentation-quality
use case, where the presence or absence of an explicit negation can
carry meaning beyond the presence or absence of a positive statement.

The third category is the enum fields: discrete-valued single-choice
fields with closed value sets, covering mental status, functional status
at discharge, discharge condition category, and others (Table
\ref{tbl:enum-fields}).

The fourth category comprises integer count fields, such as the number
of specialty consults, and a free-text evidence field populated by the
model with quoted or paraphrased note content supporting the structured
output. Count fields are part of the production extraction but were not
analyzed in this work. The evidence field is retained for audit and
qualitative review but is not consumed by any downstream metric here.

Every extracted record was validated against a Pydantic schema. The
validation enforces that types match the schema, that all enum and tag
values lie in their declared value sets, that the admission-reason tag
set is non-empty, and that the dominant admission reason is a member of
the returned tag set. Records failing validation were marked as parse
failures and excluded from downstream aggregation.

\subsection{The three controlled comparisons}\label{sec-comparisons}

The methodology holds the extraction task and the source notes fixed and
varies three configuration choices in turn: the prompt phrasing, the
model size, and the granularity of the categorical schema. Each
comparison is described below. The agreement metrics applied across all
three are defined in Section \ref{sec-cross-prompt-metrics}.

\subsubsection{Prompt phrasing: three
variants}\label{sec-three-prompt-variants}

Three prompt variants were developed to elicit the same structured
output under different phrasings. Variant A is a detailed prose prompt,
in which the task is described in continuous text with the schema
appended as JSON-style examples. Variant B presents the task in an
evidence-first structure, instructing the model to identify and quote
relevant note passages for each schema field and then to produce the
structured output keyed to those quotes. Variant C uses a questionnaire
format, in which each schema field is presented as a separate narrow
question with explicit options and the model answers field by field.

The variants were drafted and evaluated on the 200-note smoke set. A
single hand-edit was applied across all three during smoke testing to
address a vocabulary coverage gap, and no further variant-level changes
were made before three-variant extraction was launched on the refinement
set. The full prompt text for each variant is provided in supplementary
material.

\subsubsection{Model size: paired
re-extraction}\label{sec-model-size-comparison}

To compare the effect of model size on cross-prompt agreement against
sample-draw variation, a paired re-extraction was performed using the
full model. The same 1,500 admissions covered by the 1,000-note
validation set and the 500-note audit subset were re-extracted with all
three prompt variants at the full model. Both model sizes are snapshots
from a single vendor's product range: the small model is
\texttt{gpt-5.4-nano-2026-03-17} and the full model is
\texttt{gpt-5.4-2026-03-05}.

This design enables a same-note paired comparison: for every note in the
paired sample, the small-model and full-model outputs are available
across all three variants. Because the comparison is made on the same
notes, any same-note difference reflects the change in model operating
point rather than a difference in sample composition. The two model
sizes are treated as two operating points to be compared, with neither
taken as a reference standard for the other.

\subsubsection{Schema granularity: binary
collapse}\label{sec-binary-collapse}

A post-hoc re-analysis evaluated the dependence of cross-prompt
agreement on the granularity of the TriState schema. Under this
re-analysis, each TriState value was re-mapped: \texttt{yes} was
preserved as \texttt{yes}, and \texttt{no} and \texttt{not\_documented}
were merged into a single class labeled \texttt{not\_yes}. The collapse
is a re-labeling of the existing extractions and not a re-extraction;
the prompts are unchanged and the model is not asked to categorize under
a binary schema.

All cross-prompt agreement metrics (kappa, percent agreement,
disagreement decomposition, per-pair confusion analysis) were recomputed
on the re-mapped values. The binary-collapse analysis is reported
alongside the full-TriState analysis throughout Section
\ref{sec-results}. Whether cross-prompt agreement changes substantially
under binary collapse, and how that change is distributed across fields,
is a primary empirical finding of this work and is addressed in Section
\ref{sec-results-collapse-structure}.

\subsection{Cross-prompt agreement
metrics}\label{sec-cross-prompt-metrics}

Pairwise Cohen's kappa across prompt variants is a categorical-agreement
metric in the inter-rater-reliability tradition
\cite{mchugh2012kappa,uzuner2011i2b2}. Chance correction is the relevant
property here: most TriState fields have skewed marginals, with
\texttt{not\_documented} dominating, so raw percent agreement would
register as high on these fields for reasons of base rate alone, whereas
kappa discounts the agreement expected from the marginals and isolates
agreement on the categorization itself. It serves the same diagnostic
role as the entropy-based prompt-sensitivity metrics formalized for
general-domain classification
\cite{errica2025quantifying,razavi2025benchmarking}, with the practical
difference that it is interpretable at field-level resolution in a
multi-field structured-extraction setting. The choice of pairwise kappa
rather than per-input prediction entropy follows from the design of the
present study. Entropy-based prompt-sensitivity metrics estimate a
distribution over outputs and therefore need a large ensemble of
rephrasings to be meaningful; this study uses three deliberately
distinct prompt phrasings, too few to estimate an output distribution
but well suited to pairwise agreement, which is defined for as few as
two raters.

Cross-prompt agreement was measured as the pairwise Cohen's kappa
between variant outputs on the same note, computed independently per
schema field and per variant pair. For TriState fields, kappa was
computed treating the three values (\texttt{yes}, \texttt{no},
\texttt{not\_documented}) as categorical labels with no implied
ordering. The three values are not an ordered scale:
\texttt{not\_documented} is an evidence-availability state rather than a
midpoint between \texttt{yes} and \texttt{no}, so an ordinal treatment
would impose a ranking the schema does not carry. For enum fields, kappa
was computed over the declared value set. For the dominant admission
reason, kappa was computed over the 47-tag vocabulary.

Two multi-class admission-reason quantities are reported. Cross-variant
agreement on the single-choice primary admission reason is the
exact-match rate on the dominant tag, equivalent to the diagonal mass of
a \(47\times47\) confusion matrix with one label per note. Agreement on
the multi-label admission-reason tag set, where a single-label confusion
matrix does not apply, is summarized by admission-tag set agreement, the
mean per-note Jaccard index. For note \(i\) with variant tag sets
\(A_i\) and \(B_i\), per-note agreement is
\(|A_i \cap B_i| / |A_i \cup B_i|\), defined as 1 when both sets are
empty, and the reported set agreement is the mean of this quantity
across notes. A value of one indicates identical tag sets on every note;
lower values indicate divergence in which tags were assigned. Because
each prompt variant is instructed to include the primary admission
reason in the tag set, set agreement is bounded below by dominant-tag
exact-match agreement and therefore primarily reflects consistency on
the secondary, non-dominant tags. This set agreement is a descriptive
measure and is not chance-corrected; it is reported as a percentage and
is not placed on the same scale as the chance-corrected Cohen's kappa
used for the TriState and enum fields.

For each TriState field, the per-field kappa values are reported
alongside a filtered-median summary across those fields. The filter
excludes fields with low positive-class base rate, defined as fewer than
10 total positive votes across all variants on the sample at hand. In
the model-size comparison the filter is applied jointly to both model
sizes, so a field is included only if it passes the threshold at the
small model and at the full model. This filter is applied because
Cohen's kappa is sensitive to the marginal distribution of the
categorical labels. When the positive class is very rare, the point
estimate is dominated by base-rate variance and is no longer a reliable
measure of agreement structure.

Several kappa-related quantities are referenced in subsequent sections,
and the following notation is used throughout. The quantity
\(\kappa_{p,f}\) denotes Cohen's kappa for variant pair
\(p \in {\text{A-B}, \text{A-C}, \text{B-C}}\) on field \(f\). The
atomic unit of measurement is therefore a single \(\kappa\) value for
one (variant-pair, field) cell. With 17 TriState fields and three
variant pairs there are up to 51 such cells per model size before
base-rate filtering, and 48 cells after filtering on the paired
model-size sample (16 fields \(\times\) 3 pairs).

The model-size comparison is summarized by a paired per-field statistic.
For a single field, the median over the three variant pairs of
\(\kappa_{p,f}(\text{full}) - \kappa_{p,f}(\text{small})\) gives that
field's model-size difference, \(\Delta\kappa_f\). Taking the median of
\(\Delta\kappa_f\) over the filtered field set gives
\(\overline{\Delta\kappa}^{\,\mathrm{per\text{-}field}}\), the
model-size difference experienced by the typical field. Because each
\(\Delta\kappa_f\) is formed on the same notes before aggregation, this
median-of-differences is the paired quantity used for every
between-model comparison in this work. Separately, the pooled median
\(\bar{\kappa}\) is the median taken over all (variant-pair, field)
cells at one model size at once, treating those cells as a single flat
population; it is reported per model size as a benchmark-anchor level
(small and full, under the TriState schema and under collapse), so that
the agreement of each configuration can be situated against external
work. The unpaired difference of the two pooled medians is not reported
as the model-size statistic, because the field at the median of the
small-model distribution need not be the field at the median of the
full-model distribution, so their difference does not correspond to any
field's experience.

\subsection{Labeling function ensemble}\label{sec-lf-ensemble}

Independent labeling functions were developed for a subset of schema
targets to provide signal beyond the LLM extraction. Three
labeling-function families were used. The complete labeling-function
definitions are provided in Table\nobreakspace{}\ref{tbl:icd-lfs} and
Table\nobreakspace{}\ref{tbl:regex-lfs}.

ICD labeling functions (15 in total) map admission ICD-10 codes to
specific schema targets. Each ICD labeling function is associated with
one target and emits a positive vote when any code in its target's code
list appears in the admission's structured diagnosis codes, and an
abstain vote otherwise. ICD labeling functions cover targets where ICD
coverage is expected, such as cardiac heart failure, sepsis, AKI, and
hepatic failure. The full set of ICD anchors with their code lists is
given in Table \ref{tbl:icd-lfs}.

Regex labeling functions (8 in total) match curated regular expressions
against the note text. Each regex labeling function emits a positive
vote on match and abstains otherwise. The pattern set was bootstrapped
from the refinement set and is listed in Table \ref{tbl:regex-lfs}.

LLM labeling functions (96 in total) are derived from the three-variant
extraction. They cover 32 target field-value pairs: 14 admission-tag
membership targets and 9 TriState fields each split into a \texttt{yes}
and a \texttt{no} target (\(14 + 9 \times 2 = 32\)). For each target
field-value pair the three variant outputs each contribute one labeling
function, yielding \(3 \times 32 = 96\) LLM-based votes per note.

Labeling-function votes were integrated using a Snorkel LabelModel fit
per target \cite{ratner2017snorkel}. The LabelModel infers
per-labeling-function accuracies from cross-LF agreement patterns and
produces a probabilistic consensus label for each note. The label-model
output was used as the aggregated reference signal for two purposes in
this work: selecting the production prompt variant (Section
\ref{sec-variant-selection-production}), and clustering disagreement
patterns for the optimization loop (Section
\ref{sec-optimization-loop}).

Embedding-based labeling functions were also explored but produced
overlapping signal and noise distributions on the refinement set and
were not included in the final ensemble.

\subsection{Autonomous optimization loop}\label{sec-optimization-loop}

A concern during prompt development was that the three variants might
produce systematically different outputs on subsets of targets, in the
sense that one variant might consistently disagree with the consensus of
the others in a way attributable to its specific phrasing rather than to
the content of the note. To detect and address such systematic
divergence, an automated optimization procedure was developed. The
procedure is described in its general form below; in practice it was
applied only to variant C, which on the initial three-variant extraction
showed clustered disagreement against the consensus of variants A and B
on several targets.

The procedure is:

\begin{enumerate}
\def\labelenumi{\arabic{enumi}.}
\item
  Compute three-variant labeling-function votes on the refinement set
  and identify clusters of systematic disagreement. A cluster is a
  target-and-pattern combination where one variant systematically
  diverges from the other two, for example a case in which variant C
  asserts \texttt{no} on cognitive\_impairment where A and B assert
  \texttt{not\_documented}.
\item
  For each cluster, construct a prompt for a full-model rewrite step.
  The full model is given the current prompt of the diverging variant, a
  sample of disagreement examples, and instructions to revise the prompt
  to address the cluster while preserving schema structure.
\item
  The revised prompt is accepted only if four deterministic guards pass:
  the revised prompt produces output of the same schema cardinality,
  with all required fields and no unexpected fields; all enum and tag
  values remain within the declared value sets; no field is renamed; and
  no field is removed. Failed rewrites are discarded.
\item
  The accepted prompt is re-extracted on the refinement set, and the new
  cluster disagreement rate is measured. The loop continues to the next
  cluster.
\item
  Iteration stops when any one of the following holds: a maximum of 5
  iterations is reached; the per-iteration improvement in cluster
  disagreement rate falls below 2 percentage points; no remaining
  cluster exceeds a residual disagreement volume of 50 cases; or two
  consecutive iterations produce no improvement.
\end{enumerate}

The loop ran for 2 iterations on variant C, reducing the targeted
cluster disagreement count from 547 to 75 (a 86.29\% reduction). The
post-optimization prompts for all three variants were then locked and
evaluated on the holdout set (Section \ref{sec-cohort-splits}) as a
single-touch generalization test.

\subsection{Variant selection and production
extraction}\label{sec-variant-selection-production}

After holdout validation, a single prompt variant was selected for the
population-scale extraction. The selection used the 1,000-note
validation set. For each variant, the LLM extraction output was compared
to the Snorkel-aggregated reference signal across the labeling-function
targets. The agreement metric was the mean weighted agreement between
the variant's per-target output and the per-target Snorkel posterior,
weighted by per-target support. For this metric, TriState
\texttt{not\_documented} was treated as an abstain vote on the
corresponding labeling-function target.

The metric is one of several reasonable choices for variant selection.
It measures agreement with the consensus label produced by the full
labeling-function ensemble, of which the LLM variant is one input, which
introduces a degree of self-consistency in the score. The selection
nevertheless served the practical purpose of choosing a variant.
Cross-prompt agreement metrics are pairwise across all three variants
and therefore do not depend on which one is designated as production.
Variant A scored highest in the metric and was selected.

All extractions were produced with deterministic decoding: temperature
was set to zero and no reasoning mode was enabled, for both model sizes
and all three prompt variants. Cross-prompt and cross-model disagreement
reported in this work is therefore not attributable to sampling
stochasticity in the decoding step; it reflects the model's response to
the prompt and schema inputs themselves.

A population-scale extraction using variant A and the small model was
subsequently performed on the 331,793 admissions for downstream
applications outside the scope of this work.

\section{Results}\label{sec-results}

\subsection{Cross-prompt agreement is stable across sample
sizes}\label{sec-results-stability}

Before reporting the cross-prompt findings, the sample-size stability of
the agreement metrics was verified. The filtered-median cross-prompt
kappa \(\bar{\kappa}^{\mathrm{TriState}}\) on the small model is 0.70 on
the 150-note refinement set, 0.70 on the 150-note holdout set, 0.66 on
the 1,000-note validation set, 0.67 on the 500-note audit subset, and
0.65 on the 5,000-note extended validation set
(Figure\nobreakspace{}\ref{fig:sample-size-stability}). All five point
estimates sit within a narrow band. The refinement-to-holdout difference
of 0.01 percentage points
(Figure\nobreakspace{}\ref{fig:refinement-holdout}) is small relative to
that band, consistent with the optimization loop
(Section\nobreakspace{}\ref{sec-optimization-loop}) not producing a
refinement-to-holdout gap. The findings reported in the subsequent
sections are therefore not artifacts of any single sample size.

\subsection{Model size reshapes cross-prompt agreement rather than
raising it}\label{sec-results-model-size}

\begin{figure}
\centering
\pandocbounded{\includegraphics[keepaspectratio,alt={Cross-prompt agreement at two model sizes, on the same notes. Four agreement measures, each shown as paired bars for the small model (gpt-5.4-nano) and the full model (gpt-5.4) on the 1,500-note paired sample. The TriState and collapsed kappa bars are the filtered median over the per-field cross-prompt \textbackslash kappa values across the base-rate-filtered field set (Section). The admission-tag bar is admission-tag set agreement (mean per-note Jaccard), and the primary-admission-reason bar is the exact-match agreement on the single dominant tag; both are descriptive percent-agreement quantities and are not chance-corrected. Error bars are 95\% bootstrap confidence intervals. Paired sample, n = 1,500.}]{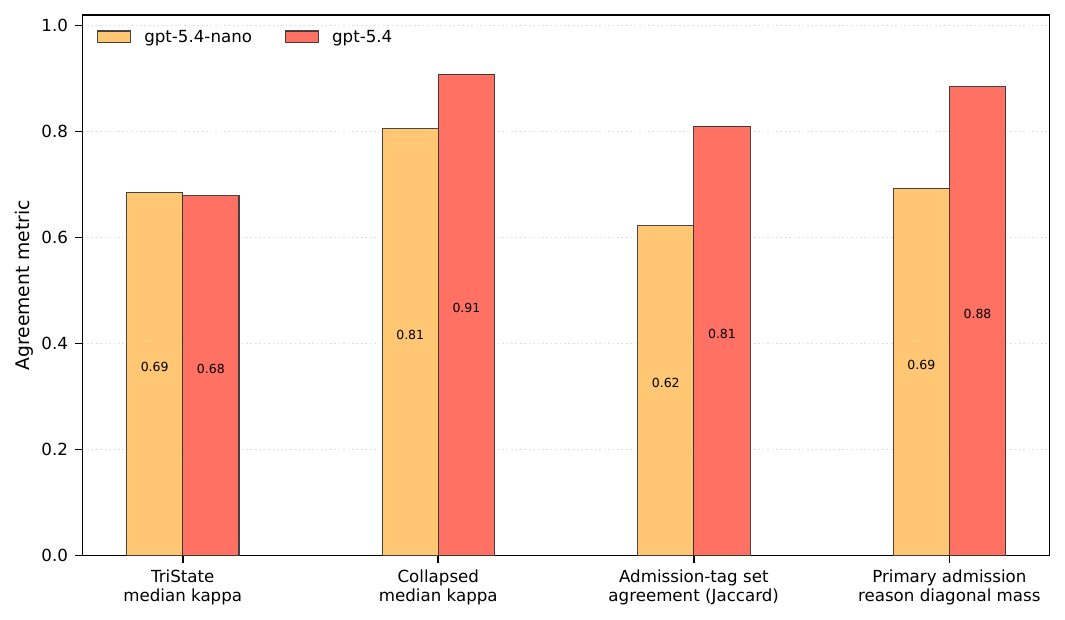}}
\caption{\textbf{Cross-prompt agreement at two model sizes, on the same
notes.} Four agreement measures, each shown as paired bars for the small
model (\texttt{gpt-5.4-nano}) and the full model (\texttt{gpt-5.4}) on
the 1,500-note paired sample. The TriState and collapsed kappa bars are
the filtered median over the per-field cross-prompt \(\kappa\) values
across the base-rate-filtered field set
(Section\nobreakspace{}\ref{sec-cross-prompt-metrics}). The
admission-tag bar is admission-tag set agreement (mean per-note
Jaccard), and the primary-admission-reason bar is the exact-match
agreement on the single dominant tag; both are descriptive
percent-agreement quantities and are not chance-corrected. Error bars
are 95\% bootstrap confidence intervals. Paired sample,
\(n = 1,500\).}\label{fig:cross-prompt-model-size}
\end{figure}

The cross-prompt agreement at the two model sizes was evaluated on the
1,500-note paired sample, where the small model and the full model both
produced extractions for the same notes under all three prompt variants
(Figure\nobreakspace{}\ref{fig:cross-prompt-model-size}).

On the TriState clinical documentation flags, the larger model does not
raise pooled cross-prompt agreement. The pooled-median cross-prompt
kappa is 0.69 at the small model (\texttt{gpt-5.4-nano}) and 0.68 at the
full model (\texttt{gpt-5.4}); the two levels are effectively the same.
The between-model comparison is made on the paired per-field statistic
rather than on the difference of these two pooled medians
(Section\nobreakspace{}\ref{sec-cross-prompt-metrics}): the median over
TriState fields of the per-field model-size difference
\(\overline{\Delta\kappa}^{\,\mathrm{per\text{-}field}}\) is -2.0
percentage points (95\% CI -8.0 to 4.4). The point estimate is slightly
negative and its confidence interval spans zero: under the TriState
schema, the typical field is no more reproducible at the larger model,
and if anything is marginally less so. Under the joint base-rate rule
(at least 10 positive votes at both model sizes),
\texttt{social\_support\_absent} is excluded from these filtered-median
summaries.

The aggregate stability is not the absence of an effect. The per-field
model-size difference is large and mixed in sign: the full model raises
cross-prompt agreement substantially on some TriState fields and lowers
it substantially on others, and these field-level changes approximately
cancel in the median. 8 of the 17 TriState fields have a negative
per-field difference, meaning the small-model variants agree with each
other more closely than the full-model variants do on those fields. The
slightly negative median is therefore the signature of a redistribution,
not a uniform decline: model size moves where in the schema the variants
agree, without moving the aggregate level. The per-field distribution is
the informative description of the model-size effect, and
Section\nobreakspace{}\ref{sec-results-per-field} presents it in full.

The behavior under binary collapse is different in kind. When the
TriState values are collapsed to a binary
\texttt{yes}-versus-\texttt{not\_yes} schema, the model-size effect
stops cancelling and becomes consistently positive: the per-field median
\(\overline{\Delta\kappa}^{\,\mathrm{per\text{-}field}}\) is +10.0
percentage points (95\% CI 7.0 to 13.6), and the pooled-median collapsed
kappa is 0.81 at the small model and 0.91 at the full model. The
contrast between the two schemas is the central observation: under the
TriState schema the model-size effect cancels across fields, whereas
under collapse it is positive across fields. Because the comparison uses
the same notes at both model sizes, the difference reflects the change
in model size and not a difference in sample composition. The per-field
structure behind this contrast is developed in
Section\nobreakspace{}\ref{sec-results-per-field} and its schema
interpretation in
Section\nobreakspace{}\ref{sec-results-collapse-structure}.

A complementary view holds each prompt variant fixed and measures how
far the small-model and full-model extractions agree with each other on
the same notes. This same-prompt cross-model agreement is itself low
under the TriState schema: the median over TriState fields of the
small-versus-full kappa is 0.32 for variant A, 0.39 for variant B, and
0.43 for variant C
(Figure\nobreakspace{}\ref{fig:per-variant-cross-model}). These values
are well below the cross-prompt agreement at a fixed model size
(0.69--0.68): holding the prompt fixed and changing the model size moves
the TriState output more than holding the model fixed and changing the
prompt does. Under binary collapse the same same-prompt cross-model
agreement rises to 0.70 (A), 0.69 (B), and 0.68 (C), confirming that
much of the small-versus-full disagreement lies on the
\texttt{no}-versus-\texttt{not\_documented} axis that collapse removes.

The model-size effect on the multi-class admission categorizations,
shown by the two right-hand bars of
Figure\nobreakspace{}\ref{fig:cross-prompt-model-size}, is of a
different character and is treated separately: the larger model brings
the variants markedly closer together on admission tags, and the
comparison of model size against prompt phrasing on these fields is
developed in Section\nobreakspace{}\ref{sec-results-tags-prevalence} and
Section\nobreakspace{}\ref{sec-results-dominant-confusion}. The TriState
reshaping reported here and the multi-class effect reported there are
two effects of the same configuration choice on two different field
types, and neither is subordinate to the other.

\subsection{The per-field model-size effect is heterogeneous in
sign}\label{sec-results-per-field}

The per-field model-size difference distribution is the primary
description of the model-size effect on TriState fields.
Figure\nobreakspace{}\ref{fig:per-field-deltas} shows \(\Delta\kappa_f\)
for each TriState field under both schemas; the orange bars report the
TriState schema and the green bars the binary-collapse re-analysis.
Under the TriState schema, the per-field difference is distributed
widely, from -29.6 percentage points for
\texttt{cardiac\_rehab\_referred} to +42.1 percentage points for
\texttt{cognitive\_impairment}, with the filtered-median
\(\overline{\Delta\kappa}^{\,\mathrm{per\text{-}field}}\) of -2.0
percentage points sitting just left of zero. The wide, sign-mixed
spread, not a uniform shift, is what the near-zero median reduces to a
single number.

The sign of the median is itself informative and is easy to overlook.
The typical TriState field is slightly less cross-prompt consistent at
the full model than at the small model, and 8 of the 17 fields move in
that direction. Fields with strongly negative TriState
\(\Delta\kappa_f\) include \texttt{aki\_present} (-39.7 pp),
\texttt{palliative\_care\_consult} (-38.2 pp),
\texttt{cardiac\_rehab\_referred} (-29.6 pp), \texttt{shock\_present}
(-25.0 pp), and \texttt{home\_health\_ordered} (-23.9 pp); on each, the
small-model variants agree with each other more closely than the
full-model variants do. Fields with strongly positive TriState
\(\Delta\kappa_f\) include \texttt{cognitive\_impairment} (+42.1 pp),
\texttt{financial\_hardship} (+35.5 pp), and
\texttt{goals\_of\_care\_flag} (+16.2 pp), where the full model brings
the variants closer together.

Read together with the collapse result from
Section\nobreakspace{}\ref{sec-results-model-size}, these two facts
describe a single mechanism. The per-field median is slightly negative
under the TriState schema (-2.0 pp) and clearly positive under collapse
(+10.0 pp). Model size therefore does not raise cross-prompt agreement
uniformly; it redistributes it. The larger model improves agreement on
the presence-versus-absence distinction, which is the part that survives
collapse, while on the same fields it slightly worsens agreement on the
\texttt{no}-versus-\texttt{not\_documented} distinction, which is the
part collapse removes. The aggregate level holds roughly constant
because these two movements offset.

\subsection{Binary collapse isolates a schema-imposed component of
disagreement}\label{sec-results-collapse-structure}

\begin{figure}
\centering
\pandocbounded{\includegraphics[keepaspectratio,alt={Per-field model-size differences in pairwise cross-prompt kappa, under the TriState schema and under binary collapse. Each row is one TriState clinical documentation flag; bars show the median across the three variant pairs of \textbackslash kappa\_\{p,f\}(\textbackslash text\{full\}) - \textbackslash kappa\_\{p,f\}(\textbackslash text\{small\}) (Section, Section). Orange bars report the TriState schema; green bars report the binary-collapse re-analysis (Section). Fields are ordered by their TriState model-size difference. The filtered-median model-size difference is -2.0 pp under the TriState schema and +10.0 pp under collapse. Paired sample, n = 1,500.}]{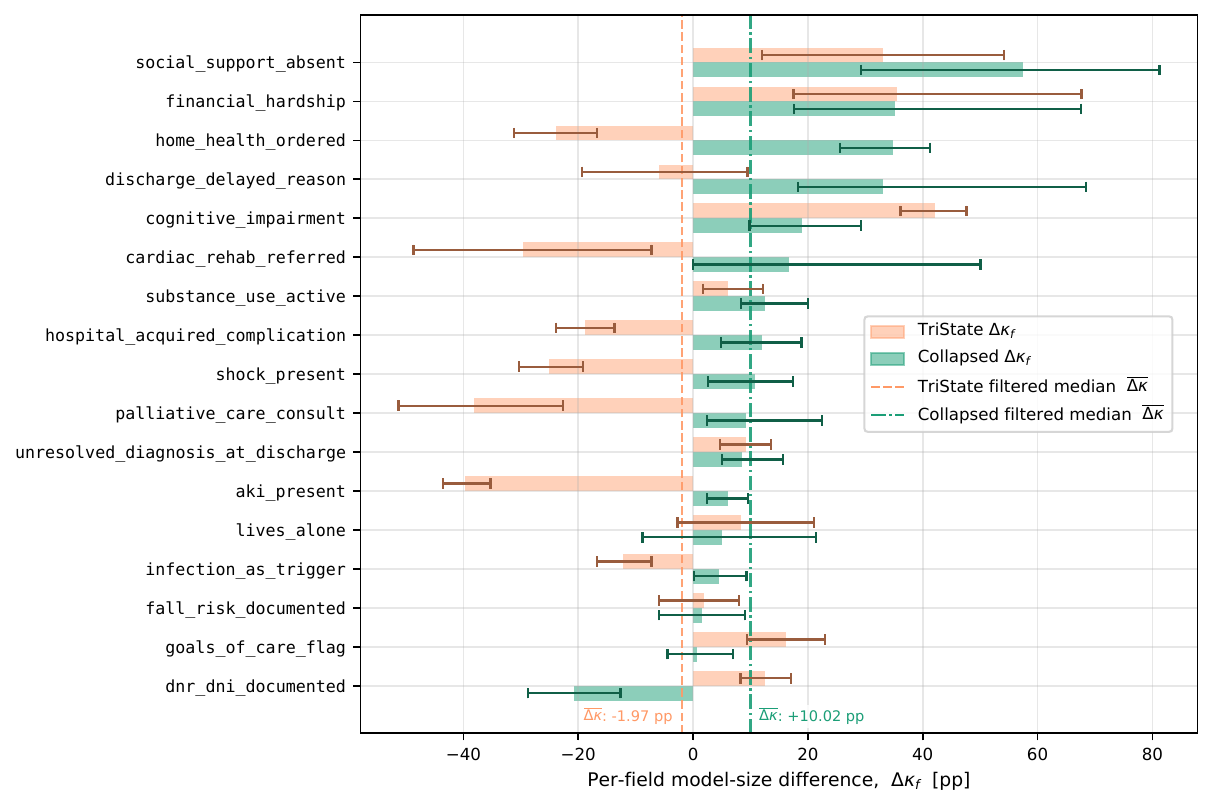}}
\caption{\textbf{Per-field model-size differences in pairwise
cross-prompt kappa, under the TriState schema and under binary
collapse.} Each row is one TriState clinical documentation flag; bars
show the median across the three variant pairs of
\(\kappa_{p,f}(\text{full}) - \kappa_{p,f}(\text{small})\)
(Section\nobreakspace{}\ref{sec-cross-prompt-metrics},
Section\nobreakspace{}\ref{sec-model-size-comparison}). Orange bars
report the TriState schema; green bars report the binary-collapse
re-analysis (Section\nobreakspace{}\ref{sec-binary-collapse}). Fields
are ordered by their TriState model-size difference. The filtered-median
model-size difference is -2.0 pp under the TriState schema and +10.0 pp
under collapse. Paired sample,
\(n = 1,500\).}\label{fig:per-field-deltas}
\end{figure}

The binary-collapse re-analysis varies one configuration choice, schema
granularity, while holding the prompts, the model, and the extracted
values fixed. Comparing the TriState and collapsed agreement on the same
extractions isolates the part of cross-prompt disagreement that is
attributable to the three-way schema from the part that is not. The
per-field median \(\Delta\kappa_f\) moves from -2.0 percentage points
under the TriState schema to +10.0 percentage points under collapse
(Figure\nobreakspace{}\ref{fig:per-field-deltas}). Under the TriState
schema the model-size effect is heterogeneous in sign and cancels in
aggregate; under collapse it is consistently positive across fields.

The pattern of per-field swings is the informative part. Fields with
strongly negative TriState \(\Delta\kappa_f\) all flip to positive under
collapse: \texttt{cardiac\_rehab\_referred} from -29.6 to +16.7 pp,
\texttt{aki\_present} from -39.7 to +6.1 pp, and
\texttt{palliative\_care\_consult} from -38.2 to +9.2 pp.~On these
fields the small-model variants disagreed primarily on the
\texttt{no}-versus-\texttt{not\_documented} axis. Collapse merges those
two values into one class, that axis of disagreement disappears, and the
small-model kappa rises sharply; the full-model kappa, already high
under TriState because the full model rarely confuses \texttt{no} and
\texttt{not\_documented}, rises only modestly. The per-field difference
therefore flips from negative to positive. Fields with positive TriState
\(\Delta\kappa_f\) keep their direction under collapse but shrink in
magnitude (\texttt{cognitive\_impairment} from +42.1 to +19.0 pp,
\texttt{goals\_of\_care\_flag} from +16.2 to +0.8 pp), because on these
fields the disagreement was already on the surviving
presence-versus-absence axis.

\begin{figure}
\centering
\pandocbounded{\includegraphics[keepaspectratio,alt={Per-variant-pair cross-prompt kappa under the TriState schema and under binary collapse. Each cell is \textbackslash kappa\_\{p,f\} for one variant pair p and one TriState field f, on the 1,500-note paired sample at the small model. The first panel reports the TriState schema, the second the binary-collapse re-analysis, and the third their per-cell difference \textbackslash kappa\_\{p,f\}\^{}\{\textbackslash mathrm\{collapsed\}\} - \textbackslash kappa\_\{p,f\}\^{}\{\textbackslash mathrm\{TriState\}\}. Paired sample, n = 1,500.}]{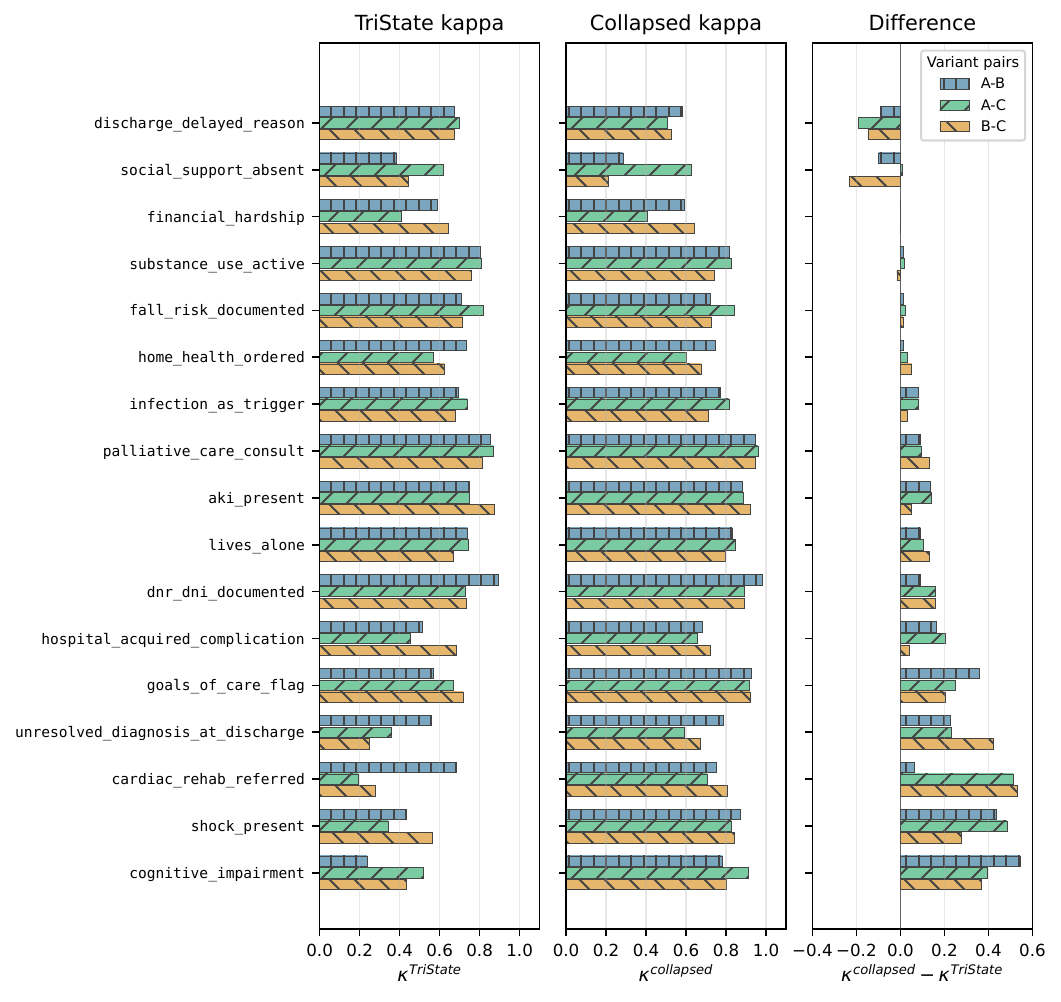}}
\caption{\textbf{Per-variant-pair cross-prompt kappa under the TriState
schema and under binary collapse.} Each cell is \(\kappa_{p,f}\) for one
variant pair \(p\) and one TriState field \(f\), on the 1,500-note
paired sample at the small model. The first panel reports the TriState
schema, the second the binary-collapse re-analysis, and the third their
per-cell difference
\(\kappa_{p,f}^{\mathrm{collapsed}} - \kappa_{p,f}^{\mathrm{TriState}}\).
Paired sample, \(n = 1,500\).}\label{fig:per-pair-kappa}
\end{figure}

The per-pair view in Figure\nobreakspace{}\ref{fig:per-pair-kappa}
confirms the pattern at finer resolution. Its third panel, the per-cell
difference
\(\kappa_{p,f}^{\mathrm{collapsed}} - \kappa_{p,f}^{\mathrm{TriState}}\),
shows where collapse changes agreement most. The largest positive
differences appear on \texttt{cognitive\_impairment},
\texttt{shock\_present}, and \texttt{cardiac\_rehab\_referred}, where
collapse raises every variant pair's kappa substantially. The smallest
appear on \texttt{substance\_use\_active},
\texttt{fall\_risk\_documented}, and \texttt{home\_health\_ordered},
where collapse changes kappa by only a few points for all pairs. A small
number of fields show negative differences
(\texttt{discharge\_delayed\_reason}, \texttt{social\_support\_absent},
\texttt{dnr\_dni\_documented}), where collapse removes signal from
variant pairs that were using the three-way structure successfully. This
per-field asymmetry, rather than the aggregate dissolution alone, is
what distinguishes a schema-imposed component from a generic coarsening
effect, and it is the empirical basis for the interpretation in
Section\nobreakspace{}\ref{sec-discussion-two-sources}.

\subsection{Most TriState disagreement is on the soft
axis}\label{sec-results-soft-disagreement}

Decomposing cross-variant TriState disagreement by type shows where the
disagreement sits
(Figure\nobreakspace{}\ref{fig:disagreement-decomposition}). Almost all
of it is soft, meaning at least one variant assigned
\texttt{not\_documented} while another assigned a definite value: soft
disagreements are 98.1\% of all cross-variant TriState disagreements,
and hard \texttt{yes}-versus-\texttt{no} flips, where the variants
assert opposite definite values, are the remaining 1.9\%. The soft total
splits into two sub-axes: 68.6\% of all disagreements are \texttt{no}
versus \texttt{not\_documented}, and 29.4\% are \texttt{yes} versus
\texttt{not\_documented}. The single largest component is therefore the
\texttt{no}-versus-\texttt{not\_documented} axis, the distinction
between an explicit negation and silence.

Binary collapse acts precisely on that largest component. Aggregated
across all TriState fields on the 6,500-note cross-variant pool, the
total disagreement count is 24,523 under the TriState schema and 7,688
under collapse, so 68.6\% of the disagreement dissolves when \texttt{no}
and \texttt{not\_documented} are merged. This dissolved fraction is the
same quantity as the \texttt{no}-versus-\texttt{not\_documented} share
above, since those are exactly the disagreements collapse removes; it is
one number seen two ways rather than two findings. The residual 31.4\%
is the disagreement that survives collapse, on the
\texttt{yes}-versus-not-yes axis. Most of the apparent cross-variant
disagreement on TriState fields is thus located on the schema-imposed
soft axis, and only a minority reflects disagreement about whether the
feature is present at all.

\subsection{Prompt phrasing shifts multi-class admission
categorization}\label{sec-results-tags-prevalence}

\begin{figure}
\centering
\pandocbounded{\includegraphics[keepaspectratio,alt={Per-tag prevalence across the three prompt variants, shown as absolute rate of tag assignment (panel A) and as share of total tag firings (panel B). Panel A: for each of the 47 admission-reason tags, three bars give the percentage of notes on which variant A, variant B, and variant C respectively assigned the tag (multi-label). Panel B: the same data normalized by the variant-specific total of tag assignments, controlling for differences in mean tags per note (means: 1.83 for A, 1.62 for B, 1.49 for C). Panel B controls for variant differences in mean tags per note. Pooled cross-variant sample, n = 6,500.}]{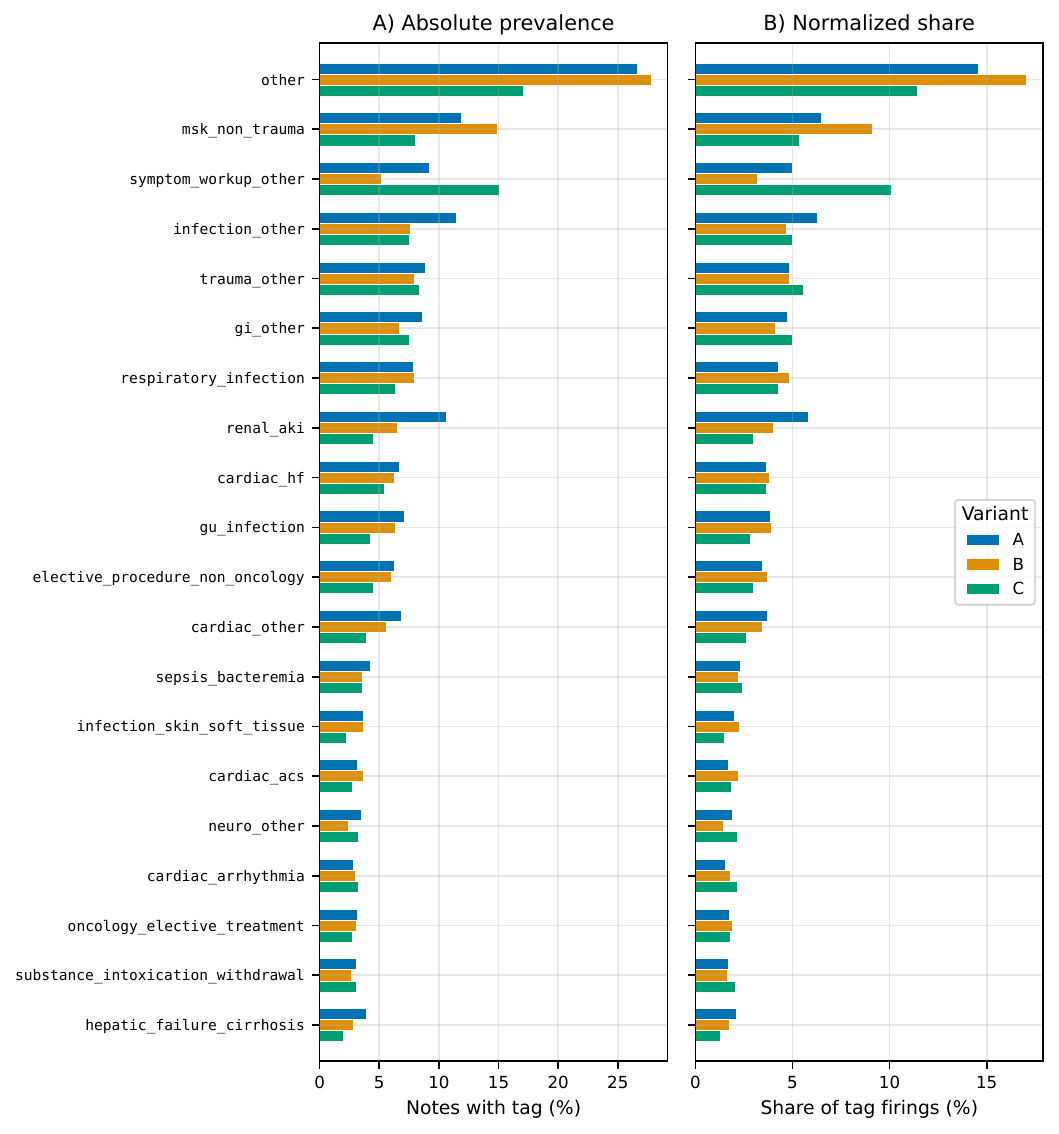}}
\caption{\textbf{Per-tag prevalence across the three prompt variants,
shown as absolute rate of tag assignment (panel A) and as share of total
tag firings (panel B).} Panel A: for each of the 47 admission-reason
tags, three bars give the percentage of notes on which variant A,
variant B, and variant C respectively assigned the tag (multi-label).
Panel B: the same data normalized by the variant-specific total of tag
assignments, controlling for differences in mean tags per note (means:
1.83 for A, 1.62 for B, 1.49 for C). Panel B controls for variant
differences in mean tags per note. Pooled cross-variant sample,
\(n = 6,500\).}\label{fig:tag-prevalence}
\end{figure}

The admission-tag prevalence by variant
(Figure\nobreakspace{}\ref{fig:tag-prevalence}, panel A) shows that the
three variants categorize admission reasons at systematically different
rates. Variant C assigns \texttt{symptom\_workup\_other} on 15.5\% of
notes, against 9.3\% for variant A and 5.1\% for variant B. Variant A
assigns \texttt{infection\_other} on 11.8\% of notes, against 7.6\% (B)
and 7.4\% (C). The clearest contrast is on the generic \texttt{other}
tag: variant C reaches for it on 16.6\% of notes, well below variant A
(26.2\%) and variant B (27.8\%). Variant C is the most specific of the
three, resorting to the undifferentiated catch-all least often.

Because the variants emit different numbers of tags per note (means 1.83
for A, 1.62 for B, 1.49 for C), panel B normalizes each tag's count by
the variant's total tag firings. The differences persist after
normalization: variant C's share of \texttt{symptom\_workup\_other}
firings is 10.4\% against 5.1\% (A) and 3.2\% (B), so the categorization
differences are not an artifact of total tag count. The differences are
consistent in direction across the validation samples; each variant
makes internally consistent but phrasing-dependent categorization
choices, and different phrasings push the model toward different
categorical readings of the same clinical content.

The same specificity gradient appears as a model-size effect, and it
sharpens under the larger model. On the 1,500-note paired sample, with
variant outputs averaged, the share of tag mass assigned to residual and
catch-all categories (the \texttt{other} tag together with the
\texttt{*\_other} family) falls from 44.1\% at the small model to 26.4\%
at the full model, driven mostly by the \texttt{other} tag itself
dropping from 13.8\% to 3.1\%. The mass moves into more specific
categories, for example \texttt{elective\_procedure\_non\_oncology}
(3.5\% to 8.9\%), \texttt{heme\_onc\_complication} (0.4\% to 4.2\%), and
\texttt{trauma\_fracture} (0.3\% to 2.9\%). Per-assignment confusion on
residual categories also falls at the full model for most of these tags,
for example \texttt{neuro\_other} (57\% to 24\%) and \texttt{gi\_other}
(46\% to 24\%), though not uniformly: \texttt{cardiac\_other} and
\texttt{trauma\_other} are flat or slightly higher. The full model thus
categorizes admission reasons more specifically on two counts, placing
less mass on catch-all categories and confusing the catch-all categories
it does use somewhat less often.

\subsection{Model size moves the dominant admission reason more than
phrasing does}\label{sec-results-dominant-confusion}

\begin{figure}
\centering
\pandocbounded{\includegraphics[keepaspectratio,alt={Cross-variant confusion structure for the primary admission reason, triangulated across the three variant pairs on the small model. A single confusion matrix is shown on the top 15 most-prevalent dominant tags; off-diagonal mass on rows for less-prevalent tags is shown in supplementary Figure, Figure, and Figure. Each (row, column) cell is divided into four triangular wedges, one per cross-variant comparison: the top wedge encodes A-vs-B, the left wedge A-vs-C, the right wedge B-vs-C, and the bottom wedge the average across the three pairs. The layout allows the reader to see at one cell whether all three pairs agree about a categorization or whether one pair diverges. Diagonal mass is 74.1\% (A-B), 69.9\% (A-C), and 67.8\% (B-C). Notable off-diagonal clusters include obstetric versus other, cardiac\_other versus cardiac\_valve\_disease, and psych\_mood\_anxiety versus symptom\_workup\_other. Pooled cross-variant sample, n = 6,500.}]{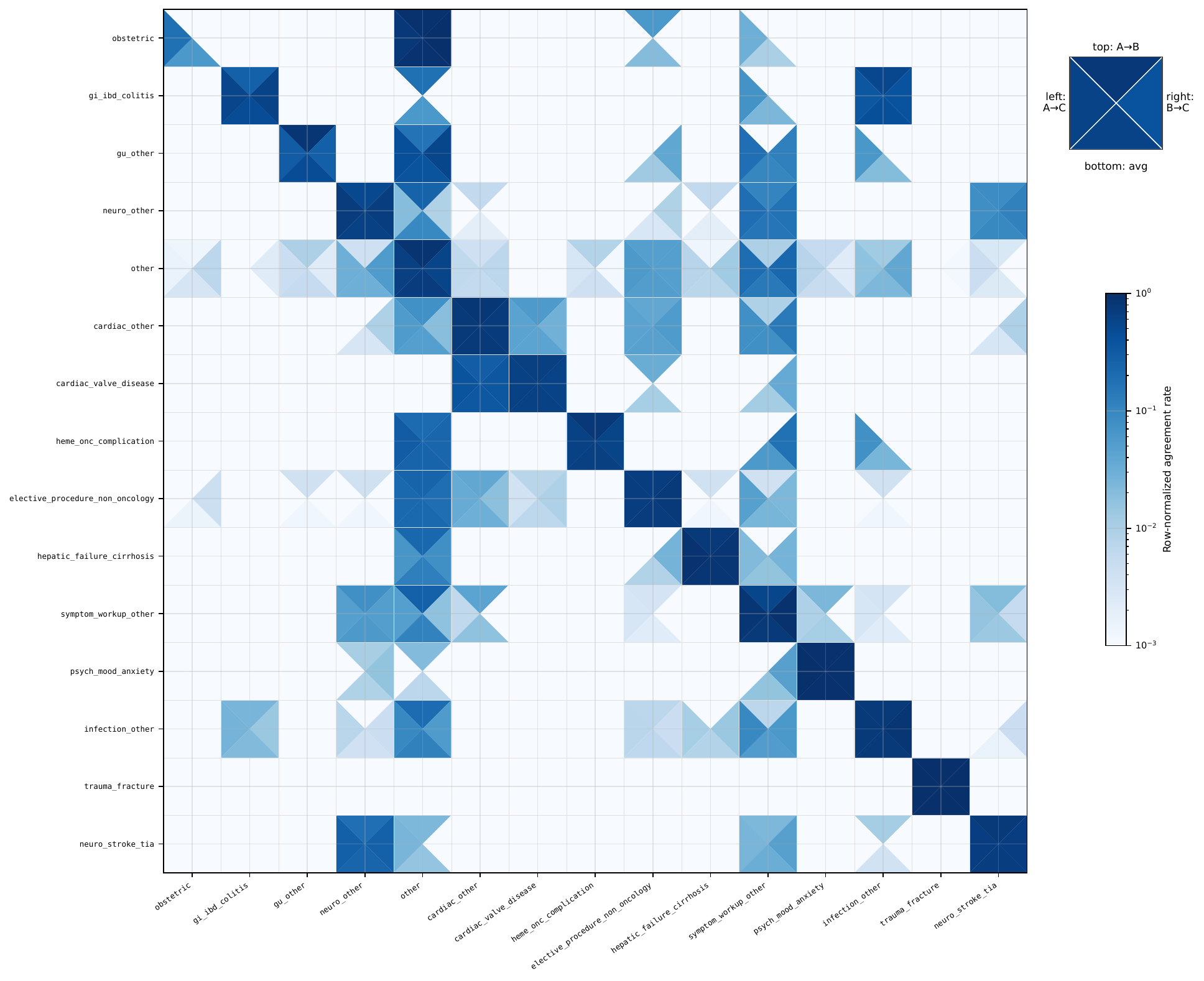}}
\caption{\textbf{Cross-variant confusion structure for the primary
admission reason, triangulated across the three variant pairs on the
small model.} A single confusion matrix is shown on the top 15
most-prevalent dominant tags; off-diagonal mass on rows for
less-prevalent tags is shown in supplementary
Figure\nobreakspace{}\ref{fig:cross-variant-ab},
Figure\nobreakspace{}\ref{fig:cross-variant-ac}, and
Figure\nobreakspace{}\ref{fig:cross-variant-bc}. Each (row, column) cell
is divided into four triangular wedges, one per cross-variant
comparison: the top wedge encodes A-vs-B, the left wedge A-vs-C, the
right wedge B-vs-C, and the bottom wedge the average across the three
pairs. The layout allows the reader to see at one cell whether all three
pairs agree about a categorization or whether one pair diverges.
Diagonal mass is 74.1\% (A-B), 69.9\% (A-C), and 67.8\% (B-C). Notable
off-diagonal clusters include \texttt{obstetric} versus \texttt{other},
\texttt{cardiac\_other} versus \texttt{cardiac\_valve\_disease}, and
\texttt{psych\_mood\_anxiety} versus \texttt{symptom\_workup\_other}.
Pooled cross-variant sample,
\(n = 6,500\).}\label{fig:admission-confusion}
\end{figure}

The single-choice primary admission reason
(Figure\nobreakspace{}\ref{fig:admission-confusion}) is the field on
which the model-size and phrasing effects can be compared most directly,
because both reduce to one quantity: the fraction of notes on which the
dominant tag changes. Holding the model fixed and changing only the
prompt, the cross-variant dominant-tag agreement at the full model is
high, 88.1\% (A-B), 88.4\% (A-C), and 88.7\% (B-C), so prompt phrasing
changes the dominant tag on roughly one note in eight. Holding the
prompt fixed and changing only the model size moves it far more often:
the same-prompt cross-model dominant-tag confusion has off-diagonal mass
of 45.7\% (variant A), 47.8\% (variant B), and 47.3\% (variant C) on the
1,500-note paired sample, meaning the small and full models select a
different dominant admission tag on close to half of all notes even when
the prompt is identical (the full \(47\times47\) same-prompt cross-model
confusion matrices are given in supplementary
Figures\nobreakspace{}\ref{fig:cross-model-a}, \ref{fig:cross-model-b},
and \ref{fig:cross-model-c}). For these three variants and two model
operating points, changing the model size is several times more
consequential for the dominant admission reason than changing the prompt
phrasing. This is the multi-class counterpart of the TriState reshaping
in Section\nobreakspace{}\ref{sec-results-model-size}: model size is the
dominant configuration choice on admission-reason categorization, just
as it redistributes agreement on the TriState fields.

At the small model, cross-variant agreement on the dominant tag is
74.1\% (A-B), 69.9\% (A-C), and 67.8\% (B-C); variants A and B are the
most consistent pair and A and C the least. The off-diagonal mass
concentrates in interpretable clusters
(Figure\nobreakspace{}\ref{fig:admission-confusion}): \texttt{obstetric}
versus \texttt{other}, where variants disagree about whether to tag
narrowly or under the catch-all; \texttt{cardiac\_other} versus
\texttt{cardiac\_valve\_disease}, where the two cardiac categories are
not cleanly separated; \texttt{symptom\_workup\_other} versus
\texttt{infection\_other}, where the same content is routed to a workup
category by one variant and an infection category by another. The
variants are not merely labeling at different rates but routing the same
admissions to different categorical destinations.

This disagreement is not spread evenly across the vocabulary. On the
pooled 6,500-note small-model sample, the per-tag confusion mass scales
strongly with per-tag prevalence: Spearman's correlation between the
two, averaged across the three variant pairs, is \(\rho = 0.92\) over
the 47 tags. The most-confused tags are the common ones, dominated by
residual or catch-all categories, and the least-confused are rare
conditions with sharp diagnostic anchors. Beyond this prevalence
scaling, the per-assignment confusion rate, confusion mass divided by
prevalence, is substantially higher for residual categories such as
\texttt{other} and the \texttt{*\_other} family than for sharply defined
categories such as diabetic ketoacidosis, COPD or asthma exacerbation,
and pulmonary embolism or deep vein thrombosis. Cross-variant
disagreement on the dominant admission reason therefore concentrates on
the categories without clear diagnostic anchors, where the prompt's
phrasing decides which of several plausible labels is applied.

\needspace{14\baselineskip}

\subsection{Enum field agreement is high with category-specific residual
confusion}\label{sec-results-enum}

The enum fields with closed value sets show high cross-variant agreement
overall, with structured residual confusion that varies by field.
Figure\nobreakspace{}\ref{fig:enum-mental-status},
Figure\nobreakspace{}\ref{fig:enum-functional-status}, and
Figure\nobreakspace{}\ref{fig:enum-discharge-condition} report the
variant-pair confusion structure for mental status, functional status at
discharge, and discharge condition category respectively, on the pooled
6,500-note cross-variant sample.

Functional status at discharge is the most consistent enum field, with
diagonal mass 96.4\% (A-B), 96.3\% (A-C), and 96.9\% (B-C). The
four-class scheme of \texttt{independent}, \texttt{assisted},
\texttt{dependent}, and \texttt{not\_documented} is reproduced
consistently across variants, and the only off-diagonal cluster at
appreciable scale is in the \texttt{dependent} row, where a minority of
\texttt{dependent} notes by one variant are read as \texttt{assisted} by
another.

Mental status shows a similar overall level, with diagonal mass 96.7\%
(A-B), 95.9\% (A-C), and 95.2\% (B-C). Off-diagonal mass concentrates
between \texttt{mild\_impairment} and \texttt{confused\_delirious}, and
between \texttt{not\_documented} and \texttt{intact}. The
\texttt{not\_documented} row is variant-pair-specific: the A-B pair
keeps most \texttt{not\_documented}-by-A notes as
\texttt{not\_documented}-by-B, whereas a smaller fraction of
\texttt{not\_documented}-by-B notes remain
\texttt{not\_documented}-by-C, with a sizeable share read as
\texttt{intact} instead. Variant C appears to treat the absence of an
explicit cognitive assessment as evidence of an intact baseline more
often than variant B does.

Discharge condition category shows the lowest agreement of the three,
with diagonal mass 71.5\% (A-B), 69.7\% (A-C), and 84.1\% (B-C).
Off-diagonal mass concentrates between \texttt{unchanged} and the more
clinically specific categories \texttt{stable}, \texttt{improved}, and
\texttt{deteriorated}, suggesting that \texttt{unchanged} functions as a
default that the variants invoke at different rates. The
\texttt{deteriorated} row shows substantial mass into \texttt{improved}
and \texttt{not\_documented}, indicating that the categorization of
clinical decline is variant-dependent, while \texttt{expired} is
essentially identical across variants because the underlying fact is
unambiguous in the source. The pattern across the three fields is
consistent with the broader findings: agreement is high where the
categories are concrete and observable, and the residual disagreement
concentrates on interpretive category boundaries.

\subsection{Labeling-function and LLM signals are
complementary}\label{sec-results-lf-llm}

The labeling-function ensemble provides an independent reference signal
against which the LLM extraction can be checked target by target, and
the relationship between the two is not uniform across targets
(Figures\nobreakspace{}\ref{fig:lf-icd-concordance},
\ref{fig:lf-regex-concordance}, and \ref{fig:aki-five-signal}).

For some targets, ICD codes capture broadly what the LLM extracts. On
acute kidney injury, the LLM-positive prevalence across variants ranges
from 14.4\% to 16.6\%, against an ICD-anchor prevalence of 15.4\%, with
variant-versus-ICD kappa in the 0.77 to 0.78 range. The five-signal pool
of three LLM variants, the ICD anchor, and the regex anchor on
\(n = 6,500\) notes contains 5,226 notes where no signal fires, 150
where ICD fires but no LLM variant does, and 103 where all three LLM
variants fire but ICD does not. That ICD-only-no-LLM cases outnumber
all-LLM-no-ICD cases is informative about each signal's coverage.

For other targets the asymmetry runs the other way. On cardiac heart
failure, the probability that the LLM fires given that ICD fires is high
(\(P(\mathrm{LLM} \mid \mathrm{ICD}) = 0.96\)), while the reverse is
lower (\(P(\mathrm{ICD} \mid \mathrm{LLM}) = 0.29\)). On trauma fracture
the probabilities flip, with ICD capturing fractures broadly and the LLM
flagging them narrowly (\(P(\mathrm{LLM} \mid \mathrm{ICD}) = 0.09\),
\(P(\mathrm{ICD} \mid \mathrm{LLM}) = 0.89\)). These asymmetries are
interpretable in clinical-practice terms
(Section\nobreakspace{}\ref{sec-discussion-implications}): ICD captures
billing-relevant codes that need not be the dominant admission reason,
while the LLM captures content prominent in the discharge narrative, so
disagreement between the signals is informative rather than purely
noisy.

Regex labeling functions have high specificity but limited coverage.
Regex prevalence on \texttt{aki\_present} is 3.5\%, well below the LLM
and ICD prevalence on the same target, so regex labels serve as a
high-precision anchor rather than a primary reference signal.

\section{Discussion}\label{sec-discussion}

The empirical findings in Section\nobreakspace{}\ref{sec-results}
describe how an LLM-based extraction responds to three configuration
choices: prompt phrasing, model size, and the granularity of the
categorical schema. The findings sort into three patterns. On the
TriState clinical flags, cross-prompt disagreement concentrates on a
schema-imposed axis that binary collapse removes. On the multi-class
admission categorizations, model size moves the assigned category far
more than prompt phrasing does. Across both field types, the model-size
effect is real but is a redistribution of agreement rather than a
uniform gain. This section interprets these patterns and draws out the
consequences for practitioners.

\subsection{Schema imposition and prompt phrasing as distinct sources of
disagreement}\label{sec-discussion-two-sources}

The findings isolate two patterns of disagreement with different
mechanical origins. The two arise from two configuration choices that
can be varied independently, the schema granularity and the prompt
phrasing, and the analysis varies each while holding the other fixed;
the patterns are discussed separately on that basis, not as orthogonal
components of a single decomposed quantity.

The first pattern is disagreement on the schema-imposed axis of the
TriState fields. The three-way
\texttt{yes}/\texttt{no}/\texttt{not\_documented} schema
(Section\nobreakspace{}\ref{sec-schema-design}) asks the extraction to
distinguish an explicit negation from silence, the assertion-status
distinction central to clinical NLP
\cite{uzuner2011i2b2,harkema2009context}. The data show that the
extraction recovers this distinction inconsistently. Decomposing the
cross-variant TriState disagreement
(Section\nobreakspace{}\ref{sec-results-soft-disagreement}) shows that
disagreements involving an affirmative \texttt{yes} are a small
minority, while the dominant axis of disagreement, 69\% of all
disagreements, is \texttt{no} versus \texttt{not\_documented}. Stated
operationally: when the note affirmatively documents a feature, the
variants agree on \texttt{yes}; the instability is in distinguishing an
explicitly declined feature from an unmentioned one. This is a statement
about the extraction's behavior, which the analysis measures directly,
rather than a claim about what the source text does or does not encode,
which it does not.

Binary collapse acts on exactly this axis, and its effect is the
empirical signature of the pattern. Merging \texttt{no} and
\texttt{not\_documented} post-hoc, without re-extraction, dissolves 69\%
of the cross-variant disagreement
(Section\nobreakspace{}\ref{sec-results-soft-disagreement}) and flips
the per-field model-size differences from negative under the TriState
schema to positive under collapse
(Section\nobreakspace{}\ref{sec-results-collapse-structure}). The
dissolution alone is not proof of schema imposition: any coarsening of a
label space can mechanically raise agreement by merging cases the finer
schema separated. What distinguishes the present finding is that the
aggregate dissolution is a raw disagreement count rather than a
chance-corrected quantity, and that the per-field flips are directional,
with the fields whose TriState model-size difference is negative turning
positive under collapse. The per-field collapse magnitudes are not read
as a severity ranking, because Cohen's kappa normalizes by each field's
class marginals, so collapse does not raise kappa by a fixed amount
across fields even where the underlying confusion is identical; the
magnitudes mix the schema-imposition signal with a base-rate-dependent
component. The interpretive weight rests on the aggregate dissolution
and the direction of the flips, neither of which is subject to that
caveat. The interpretation, that the schema asks the extraction to draw
a distinction it does not draw consistently across prompt phrasings, is
a hypothesis explaining the observed pattern; the pattern itself, the
asymmetric dissolution under collapse, is observed directly.

The second pattern is disagreement on the choice of category among the
multi-class admission-reason tags. The 47-tag vocabulary is fixed across
the three variants; the prompts differ only in how the assignment is
requested. Different phrasings produce different category preferences
(Section\nobreakspace{}\ref{sec-results-tags-prevalence}): variant A
assigns \texttt{infection\_other} on 11.8\% of notes against 7.6\% for B
and 7.4\% for C, and variant C, the most specific of the three, reaches
for the generic \texttt{other} tag least often. These preferences are
internally consistent within a variant and produce interpretable
confusion clusters
(Section\nobreakspace{}\ref{sec-results-dominant-confusion}). The
disagreement is uneven across the vocabulary, concentrating on residual
and catch-all categories without sharp diagnostic anchors, where the
phrasing decides which of several plausible labels is applied. The
mechanism that links a phrasing to a category preference is not
transparent from the prompts and would require targeted attribution
analysis to identify.

These two patterns affect different field types differently. The
TriState fields can carry schema-imposition disagreement because their
schema embeds the \texttt{no}-versus-\texttt{not\_documented}
distinction the extraction does not draw consistently. The
admission-reason tags carry phrasing-dependent category preferences
instead, because their schema offers only choices over a fixed
vocabulary. The enum fields with concrete observable categories,
functional status and mental status
(Section\nobreakspace{}\ref{sec-results-enum}), show high agreement
under both patterns, because they ask only for categorical
correspondences that are unambiguous in the source. A reported
cross-prompt kappa is therefore informative about the presence of
disagreement, but the remedy depends on which pattern is active:
schema-imposition disagreement is addressable through schema redesign,
such as collapse or moving the documentation-state question to a
separate auxiliary field, whereas phrasing-dependent disagreement is
addressable through prompt engineering or ensemble integration.

\subsection{Model size versus prompt
phrasing}\label{sec-discussion-model-size-vs-prompt}

The model-size effect is substantial but specific in form. On the
TriState fields it is the redistribution described in
Section\nobreakspace{}\ref{sec-results-per-field}: the larger model
improves agreement on the presence-versus-absence distinction that
survives collapse and slightly worsens it on the
\texttt{no}-versus-\texttt{not\_documented} axis that collapse removes,
so the per-field median moves from -2.0 percentage points under the
TriState schema to +10.0 percentage points under collapse. On the
multi-class admission categorizations it is a more direct improvement:
the full model brings the variants markedly closer together on both the
dominant tag and the secondary tag set
(Section\nobreakspace{}\ref{sec-results-dominant-confusion}), and it
places less tag mass on residual catch-all categories while confusing
those categories less often when it does use them
(Section\nobreakspace{}\ref{sec-results-tags-prevalence}).

The comparison between the two configuration choices is sharpest on the
single-choice primary admission reason, where both reduce to one
quantity, the fraction of notes on which the dominant tag changes
(Section\nobreakspace{}\ref{sec-results-dominant-confusion}). Changing
only the model size while holding the prompt fixed changes the dominant
tag on close to half of all notes; changing only the prompt while
holding the model fixed changes it on roughly one note in eight. For the
three prompts and two model operating points studied here, model size is
several times the larger driver of variability in the dominant admission
reason. This is the multi-class counterpart of the TriState result:
model size is the dominant configuration choice for admission-reason
categorization, as it is the agent of the redistribution on the TriState
fields. The two are co-equal findings about the same configuration
choice acting on two different field types, not a single effect seen
twice.

\subsection{Implications for clinical NLP
practitioners}\label{sec-discussion-implications}

These findings carry methodological implications for practitioners
building LLM-based extraction pipelines for cohort construction, patient
classification, or feature engineering. The discussion applies
specifically to zero-shot prompt-engineering deployments, in which the
prompt is the primary lever the engineer holds. Parameter-efficient
fine-tuning or instruction-tuning on task-specific data would likely
reduce the prompt-phrasing sensitivity reported here
\cite{hu2026information,hsu2025weak}, and the cross-prompt agreement
findings are not intended to predict the behavior of fine-tuned systems.

Cross-prompt agreement is most usefully read as a stability measure, not
an accuracy measure \cite{errica2025quantifying,razavi2025benchmarking}.
Prompt-stability and task-accuracy are weakly correlated at best: two
variants can agree on a wrong categorization, and two variants can
disagree on cases where the source genuinely admits multiple readings
\cite{errica2025quantifying}. A low cross-prompt kappa on a field
indicates that the field's value would not be reproducible if the prompt
were redrafted by a different engineer; it does not by itself indicate
incorrectness, and a high kappa indicates reproducibility without
bearing on correspondence to clinical reality. The interpretation
depends on the downstream use.

For applications where the extracted variables feed a machine-learning
model, the two interpretations interact. A feature with low cross-prompt
agreement that is ranked highly by a downstream model should prompt the
question of whether the model is exploiting variant-specific phrasing
artifacts rather than clinical content; the phrasing-dependent category
preferences of Section\nobreakspace{}\ref{sec-results-tags-prevalence}
are a concrete mechanism by which such artifacts can leak into
downstream signals. Conversely, a clinically meaningful feature with
unexpectedly low importance may be losing signal to cross-prompt
instability. Cross-prompt agreement is therefore worth reporting
alongside feature-importance scores in studies that use LLM extractions
for downstream prediction.

For cohort construction and patient classification, the binary-collapse
analysis offers a schema-validation tool. A practitioner adopting a
three-way \texttt{yes}/\texttt{no}/\texttt{not\_documented} schema for
documentation-quality reasons can check, per field, whether collapse
substantially changes the cross-prompt agreement; a large positive shift
indicates that the three-way distinction is being supplied by the model
rather than recovered consistently from the source. On such fields, a
downstream consumer of the binary \texttt{yes}-versus-not-\texttt{yes}
signal is likely to benefit from collapsing before consumption, and the
\texttt{no}-versus-\texttt{not\_documented} distinction, where it is
wanted, may be more reliably obtained through a targeted auxiliary field
than through the TriState schema.

This bears on a choice made in the present work: the population-scale
extraction
(Section\nobreakspace{}\ref{sec-variant-selection-production}) was run
with the small model. The decision was partly one of cost, but it is
consistent with the evidence here. The model-size effect on cross-prompt
stability is a redistribution rather than a uniform gain
(Section\nobreakspace{}\ref{sec-results-per-field}): under the TriState
schema the two model sizes reach the same pooled agreement and the
per-field median model-size difference is indistinguishable from zero,
while the clear positive effect under collapse is concentrated on the
\texttt{no}-versus-\texttt{not\_documented} axis, precisely the axis a
documentation-quality consumer can choose to collapse. The
labeling-function analysis
(Section\nobreakspace{}\ref{sec-results-lf-llm}) reinforces this reading
from another direction: combining the LLM extraction with independent
labeling functions yields a consensus signal no longer dependent on a
single prompt's phrasing
\cite{ratner2017snorkel,smith2022prompting,fries2021trove}, and the
asymmetry between ICD-only and LLM-only positive cases on AKI (150
versus 103 on an \(n = 6,500\) pool) shows that billing codes and
discharge-summary narrative capture different aspects of the clinical
event, so disagreement between the two is informative rather than noise.

The model-size comparison also has an implication beyond capability. The
two model sizes are two points in a vendor product range that turns over
rapidly, and commercial model versions are routinely deprecated and
replaced; the operating point used for a population-scale extraction may
not remain available for a subsequent one. Because changing the model
moves the dominant admission categorization on close to half of all
notes, more than changing the prompt phrasing does, a model version is
not an interchangeable implementation detail but a controlled dependency
of the extraction. Pipelines that require reproducibility across
re-extractions therefore have reason to pin a fixed model version, or to
self-host a fixed model, and to treat any model substitution as a change
that warrants re-running the agreement and disagreement diagnostics
rather than as a transparent upgrade.

\subsection{Methodology
contributions}\label{sec-discussion-contributions}

The methodology assembles into a deployment-readiness check applicable
without human-annotated ground truth, contributing to clinical NLP
practice in five ways.

Cross-prompt kappa, measured at multiple sample sizes with a base-rate
filter for low-prevalence fields
(Section\nobreakspace{}\ref{sec-cross-prompt-metrics}), provides a
deployment-time stability metric that requires no human labels and
replicates on any extraction task with multiple prompt variants
targeting one schema. The paired same-note model-size comparison
(Section\nobreakspace{}\ref{sec-model-size-comparison}) isolates the
model-size effect from sample-composition confounds, available wherever
budget permits re-extracting the same admissions with a second model.
The binary-collapse re-analysis
(Section\nobreakspace{}\ref{sec-binary-collapse}) is a schema-validation
diagnostic: re-map the categorical values, recompute agreement, and read
the per-field shifts to see which schema distinctions the model is
supplying rather than recovering. The autonomous optimization loop
(Section\nobreakspace{}\ref{sec-optimization-loop}) addresses systematic
cross-variant divergence with four deterministic guards (cardinality,
value-set membership, no field renaming, no field removal) that prevent
the rewrite step from silently altering the extraction's structural
contract; it builds on general-domain prompt rewriting
\cite{ye2023pe2,wu2025automedprompt} and clinician-in-the-loop
refinement \cite{hein2025iterative,huang2024chatgpt}, with the
contribution being the pairing of autonomous rewriting with structural
safety constraints applied at the level of observed disagreement
clusters. The integration of LLM extractions with independent labeling
functions through a Snorkel LabelModel
(Section\nobreakspace{}\ref{sec-lf-ensemble}) \cite{ratner2017snorkel}
yields a consensus signal more stable than any single variant;
practitioners adopting it should note that the LLM is itself one
labeling function, so the consensus is not an LLM-independent reference,
but it is a more stable aggregation target than a single prompt's
output.

\section{Limitations}\label{sec-limitations}

Several limitations bound the interpretation of these findings.

Cross-prompt kappa measures the stability of extraction under
prompt-engineering variation, not its correctness relative to clinical
reality. High cross-prompt agreement is a reproducibility property, not
an accuracy guarantee; validating clinical correctness would require
human annotation of the source documents against a curated reference,
which was not performed here.

The paired same-note comparison
(Section\nobreakspace{}\ref{sec-model-size-comparison}) treats the two
model sizes as two operating points to be compared, with neither serving
as a ground truth for the other. Both are LLMs subject to the same
families of error, so a small-versus-full difference is a difference
between two operating points and not a measurement of small-model error
against a validated reference.

The schema-imposition reading
(Section\nobreakspace{}\ref{sec-discussion-two-sources}) is a hypothesis
explaining the observed dissolution of disagreement under binary
collapse; the pattern is measured directly, but the explanation is not
directly tested and would benefit from targeted auxiliary analysis on
selected fields. Relatedly, the binary-collapse re-analysis re-labels
the existing TriState extractions rather than re-running extraction
under a prompt that asks directly for a binary schema, so whether a
direct binary extraction would reproduce the same agreement profile is
an open question.

The experimental scope is bounded in three ways. The three prompt
variants sample but do not exhaust the space of plausible phrasings for
the same schema. The two model operating points are both from the
\texttt{gpt-5.4} family, and the gap between them is not resolved by
intermediate sizes; no comparison is made across providers or
architectures. The clinical context is MIMIC-IV v3.1, which reflects US
documentation and billing practice in a single tertiary-care setting,
and the analysis covers discharge summaries only. Field-level findings
(which fields show schema-imposition disagreement, which categorical
confusion clusters arise) may not transfer to other documentation
contexts or note types without revalidation; the methodology
generalizes, but the specific empirical findings are conditional on the
source data and the (schema, prompt-set, model-family) combination
tested.

The variant-selection metric
(Section\nobreakspace{}\ref{sec-variant-selection-production}) measures
each variant's agreement with a Snorkel-aggregated consensus of which
the variant is itself an input, introducing a degree of self-consistency
the metric does not separate from substantive agreement. The
cross-prompt findings reported here are pairwise across all three
variants and do not depend on which variant was selected for production,
but the selection metric has this known limitation as a stand-alone
variant-quality measure.

Per-field cross-variant agreement varies even under the most favorable
conditions tested. On the filtered field set the median pairwise kappa
under the full model and binary collapse is 0.91, but the lower tail
sits well below this; the four lowest-agreement fields under these
conditions are \texttt{dnr\_dni\_documented},
\texttt{unresolved\_diagnosis\_at\_discharge},
\texttt{hospital\_acquired\_complication}, and
\texttt{fall\_risk\_documented}, which warrant individual scrutiny
before downstream use. The free-text evidence field and the integer
count fields (Section\nobreakspace{}\ref{sec-schema-design}) were
retained in the extraction but not analyzed, and their behavior is
unaddressed here.

\clearpage
\bibliography{../references}

\clearpage
\section*{Supplementary Material}
\appendix
\setcounter{section}{0}
\renewcommand{\thesection}{\Alph{section}}
\renewcommand{\thesubsection}{\Alph{section}.\arabic{subsection}}
\setcounter{figure}{0}
\renewcommand{\thefigure}{S\arabic{figure}}
\setcounter{table}{0}
\renewcommand{\thetable}{S\arabic{table}}
\floatplacement{figure}{H}

\section{Supplementary Material}\label{supplementary-material}

This supplement provides the schema and sampling definitions referenced
in the Methods, followed by the analyses that extend the main-text
figures, presented in the order in which the main text invokes them. The
opening definitions section gives the sample design
(Table\nobreakspace{}\ref{tbl:splits}), the schema definitions for the
admission-reason vocabulary
(Table\nobreakspace{}\ref{tbl:admission-tags}), the TriState clinical
flags (Table\nobreakspace{}\ref{tbl:tristate-fields}), and the
enumerated fields (Table\nobreakspace{}\ref{tbl:enum-fields}), together
with the labeling-function definitions
(Table\nobreakspace{}\ref{tbl:icd-lfs} and
Table\nobreakspace{}\ref{tbl:regex-lfs}). The analytical sections then
expand, in main-text reference order, on sample-size stability of the
cross-prompt agreement
(Section\nobreakspace{}\ref{sec-results-stability}), the
refinement-to-holdout generalization check
(Section\nobreakspace{}\ref{sec-optimization-loop}), the same-prompt
cross-model agreement
(Section\nobreakspace{}\ref{sec-results-model-size}), the disagreement
decomposition under the TriState schema versus binary collapse
(Section\nobreakspace{}\ref{sec-results-soft-disagreement}), the
full-vocabulary admission-tag confusion matrices
(Section\nobreakspace{}\ref{sec-results-dominant-confusion}), and the
enum-field cross-variant agreement
(Section\nobreakspace{}\ref{sec-results-enum}), followed by the
labeling-function and LLM concordance analysis
(Section\nobreakspace{}\ref{sec-results-lf-llm}). The verbatim text of
the three prompt variants is provided in the final section.

\clearpage

\section{Definitions and Source
Tables}\label{definitions-and-source-tables}

This section collects the sample-design, schema, and labeling-function
definitions referenced by the Methods and Results.

\subsection{Sample Design}\label{sample-design}

\begin{longtable}[]{|>{\raggedright\arraybackslash}p{(\linewidth - 6\tabcolsep) * \real{0.2000}}|>{\raggedright\arraybackslash}p{(\linewidth - 6\tabcolsep) * \real{0.0800}}|>{\raggedright\arraybackslash}p{(\linewidth - 6\tabcolsep) * \real{0.4200}}|>{\raggedright\arraybackslash}p{(\linewidth - 6\tabcolsep) * \real{0.3000}}|}
\caption{\label{tbl:splits} List of study samples, fixed sizes, and
methodological purposes for all main and supplement
analyses.}\tabularnewline\hline
\hline
\begin{minipage}[b]{\linewidth}\raggedright
Sample name
\end{minipage} & \begin{minipage}[b]{\linewidth}\raggedright
\(N\)
\end{minipage} & \begin{minipage}[b]{\linewidth}\raggedright
Purpose
\end{minipage} & \begin{minipage}[b]{\linewidth}\raggedright
Referenced in section
\end{minipage} \\
\hline
\endfirsthead
\hline
\begin{minipage}[b]{\linewidth}\raggedright
Sample name
\end{minipage} & \begin{minipage}[b]{\linewidth}\raggedright
\(N\)
\end{minipage} & \begin{minipage}[b]{\linewidth}\raggedright
Purpose
\end{minipage} & \begin{minipage}[b]{\linewidth}\raggedright
Referenced in section
\end{minipage} \\
\hline
\endhead
\hline
\endlastfoot
Smoke & 200 & Initial prompt drafting and smoke validation before
three-variant extraction & Methods (three prompt variants) \\
Refinement & 150 & Prompt refinement and disagreement-audit development
set & Methods (splits), refinement-to-holdout generalization;
sample-size stability \\
Holdout & 150 & Firewalled single-touch validation set & Methods
(holdout), refinement-to-holdout generalization; sample-size
stability \\
Methodology & 1,000 & Variant comparison and methodology validation
sample & Methods/Results, refinement-to-holdout generalization;
sample-size stability \\
Methodology 5k Audit & 500 & Pre-production audit subset for tri-variant
diagnostics & Methods/Results, disagreement decomposition; sample-size
stability \\
Methodology Paired & 1,500 & Same-note paired small-vs-full model-size
analyses & Results model-size sections,
Figure\nobreakspace{}\ref{fig:cross-prompt-model-size}/Figure\nobreakspace{}\ref{fig:per-field-deltas}
and Figure\nobreakspace{}\ref{fig:per-variant-cross-model},
Figure\nobreakspace{}\ref{fig:cross-model-a}--Figure\nobreakspace{}\ref{fig:cross-model-c} \\
Extended & 5,000 & Large post-lock tri-variant validation sample &
Results stability and confusion analyses,
Figure\nobreakspace{}\ref{fig:per-pair-kappa}/Figure\nobreakspace{}\ref{fig:tag-prevalence}/Figure\nobreakspace{}\ref{fig:admission-confusion}
and
Figure\nobreakspace{}\ref{fig:cross-variant-ab}--Figure\nobreakspace{}\ref{fig:cross-variant-bc},
Figure\nobreakspace{}\ref{fig:sample-size-stability} \\
Pooled cross-variant & 6,500 & Intersection-pooled A/B/C sample (1k +
500 + extended) for cross-variant diagnostics &
Figure\nobreakspace{}\ref{fig:per-pair-kappa}/Figure\nobreakspace{}\ref{fig:tag-prevalence}/Figure\nobreakspace{}\ref{fig:admission-confusion},
Figure\nobreakspace{}\ref{fig:lf-icd-concordance}--Figure\nobreakspace{}\ref{fig:aki-five-signal},
Figure\nobreakspace{}\ref{fig:cross-variant-ab}--Figure\nobreakspace{}\ref{fig:cross-variant-bc},
Figure\nobreakspace{}\ref{fig:disagreement-decomposition},
Figure\nobreakspace{}\ref{fig:enum-mental-status}--Figure\nobreakspace{}\ref{fig:enum-discharge-condition} \\
Production & 331,793 & Population-scale extraction cohort & Methods
production section \\
\end{longtable}

\clearpage

\subsection{Admission-Reason Tag
Vocabulary}\label{admission-reason-tag-vocabulary}

\begin{longtable}[]{|>{\raggedright\arraybackslash}p{(\linewidth - 4\tabcolsep) * \real{0.3000}}|>{\raggedright\arraybackslash}p{(\linewidth - 4\tabcolsep) * \real{0.4000}}|>{\raggedright\arraybackslash}p{(\linewidth - 4\tabcolsep) * \real{0.3000}}|}
\caption{\label{tbl:admission-tags} Forty-seven admission-reason tags
with definitions and anchor ICD code families used for admission-reason
analysis.}\tabularnewline\hline
\hline
\begin{minipage}[b]{\linewidth}\raggedright
Tag ID
\end{minipage} & \begin{minipage}[b]{\linewidth}\raggedright
Definition
\end{minipage} & \begin{minipage}[b]{\linewidth}\raggedright
Anchor codes or patterns
\end{minipage} \\
\hline
\endfirsthead
\hline
\begin{minipage}[b]{\linewidth}\raggedright
Tag ID
\end{minipage} & \begin{minipage}[b]{\linewidth}\raggedright
Definition
\end{minipage} & \begin{minipage}[b]{\linewidth}\raggedright
Anchor codes or patterns
\end{minipage} \\
\hline
\endhead
\hline
\endlastfoot
\texttt{cardiac\_hf} & Acute decompensated heart failure or cardiogenic
pulmonary edema & ICD-10-CM: I50, I11.0, I13.0, I13.2; ICD-9-CM: 428,
402.01, 402.11, 402.91, 404.01, 404.03, 404.11, 404.13, 404.91,
404.93 \\
\texttt{cardiac\_acs} & Acute coronary syndrome (STEMI, NSTEMI, unstable
angina) & ICD-10-CM: I21, I22; ICD-9-CM: 410, 411 \\
\texttt{cardiac\_arrhythmia} & New or worsening arrhythmia as primary
driver (AFib with RVR, VT, symptomatic bradyarrhythmia) & ICD-10-CM:
I47, I48, I49; ICD-9-CM: 427 \\
\texttt{cardiac\_htn\_emergency} & Hypertensive emergency or urgency
with end-organ involvement & --- \\
\texttt{cardiac\_valve\_disease} & Symptomatic valvular disease
admission (e.g., severe AS, acute MR) & --- \\
\texttt{cardiac\_other} & Other cardiac reason not matching the above
(pericarditis, myocarditis, etc.) & --- \\
\texttt{respiratory\_infection} & Community or hospital acquired
pneumonia, bronchitis, etc. & ICD-10-CM: J12, J13, J14, J15, J16, J17,
J18, J20, J21, J22; ICD-9-CM: 480, 481, 482, 483, 484, 485, 486, 487 \\
\texttt{respiratory\_copd\_asthma\_\newline exacerbation} & COPD or
asthma exacerbation & ICD-10-CM: J44, J45, J46; ICD-9-CM: 491, 492, 493,
496 \\
\texttt{respiratory\_pe\_dvt} & Pulmonary embolism or DVT & ICD-10-CM:
I26, I82; ICD-9-CM: 415.1, 451, 453 \\
\texttt{respiratory\_failure\_other} & Hypoxemic or hypercapnic
respiratory failure without infectious or COPD/asthma driver & --- \\
\texttt{gi\_bleed} & Upper or lower GI bleed & ICD-10-CM: K25.0, K25.2,
K25.4, K25.6, K26.0, K26.2, K26.4, K26.6, K27.0, K27.2, K27.4, K27.6,
K28.0, K28.2, K28.4, K28.6, K29.01, K29.21, K29.31, K29.41, K29.51,
K29.61, K29.71, K29.81, K29.91, K62.5, K92.0, K92.1, K92.2; ICD-9-CM:
530.82, 531.00, 531.01, 531.20, 531.21, 531.40, 531.41, 531.60, 531.61,
532, 533, 534, 535.01, 535.11, 535.21, 535.31, 535.41, 535.51, 535.61,
535.71, 569.3, 578 \\
\texttt{gi\_obstruction\_ileus} & Small or large bowel obstruction,
ileus, volvulus & --- \\
\texttt{gi\_pancreatitis} & Acute pancreatitis & --- \\
\texttt{gi\_ibd\_colitis} & IBD flare, C. diff or other colitis & --- \\
\texttt{hepatic\_failure\_cirrhosis} & Acute or chronic liver failure,
hepatic encephalopathy, cirrhosis decompensation & ICD-10-CM: K70, K71,
K72, K73, K74, K76.6, K76.7; ICD-9-CM: 571, 572, 573 \\
\texttt{gi\_other} & Other GI reason (gastritis, hernia, etc.) & --- \\
\texttt{renal\_aki} & Acute kidney injury as primary admission driver &
--- \\
\texttt{renal\_ckd\_esrd\_crisis} & CKD/ESRD complication driving
admission (uremia, fluid overload, missed dialysis) & --- \\
\texttt{gu\_infection} & UTI, pyelonephritis, prostatitis & --- \\
\texttt{gu\_other} & Other genitourinary reason & --- \\
\texttt{sepsis\_bacteremia} & Sepsis or bacteremia as the primary
admission reason regardless of source & ICD-10-CM: A40, A41, R65.2;
ICD-9-CM: 038, 995.91, 995.92, 790.7 \\
\texttt{infection\_skin\_soft\_tissue} & Cellulitis, abscess,
necrotizing fasciitis & --- \\
\texttt{infection\_cns} & Meningitis, encephalitis, brain abscess &
--- \\
\texttt{infection\_other} & Other infection (endocarditis,
osteomyelitis, fungemia, etc.) & --- \\
\texttt{neuro\_stroke\_tia} & Ischemic or hemorrhagic stroke, TIA &
ICD-10-CM: I60, I61, I62, I63, I64, G45; ICD-9-CM: 430, 431, 432, 433,
434, 435, 436 \\
\texttt{neuro\_seizure} & Seizure or status epilepticus & --- \\
\texttt{neuro\_altered\_mental\_status} & Encephalopathy or delirium as
primary reason & --- \\
\texttt{neuro\_other} & Other neurologic reason (Parkinson, MS,
neuromuscular, etc.) & --- \\
\texttt{metabolic\_dka\_hhs} & Diabetic ketoacidosis or hyperosmolar
hyperglycemic state & ICD-10-CM: E10.1, E11.0, E11.1, E13.0, E13.1;
ICD-9-CM: 250.1, 250.2, 250.3 \\
\texttt{metabolic\_electrolyte\_crisis} & Severe electrolyte disturbance
(hyperkalemia, hyponatremia crisis, etc.) & --- \\
\texttt{endocrine\_other} & Other endocrine reason (thyroid storm,
adrenal crisis, etc.) & --- \\
\texttt{heme\_anemia\_bleed} & Severe anemia or hemorrhage not clearly
GI or trauma & --- \\
\texttt{heme\_onc\_complication} & Complication of cancer or its
treatment (neutropenic fever, tumor lysis, etc.) & --- \\
\texttt{oncology\_elective\_treatment} & Planned chemotherapy,
transplant, or oncology procedure admission & ICD-10-CM: Z51.0, Z51.1;
ICD-9-CM: V58.0, V58.1 \\
\texttt{trauma\_fracture} & Fracture from trauma & ICD-10-CM: S02, S12,
S22, S32, S42, S52, S62, S72, S82, S92; ICD-9-CM: 800, 801, 802, 803,
804, 805, 806, 807, 808, 809, 810, 811, 812, 813, 814, 815, 816, 817,
818, 819, 820, 821, 822, 823, 824, 825, 826, 827, 828, 829 \\
\texttt{trauma\_other} & Other trauma (blunt, penetrating, falls without
fracture) & --- \\
\texttt{msk\_non\_trauma} & MSK admission without trauma (joint
infection, non-traumatic back pain workup, etc.) & --- \\
\texttt{psych\_mood\_anxiety} & Depression, anxiety, bipolar mood
episode (non-psychotic) & --- \\
\texttt{psych\_psychosis\_crisis} & Psychotic episode, suicidal ideation
with plan, acute psychiatric crisis & --- \\
\texttt{substance\_intoxication\_\newline withdrawal} & Alcohol or drug
intoxication or withdrawal & --- \\
\texttt{substance\_overdose} & Intentional or unintentional overdose
requiring admission & ICD-10-CM: T40, T42, T43, T50; ICD-9-CM: 965, 967,
969, 977 \\
\texttt{symptom\_workup\_chest\_pain} & Chest pain admission for
rule-out, workup inconclusive or ruled out & --- \\
\texttt{symptom\_workup\_syncope} & Syncope admission for workup &
--- \\
\texttt{symptom\_workup\_other} & Other symptom-based admission for
workup without definitive diagnosis at discharge & --- \\
\texttt{elective\_procedure\_non\_oncology} & Planned non-oncology
procedure (elective surgery, cardiac cath, etc.) & --- \\
\texttt{obstetric} & Pregnancy, labor, postpartum complications & --- \\
\texttt{other} & Reason does not fit any of the above categories &
--- \\
\end{longtable}

\clearpage

\subsection{TriState Field
Definitions}\label{tristate-field-definitions}

\begin{longtable}[]{|>{\raggedright\arraybackslash}p{(\linewidth - 2\tabcolsep) * \real{0.3800}}|>{\raggedright\arraybackslash}p{(\linewidth - 2\tabcolsep) * \real{0.6200}}|}
\caption{\label{tbl:tristate-fields} TriState field identifiers and
definitions used for disagreement decomposition.}\tabularnewline\hline
\hline
\begin{minipage}[b]{\linewidth}\raggedright
Field ID
\end{minipage} & \begin{minipage}[b]{\linewidth}\raggedright
Definition
\end{minipage} \\
\hline
\endfirsthead
\hline
\begin{minipage}[b]{\linewidth}\raggedright
Field ID
\end{minipage} & \begin{minipage}[b]{\linewidth}\raggedright
Definition
\end{minipage} \\
\hline
\endhead
\hline
\endlastfoot
\texttt{shock\_present} & Any form of shock documented during admission
(cardiogenic, septic, hypovolemic, distributive). \\
\texttt{infection\_as\_trigger} & Infection identified as trigger or
precipitant for the admission event, even if not the primary reason. \\
\texttt{aki\_present} & Acute kidney injury present at any point during
admission. \\
\texttt{lives\_alone} & Patient lives alone at home per social
history. \\
\texttt{social\_support\_absent} & Explicit documentation of lack of
social support (isolated, no family, etc.). \\
\texttt{financial\_hardship} & Documented financial hardship, uninsured,
cost-related medication nonadherence. \\
\texttt{substance\_use\_active} & Active substance use (alcohol, illicit
drugs, tobacco excluded). Historical use without active = `no'. \\
\texttt{fall\_risk\_documented} & Fall risk explicitly documented. \\
\texttt{cognitive\_impairment} & Baseline cognitive impairment
(dementia, MCI) documented --- distinct from delirium. \\
\texttt{goals\_of\_care\_flag} & Goals-of-care discussion documented
during admission, including phrases like `comfort-focused', `discussed
prognosis', `family meeting re: goals'. \\
\texttt{palliative\_care\_consult} & Palliative care team formally
consulted during this admission. \\
\texttt{dnr\_dni\_documented} & DNR, DNI, or DNR/DNI code status
documented (not just discussed). \\
\texttt{home\_health\_ordered} & Home health services (nursing, PT/OT at
home) ordered at discharge. \\
\texttt{cardiac\_rehab\_referred} & Referral to cardiac rehabilitation
program at discharge. \\
\texttt{discharge\_delayed\_reason} & Discharge was delayed for
non-medical reasons (placement, insurance, social). `yes' only if
explicitly documented. \\
\texttt{hospital\_acquired\_complication} & Any hospital-acquired
complication documented: HAI, hospital-acquired AKI, hospital-acquired
delirium, fall, pressure ulcer, etc. \\
\texttt{unresolved\_diagnosis\_at\_discharge} & Language indicating the
diagnosis remained unclear or workup pending at discharge (`etiology
unclear', `workup pending', `to be followed up as outpatient'). \\
\end{longtable}

\clearpage

\subsection{Enum Fields and Value
Sets}\label{enum-fields-and-value-sets}

\begin{longtable}[]{|>{\raggedright\arraybackslash}p{(\linewidth - 4\tabcolsep) * \real{0.3000}}|>{\raggedright\arraybackslash}p{(\linewidth - 4\tabcolsep) * \real{0.1800}}|>{\raggedright\arraybackslash}p{(\linewidth - 4\tabcolsep) * \real{0.5200}}|}
\caption{\label{tbl:enum-fields} Enum field value sets and operational
definitions used for cross-variant confusion analysis.}\tabularnewline\hline
\hline
\begin{minipage}[b]{\linewidth}\raggedright
Field ID
\end{minipage} & \begin{minipage}[b]{\linewidth}\raggedright
Value set
\end{minipage} & \begin{minipage}[b]{\linewidth}\raggedright
Value definitions
\end{minipage} \\
\hline
\endfirsthead
\hline
\begin{minipage}[b]{\linewidth}\raggedright
Field ID
\end{minipage} & \begin{minipage}[b]{\linewidth}\raggedright
Value set
\end{minipage} & \begin{minipage}[b]{\linewidth}\raggedright
Value definitions
\end{minipage} \\
\hline
\endhead
\hline
\endlastfoot
\texttt{functional\_status} & \texttt{independent} & Performs activities
of daily living independently. \\
& \texttt{assisted} & Needs some assistance and/or assistive devices. \\
& \texttt{dependent} & Requires substantial or full assistance for daily
activities. \\
& \texttt{not\_documented} & No functional-status statement found in
note. \\
\texttt{mental\_status} & \texttt{intact} & Alert/oriented or documented
at baseline mental status. \\
& \texttt{mild\_impairment} & Mild cognitive/mental-status impairment
documented. \\
& \texttt{confused\_delirious} & Confusion, delirium, or marked
disorientation documented. \\
& \texttt{not\_documented} & No mental-status statement found in
note. \\
\texttt{discharge\_condition\_category} & \texttt{stable} & Discharge
condition documented as stable. \\
& \texttt{improved} & Discharge condition documented as improved. \\
& \texttt{unchanged} & Discharge condition documented as unchanged. \\
& \texttt{deteriorated} & Discharge condition documented as
deteriorated. \\
& \texttt{expired} & Patient died during admission. \\
& \texttt{not\_documented} & No discharge-condition category
documented. \\
\texttt{new\_meds\_started\_count} &
\texttt{integer\ \textgreater{}=\ 0} & Count of distinct medications
started during admission. \\
& \texttt{null} & Medication reconciliation is indeterminate in note. \\
\texttt{meds\_stopped\_count} & \texttt{integer\ \textgreater{}=\ 0} &
Count of distinct medications stopped during admission. \\
& \texttt{null} & Medication reconciliation is indeterminate in note. \\
\end{longtable}

\clearpage

\subsection{ICD-Based Labeling
Functions}\label{icd-based-labeling-functions}

\begin{longtable}[]{|>{\raggedright\arraybackslash}p{(\linewidth - 6\tabcolsep) * \real{0.2400}}|>{\raggedright\arraybackslash}p{(\linewidth - 6\tabcolsep) * \real{0.3000}}|>{\raggedright\arraybackslash}p{(\linewidth - 6\tabcolsep) * \real{0.1600}}|>{\raggedright\arraybackslash}p{(\linewidth - 6\tabcolsep) * \real{0.3000}}|}
\caption{\label{tbl:icd-lfs} ICD-based labeling functions used in the
weak-supervision ensemble, including target mapping, match position, and
anchor code families.}\tabularnewline\hline
\hline
\begin{minipage}[b]{\linewidth}\raggedright
LF Name
\end{minipage} & \begin{minipage}[b]{\linewidth}\raggedright
Target field
\end{minipage} & \begin{minipage}[b]{\linewidth}\raggedright
Match position
\end{minipage} & \begin{minipage}[b]{\linewidth}\raggedright
ICD codes/prefixes
\end{minipage} \\
\hline
\endfirsthead
\hline
\begin{minipage}[b]{\linewidth}\raggedright
LF Name
\end{minipage} & \begin{minipage}[b]{\linewidth}\raggedright
Target field
\end{minipage} & \begin{minipage}[b]{\linewidth}\raggedright
Match position
\end{minipage} & \begin{minipage}[b]{\linewidth}\raggedright
ICD codes/prefixes
\end{minipage} \\
\hline
\endhead
\hline
\endlastfoot
\texttt{icd\_aki\_primary} & \texttt{aki\_present} & Any & ICD-10-CM:
N17\newline ICD-9-CM: 584 \\
\texttt{icd\_hf\_admission} & \texttt{cardiac\_hf} & Primary &
ICD-10-CM: I50, I11.0, I13.0, I13.2\newline ICD-9-CM: 428, 402.01,
402.11, 402.91, 404.01, 404.03, 404.11, 404.13, 404.91, 404.93 \\
\texttt{icd\_acs\_admission} & \texttt{cardiac\_acs} & Primary &
ICD-10-CM: I21, I22\newline ICD-9-CM: 410, 411 \\
\texttt{icd\_stroke\_admission} & \texttt{neuro\_stroke\_tia} & Primary
& ICD-10-CM: I60, I61, I62, I63, I64, G45\newline ICD-9-CM: 430, 431,
432, 433, 434, 435, 436 \\
\texttt{icd\_sepsis\_admission} & \texttt{sepsis\_bacteremia} & Any &
ICD-10-CM: A40, A41, R65.2\newline ICD-9-CM: 038, 995.91, 995.92,
790.7 \\
\texttt{icd\_afib\_admission} & \texttt{cardiac\_arrhythmia} & Primary &
ICD-10-CM: I47, I48, I49\newline ICD-9-CM: 427 \\
\texttt{icd\_copd\_exacerbation} &
\texttt{respiratory\_copd\_asthma\_\newline exacerbation} & Primary &
ICD-10-CM: J44, J45, J46\newline ICD-9-CM: 491, 492, 493, 496 \\
\texttt{icd\_pneumonia\_admission} & \texttt{respiratory\_infection} &
Primary & ICD-10-CM: J12, J13, J14, J15, J16, J17, J18, J20, J21,
J22\newline ICD-9-CM: 480, 481, 482, 483, 484, 485, 486, 487 \\
\texttt{icd\_pe\_admission} & \texttt{respiratory\_pe\_dvt} & Primary &
ICD-10-CM: I26, I82\newline ICD-9-CM: 415.1, 451, 453 \\
\texttt{icd\_gi\_bleed\_admission} & \texttt{gi\_bleed} & Primary &
ICD-10-CM: K25.0, K25.2, K25.4, K25.6, K26.0, K26.2, K26.4, K26.6,
K27.0, K27.2, K27.4, K27.6, K28.0, K28.2, K28.4, K28.6, K29.01, K29.21,
K29.31, K29.41, K29.51, K29.61, K29.71, K29.81, K29.91, K62.5, K92.0,
K92.1, K92.2\newline ICD-9-CM: 530.82, 531.00, 531.01, 531.20, 531.21,
531.40, 531.41, 531.60, 531.61, 532, 533, 534, 535.01, 535.11, 535.21,
535.31, 535.41, 535.51, 535.61, 535.71, 569.3, 578 \\
\texttt{icd\_cirrhosis\_admission} &
\texttt{hepatic\_failure\_cirrhosis} & Any & ICD-10-CM: K70, K71, K72,
K73, K74, K76.6, K76.7\newline ICD-9-CM: 571, 572, 573 \\
\texttt{icd\_dka\_hhs\_admission} & \texttt{metabolic\_dka\_hhs} &
Primary & ICD-10-CM: E10.1, E11.0, E11.1, E13.0, E13.1\newline ICD-9-CM:
250.1, 250.2, 250.3 \\
\texttt{icd\_oncology\_treatment\_\newline admission} &
\texttt{oncology\_elective\_treatment} & Primary & ICD-10-CM: Z51.0,
Z51.1\newline ICD-9-CM: V58.0, V58.1 \\
\texttt{icd\_fracture\_admission} & \texttt{trauma\_fracture} & Primary
& ICD-10-CM: S02, S12, S22, S32, S42, S52, S62, S72, S82,
S92\newline ICD-9-CM: 800-829 \\
\texttt{icd\_overdose\_admission} & \texttt{substance\_overdose} &
Primary & ICD-10-CM: T40, T42, T43, T50\newline ICD-9-CM: 965, 967, 969,
977 \\
\end{longtable}

\clearpage

\subsection{Regex Labeling Functions}\label{regex-labeling-functions}

\begin{longtable}[]{|>{\raggedright\arraybackslash}p{(\linewidth - 4\tabcolsep) * \real{0.4000}}|>{\raggedright\arraybackslash}p{(\linewidth - 4\tabcolsep) * \real{0.3000}}|>{\raggedright\arraybackslash}p{(\linewidth - 4\tabcolsep) * \real{0.3000}}|}
\caption{\label{tbl:regex-lfs} Regex-based labeling functions used in
the weak-supervision ensemble, including target mapping and anchor
patterns.}\tabularnewline\hline
\hline
\begin{minipage}[b]{\linewidth}\raggedright
LF Name
\end{minipage} & \begin{minipage}[b]{\linewidth}\raggedright
Target field
\end{minipage} & \begin{minipage}[b]{\linewidth}\raggedright
Regex pattern(s)
\end{minipage} \\
\hline
\endfirsthead
\hline
\begin{minipage}[b]{\linewidth}\raggedright
LF Name
\end{minipage} & \begin{minipage}[b]{\linewidth}\raggedright
Target field
\end{minipage} & \begin{minipage}[b]{\linewidth}\raggedright
Regex pattern(s)
\end{minipage} \\
\hline
\endhead
\hline
\endlastfoot
\texttt{regex\_aki\_present\_yes} & \texttt{aki\_present} &
\texttt{\textbackslash{}bAKI\textbackslash{}b},
\texttt{\textbackslash{}bacute\ kidney\ injury\textbackslash{}b},
\texttt{\textbackslash{}bacute\ renal\ failure\textbackslash{}b},
\texttt{\textbackslash{}bARF\textbackslash{}b} \\
\texttt{regex\_cardiac\_rehab\_referred\_yes} &
\texttt{cardiac\_rehab\_referred} &
\texttt{\textbackslash{}bcardiac\ rehab(ilitation)?\textbackslash{}b} \\
\texttt{regex\_cognitive\_impairment\_yes} &
\texttt{cognitive\_impairment} &
\texttt{\textbackslash{}bdementia\textbackslash{}b},
\texttt{\textbackslash{}bAlzheimer(\textquotesingle{}s)?\textbackslash{}b},
\texttt{\textbackslash{}bMCI\textbackslash{}b},
\texttt{\textbackslash{}bmild\ cognitive\ impairment\textbackslash{}b},
\texttt{\textbackslash{}bbaseline\ (confusion\ OR\ dementia)\textbackslash{}b} \\
\texttt{regex\_dnr\_dni\_documented\_yes} &
\texttt{dnr\_dni\_documented} &
\texttt{\textbackslash{}bDNR\textbackslash{}s*/\textbackslash{}s*DNI\textbackslash{}b},
\texttt{\textbackslash{}bDo\ Not\ Resuscitate\textbackslash{}b},
\texttt{\textbackslash{}bcomfort\ measures\ only\textbackslash{}b} \\
\texttt{regex\_fall\_risk\_documented\_yes} &
\texttt{fall\_risk\_documented} &
\texttt{\textbackslash{}bfall\ risk\textbackslash{}b},
\texttt{\textbackslash{}bhigh\ fall\ risk\textbackslash{}b};
\texttt{compound(all\ of={[}\textquotesingle{}fell\ OR\ fall\textquotesingle{},\textquotesingle{}at\ home\ OR\ mechanical\ OR\ witnessed\ OR\ unwitnessed\textquotesingle{}{]},\ window\ chars=20)} \\
\texttt{regex\_goals\_of\_care\_flag\_yes} &
\texttt{goals\_of\_care\_flag} &
\texttt{\textbackslash{}bgoals{[}-\ {]}of{[}-\ {]}care\textbackslash{}b},
\texttt{\textbackslash{}bGOC\textbackslash{}b},
\texttt{\textbackslash{}bfamily\ meeting\textbackslash{}b},
\texttt{\textbackslash{}bcomfort{[}-\ {]}focused\textbackslash{}b},
\texttt{\textbackslash{}bhospice\textbackslash{}b} \\
\texttt{regex\_home\_health\_ordered\_yes} &
\texttt{home\_health\_ordered} &
\texttt{\textbackslash{}bvisiting\ nurse\textbackslash{}b},
\texttt{\textbackslash{}bvisiting\ nursing\textbackslash{}b},
\texttt{\textbackslash{}bVNA\textbackslash{}b} \\
\texttt{regex\_palliative\_care\_consult\_yes} &
\texttt{palliative\_care\_consult} &
\texttt{\textbackslash{}bpalliative\ care\textbackslash{}b} \\
\texttt{regex\_substance\_use\_active\_yes} &
\texttt{substance\_use\_active} &
\texttt{\textbackslash{}bIVDU\textbackslash{}b},
\texttt{\textbackslash{}bpolysubstance\textbackslash{}b},
\texttt{\textbackslash{}bactive\textbackslash{}s+(alcohol\ OR\ drug\ OR\ substance)\textbackslash{}s+(use\ OR\ abuse)\textbackslash{}b},
\texttt{\textbackslash{}bCIWA\textbackslash{}b},
\texttt{\textbackslash{}bopiate\ withdrawal\textbackslash{}b},
\texttt{\textbackslash{}balcohol\ withdrawal\textbackslash{}b};
\texttt{compound(all\ of={[}\textquotesingle{}alcohol\ OR\ ethanol\ OR\ etoh\textquotesingle{},\textquotesingle{}abuse\ OR\ use\ OR\ consum\ OR\ overdose\ OR\ intox\ OR\ depend\textquotesingle{}{]},\ window\ chars=40)};
\texttt{compound(all\ of={[}\textquotesingle{}drink(er\ OR\ ing\ OR\ s)?\textquotesingle{},\textquotesingle{}daily\ OR\ heavy\ OR\ continues\ OR\ currently\ OR\ active\textquotesingle{}{]},\ window\ chars=30)};
\texttt{compound(all\ of={[}\textquotesingle{}heroin\ OR\ cocaine\ OR\ fentanyl\ OR\ meth\ OR\ oxycodone\ OR\ opioid\ OR\ opiate\textquotesingle{},\textquotesingle{}use\ OR\ abuse\ OR\ dependence\ OR\ active\ OR\ current\textquotesingle{}{]},\ window\ chars=40)} \\
\end{longtable}

\clearpage

\section{Sample-Size Stability}\label{sample-size-stability}

\begin{figure}
\centering
\pandocbounded{\includegraphics[keepaspectratio,alt={Filtered-median cross-prompt agreement as a function of sample size, with 95\% bootstrap confidence intervals. Five points report \textbackslash bar\{\textbackslash kappa\}\^{}\{\textbackslash mathrm\{TriState\}\} on the refinement set (n = 150, \textbackslash bar\{\textbackslash kappa\} = 0.70), holdout set (n = 150, \textbackslash bar\{\textbackslash kappa\} = 0.70), validation set (n = 1,000, \textbackslash bar\{\textbackslash kappa\} = 0.66), audit subset (n = 500, \textbackslash bar\{\textbackslash kappa\} = 0.67), and extended validation set (n = 5,000, \textbackslash bar\{\textbackslash kappa\} = 0.65). Confidence intervals overlap across all five sample sizes.}]{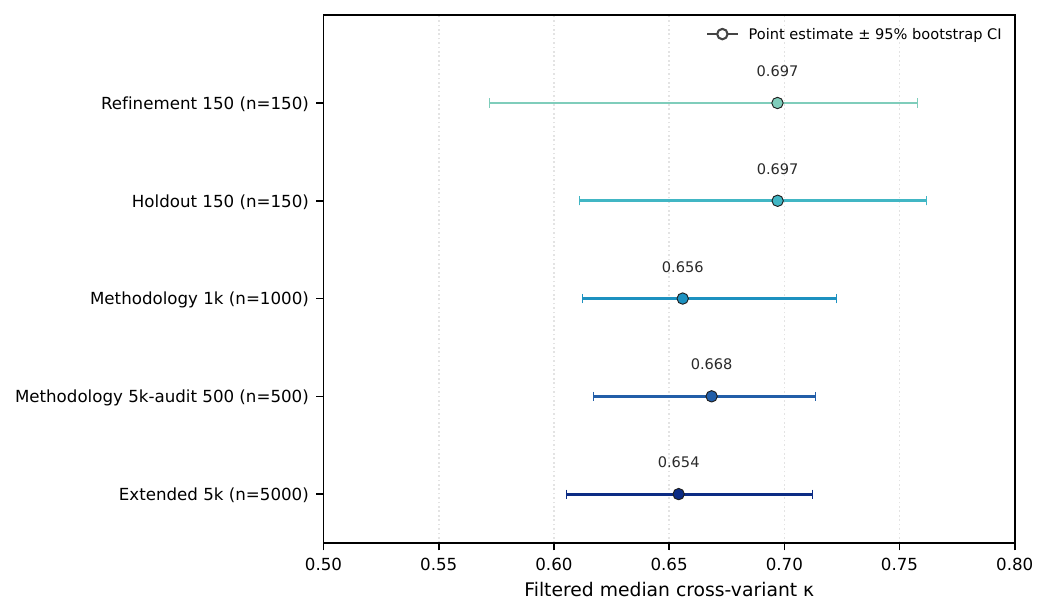}}
\caption{\textbf{Filtered-median cross-prompt agreement as a function of
sample size, with 95\% bootstrap confidence intervals.} Five points
report \(\bar{\kappa}^{\mathrm{TriState}}\) on the refinement set
(\(n = 150\), \(\bar{\kappa} = 0.70\)), holdout set (\(n = 150\),
\(\bar{\kappa} = 0.70\)), validation set (\(n = 1,000\),
\(\bar{\kappa} = 0.66\)), audit subset (\(n = 500\),
\(\bar{\kappa} = 0.67\)), and extended validation set (\(n = 5,000\),
\(\bar{\kappa} = 0.65\)). Confidence intervals overlap across all five
sample sizes.}\label{fig:sample-size-stability}
\end{figure}

This section documents that the cross-prompt agreement findings reported
throughout the paper are stable across the sample sizes used for
development and validation
(Section\nobreakspace{}\ref{sec-results-stability}).
Figure\nobreakspace{}\ref{fig:sample-size-stability} reports
filtered-median \(\bar{\kappa}^{\text{TriState}}\) with 95\% bootstrap
confidence intervals at five sample sizes spanning two orders of
magnitude, from a 150-note refinement set to a 5,000-note extended
validation set. Confidence intervals overlap across sample sizes,
indicating that the agreement metric has converged sufficiently that
subsequent comparisons reflect substantive findings rather than
sample-size artifacts.

\clearpage

\section{Refinement-to-Holdout
Generalization}\label{refinement-to-holdout-generalization}

This section reports the agreement metrics on the refinement set used
during the autonomous optimization loop
(Section\nobreakspace{}\ref{sec-optimization-loop}) and on the
firewalled holdout set evaluated once after the loop completed.
Per-variant median \(\kappa\) values are reported on four small-model
samples to confirm that the optimization loop did not produce a
noticeable refinement-to-holdout gap: the 150-note refinement set, the
150-note holdout set, the 1,000-note validation set, and the 5,000-note
extended validation set. The refinement-to-holdout differences are small
for all three variants, supporting the use of the post-optimization
prompts at population scale.

\begin{figure}
\centering
\pandocbounded{\includegraphics[keepaspectratio,alt={Filtered-median cross-prompt agreement on four small-model samples, reported per prompt variant. For each variant (A, B, C), the median \textbackslash kappa over TriState fields is reported on four samples: the 150-note refinement set used during the optimization loop, the 150-note holdout set evaluated once after the loop completed, the 1,000-note validation set, and the 5,000-note extended validation set. The refinement-to-holdout difference is small for all three variants, consistent with the optimization loop not producing a noticeable refinement-to-holdout gap. The slight decrease in median \textbackslash kappa from the 150-note samples to the 1,000-note and 5,000-note samples is consistent with convergence to a stable population-level estimate as sample size increases.}]{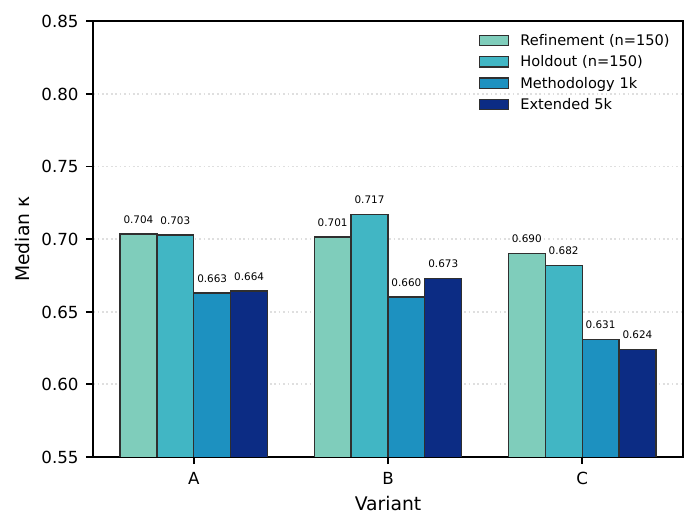}}
\caption{\textbf{Filtered-median cross-prompt agreement on four
small-model samples, reported per prompt variant.} For each variant (A,
B, C), the median \(\kappa\) over TriState fields is reported on four
samples: the 150-note refinement set used during the optimization loop,
the 150-note holdout set evaluated once after the loop completed, the
1,000-note validation set, and the 5,000-note extended validation set.
The refinement-to-holdout difference is small for all three variants,
consistent with the optimization loop not producing a noticeable
refinement-to-holdout gap. The slight decrease in median \(\kappa\) from
the 150-note samples to the 1,000-note and 5,000-note samples is
consistent with convergence to a stable population-level estimate as
sample size increases.}\label{fig:refinement-holdout}
\end{figure}

\clearpage

\section{Per-Variant Cross-Model
Agreement}\label{per-variant-cross-model-agreement}

This section reports the same-prompt small-model-versus-full-model
agreement on TriState clinical documentation flags for each prompt
variant separately, complementing the pooled cross-prompt agreement
summary in Figure\nobreakspace{}\ref{fig:cross-prompt-model-size} of the
main text. The figure quantifies how much the extraction changes when
only the model size is varied, holding the prompt fixed. The same-prompt
cross-model agreement is substantially lower than the cross-variant
agreement at either model size, supporting the interpretation in
Section\nobreakspace{}\ref{sec-discussion-model-size-vs-prompt} that
model size is a larger driver of extraction-output variability than
prompt phrasing on the same notes.

\begin{figure}
\centering
\pandocbounded{\includegraphics[keepaspectratio,alt={Cross-model agreement on TriState fields, computed per prompt variant, under both the TriState schema and binary collapse. For each variant (A, B, C), \textbackslash bar\{\textbackslash kappa\} measures the agreement between the small-model and full-model extractions on the same notes under the same prompt, filtered-median over the included TriState documentation fields under the paired base-rate rule (Section, Section). Per-variant values are \textbackslash bar\{\textbackslash kappa\}\^{}\{\textbackslash mathrm\{TriState\}\} = 0.32 (A), 0.39 (B), 0.43 (C) and \textbackslash bar\{\textbackslash kappa\}\^{}\{\textbackslash mathrm\{collapsed\}\} = 0.70 (A), 0.69 (B), 0.68 (C). The level of agreement between two model sizes on the same prompt is substantially lower than the cross-variant agreement at the small model (Figure). Paired sample, n = 1,500.}]{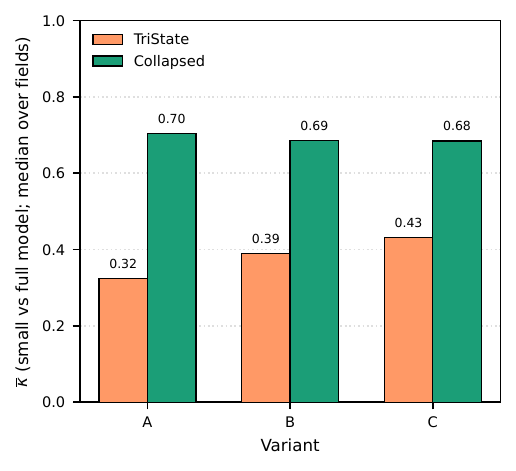}}
\caption{\textbf{Cross-model agreement on TriState fields, computed per
prompt variant, under both the TriState schema and binary collapse.} For
each variant (A, B, C), \(\bar{\kappa}\) measures the agreement between
the small-model and full-model extractions on the same notes under the
same prompt, filtered-median over the included TriState documentation
fields under the paired base-rate rule
(Section\nobreakspace{}\ref{sec-cross-prompt-metrics},
Section\nobreakspace{}\ref{sec-model-size-comparison}). Per-variant
values are \(\bar{\kappa}^{\mathrm{TriState}} = 0.32\) (A), 0.39 (B),
0.43 (C) and \(\bar{\kappa}^{\mathrm{collapsed}} = 0.70\) (A), 0.69 (B),
0.68 (C). The level of agreement between two model sizes on the same
prompt is substantially lower than the cross-variant agreement at the
small model (Figure\nobreakspace{}\ref{fig:cross-prompt-model-size}).
Paired sample, \(n = 1,500\).}\label{fig:per-variant-cross-model}
\end{figure}

\clearpage

\section{TriState Disagreement
Decomposition}\label{tristate-disagreement-decomposition}

\begin{figure}
\centering
\pandocbounded{\includegraphics[keepaspectratio,alt={Decomposition of cross-variant disagreement on TriState fields by disagreement type. Stacked bars show the share of disagreements occurring on the soft no-vs-not\_documented axis versus the hard yes-vs-not-yes axis, on the pooled small-model cross-variant sample. Soft disagreement totals 98.1\% of disagreement cases; hard yes-vs-no flips total 1.9\%. Binary collapse removes the soft component, dissolving 68.6\% of the disagreement count. Pooled cross-variant sample, n = 6,500.}]{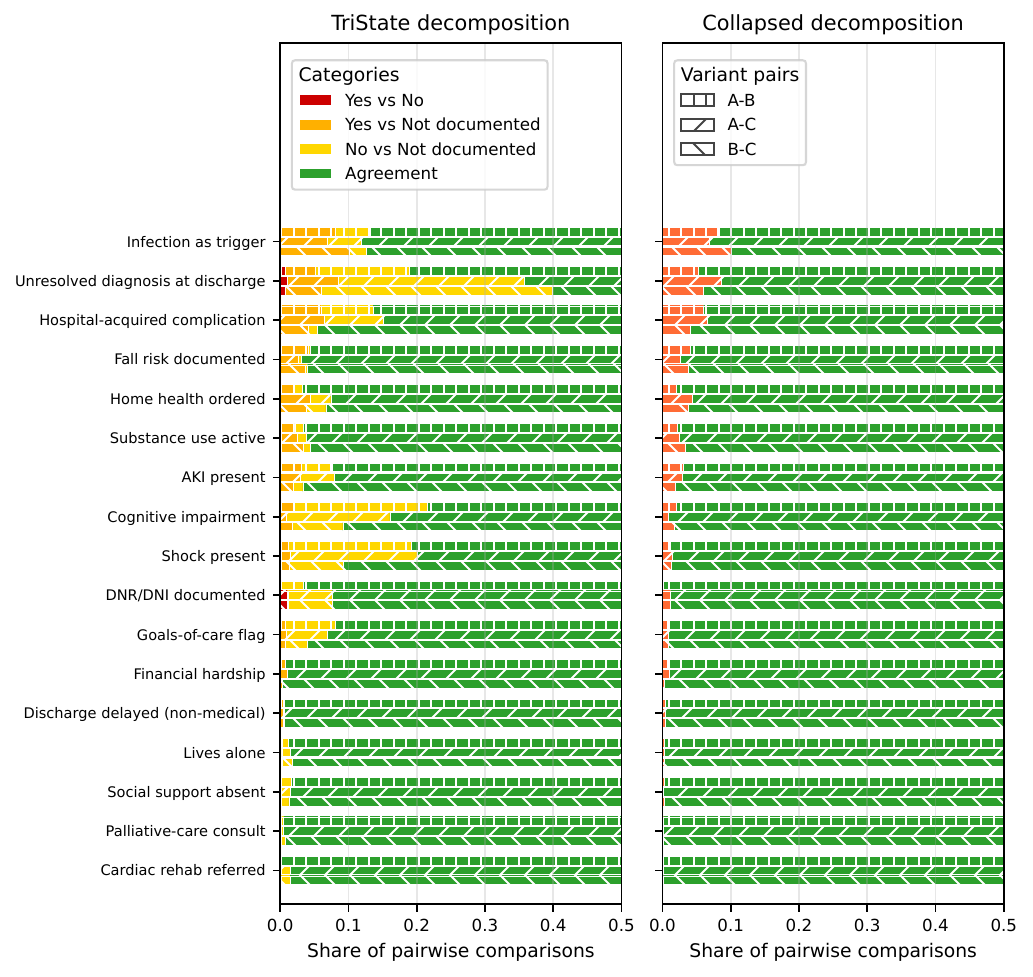}}
\caption{\textbf{Decomposition of cross-variant disagreement on TriState
fields by disagreement type.} Stacked bars show the share of
disagreements occurring on the soft
\texttt{no}-vs-\texttt{not\_documented} axis versus the hard
\texttt{yes}-vs-not-yes axis, on the pooled small-model cross-variant
sample. Soft disagreement totals 98.1\% of disagreement cases; hard
yes-vs-no flips total 1.9\%. Binary collapse removes the soft component,
dissolving 68.6\% of the disagreement count. Pooled cross-variant
sample, \(n = 6,500\).}\label{fig:disagreement-decomposition}
\end{figure}

This section accompanies
Section\nobreakspace{}\ref{sec-results-soft-disagreement}, decomposing
cross-variant disagreement on the 17 TriState clinical documentation
flags into the soft \texttt{no}-versus-\texttt{not\_documented}
component and the hard \texttt{yes}-versus-not-yes component.
Table\nobreakspace{}\ref{tbl:tristate-fields} lists the 17 TriState
fields with their definitions used throughout the schema.
Figure\nobreakspace{}\ref{fig:disagreement-decomposition} reports the
share of disagreement cases attributable to each component on the pooled
small-model cross-variant sample (\(n = 6,500\) shared notes), and
quantifies the fraction of disagreement that dissolves under post-hoc
binary collapse (Section\nobreakspace{}\ref{sec-binary-collapse}).

\subsection{Disagreement
Decomposition}\label{disagreement-decomposition}

\clearpage

\section{Admission-Tag Full Vocabulary
Confusion}\label{admission-tag-full-vocabulary-confusion}

\begin{figure}
\centering
\pandocbounded{\includegraphics[keepaspectratio,alt={Full-vocabulary cross-variant confusion matrix on the primary admission reason for variant pair A-B, 47x47. This extends Figure to the complete 47-tag vocabulary. Color encodes row-normalized confusion rate on log scale. Diagonal cells are the per-tag agreement rate within the pair; off-diagonal mass indicates phrasing-dependent categorization differences. Diagonal mass is 74.1\%. Pooled cross-variant sample, n = 6,500.}]{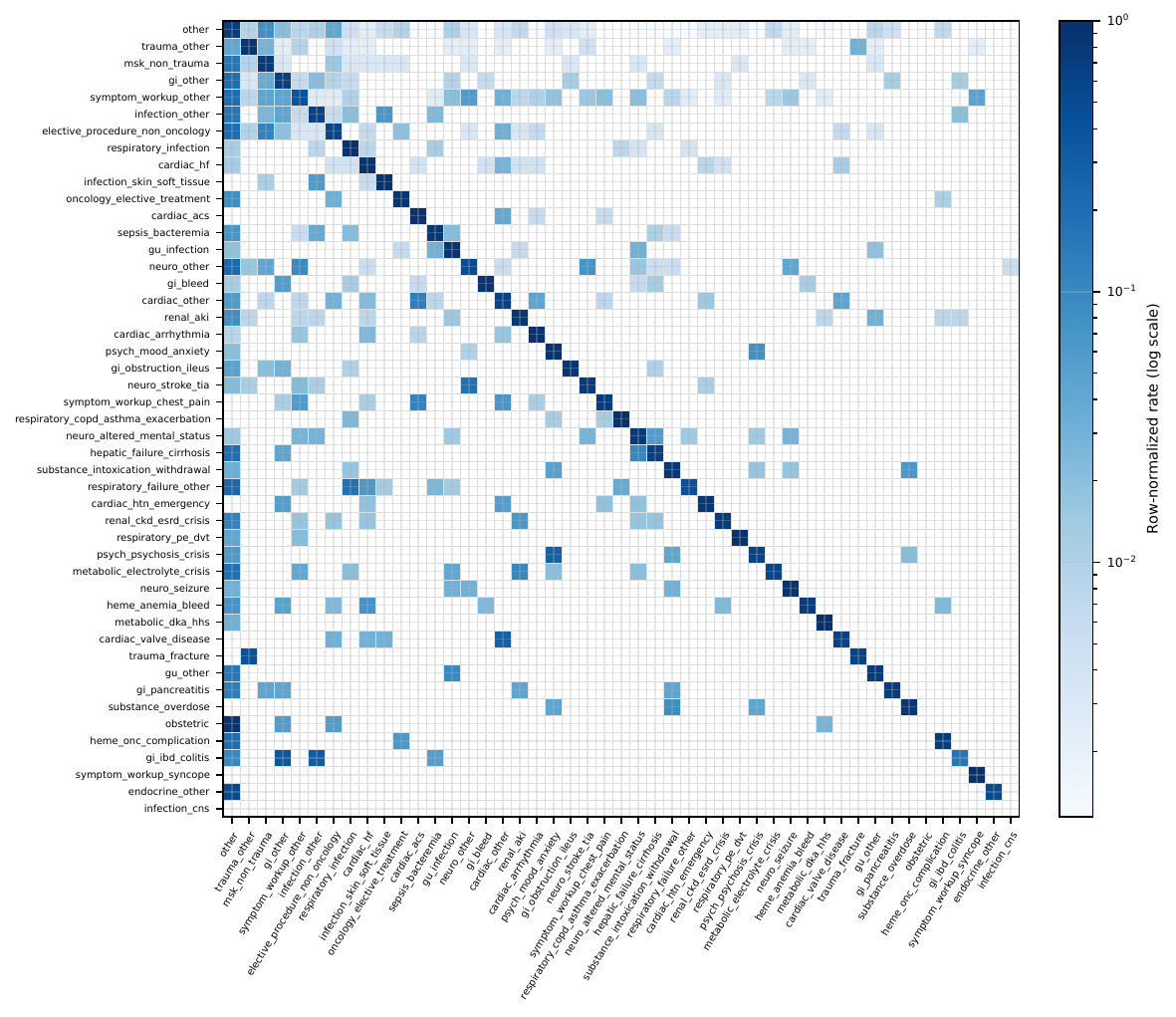}}
\caption{\textbf{Full-vocabulary cross-variant confusion matrix on the
primary admission reason for variant pair A-B, 47x47.} This extends
Figure\nobreakspace{}\ref{fig:admission-confusion} to the complete
47-tag vocabulary. Color encodes row-normalized confusion rate on log
scale. Diagonal cells are the per-tag agreement rate within the pair;
off-diagonal mass indicates phrasing-dependent categorization
differences. Diagonal mass is 74.1\%. Pooled cross-variant sample,
\(n = 6,500\).}\label{fig:cross-variant-ab}
\end{figure}

\begin{figure}
\centering
\pandocbounded{\includegraphics[keepaspectratio,alt={Full-vocabulary cross-variant confusion matrix on the primary admission reason for variant pair A-C, 47x47. This extends Figure to the complete 47-tag vocabulary. Color encodes row-normalized confusion rate on log scale. Diagonal cells are the per-tag agreement rate within the pair; off-diagonal mass indicates phrasing-dependent categorization differences. Diagonal mass is 69.9\%. Pooled cross-variant sample, n = 6,500.}]{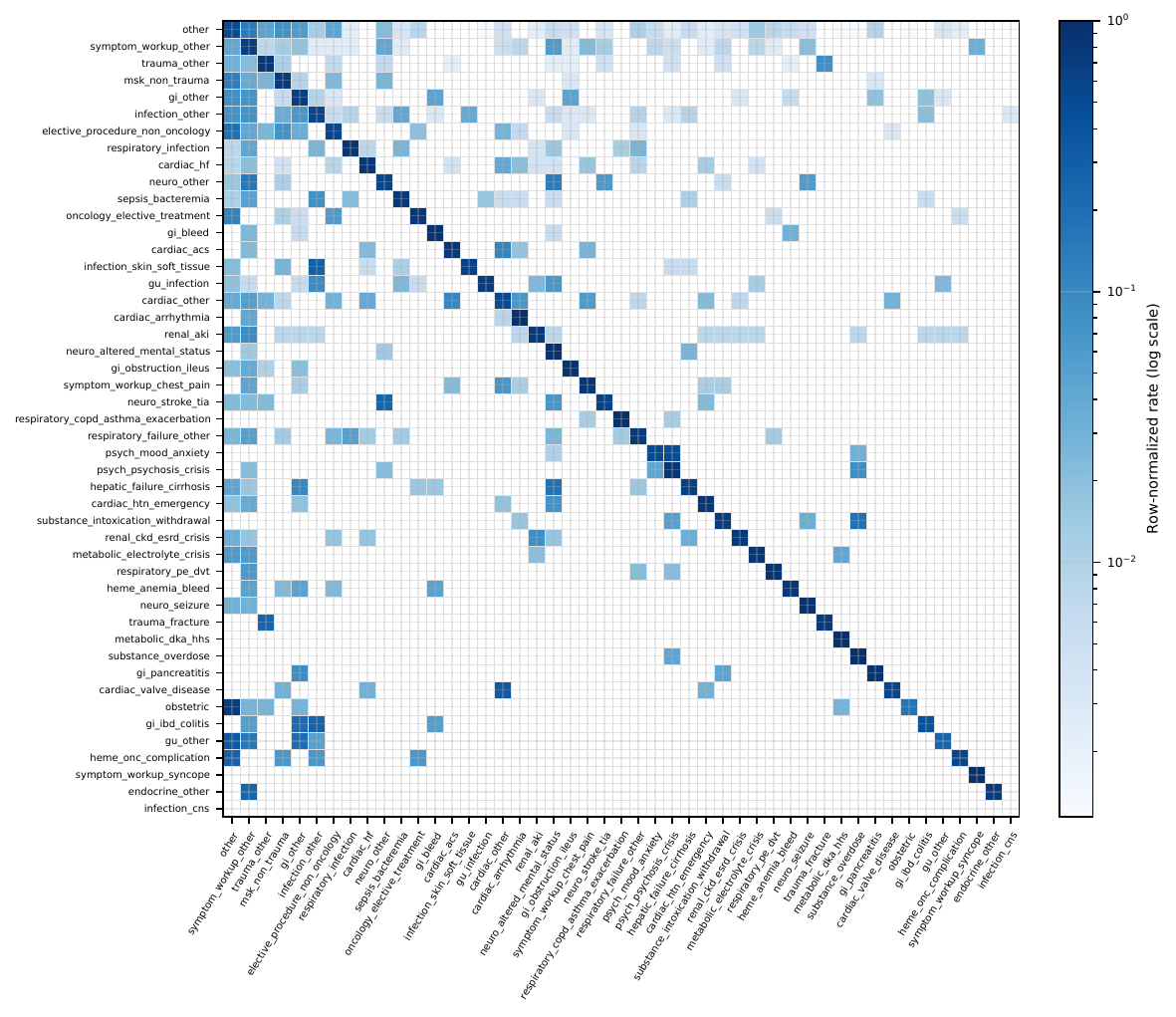}}
\caption{\textbf{Full-vocabulary cross-variant confusion matrix on the
primary admission reason for variant pair A-C, 47x47.} This extends
Figure\nobreakspace{}\ref{fig:admission-confusion} to the complete
47-tag vocabulary. Color encodes row-normalized confusion rate on log
scale. Diagonal cells are the per-tag agreement rate within the pair;
off-diagonal mass indicates phrasing-dependent categorization
differences. Diagonal mass is 69.9\%. Pooled cross-variant sample,
\(n = 6,500\).}\label{fig:cross-variant-ac}
\end{figure}

\begin{figure}
\centering
\pandocbounded{\includegraphics[keepaspectratio,alt={Full-vocabulary cross-variant confusion matrix on the primary admission reason for variant pair B-C, 47x47. This extends Figure to the complete 47-tag vocabulary. Color encodes row-normalized confusion rate on log scale. Diagonal cells are the per-tag agreement rate within the pair; off-diagonal mass indicates phrasing-dependent categorization differences. Diagonal mass is 67.8\%. Pooled cross-variant sample, n = 6,500.}]{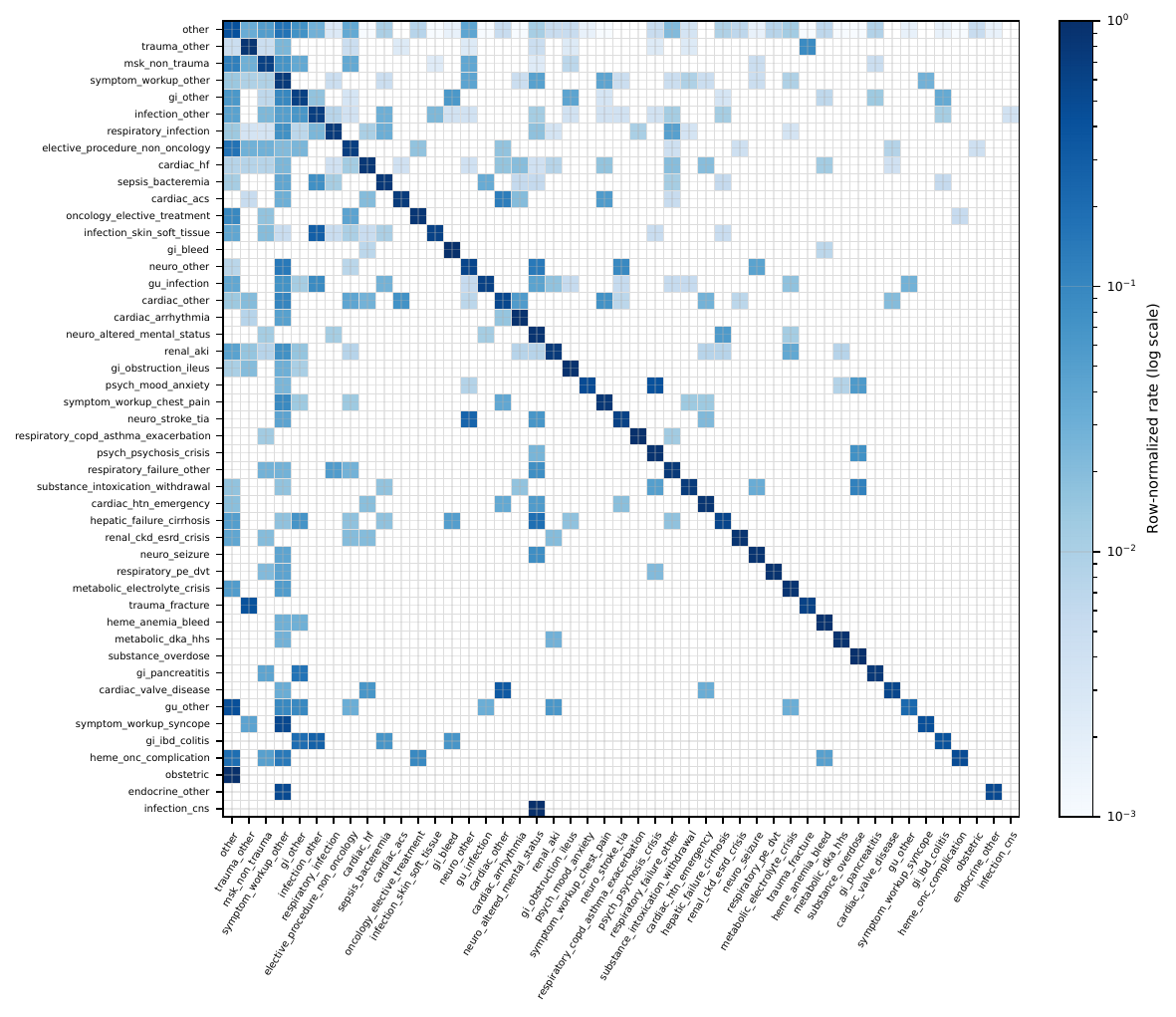}}
\caption{\textbf{Full-vocabulary cross-variant confusion matrix on the
primary admission reason for variant pair B-C, 47x47.} This extends
Figure\nobreakspace{}\ref{fig:admission-confusion} to the complete
47-tag vocabulary. Color encodes row-normalized confusion rate on log
scale. Diagonal cells are the per-tag agreement rate within the pair;
off-diagonal mass indicates phrasing-dependent categorization
differences. Diagonal mass is 67.8\%. Pooled cross-variant sample,
\(n = 6,500\).}\label{fig:cross-variant-bc}
\end{figure}

\begin{figure}
\centering
\pandocbounded{\includegraphics[keepaspectratio,alt={Full-vocabulary same-prompt cross-model confusion matrix on the primary admission reason for variant A, 47x47. The matrix compares the small-model and full-model extractions on the same notes under the same prompt. Color encodes row-normalized confusion rate on log scale. Diagonal mass is 54.3\%, quantifying same-prompt agreement across model sizes for variant A. Paired sample, n = 1,500.}]{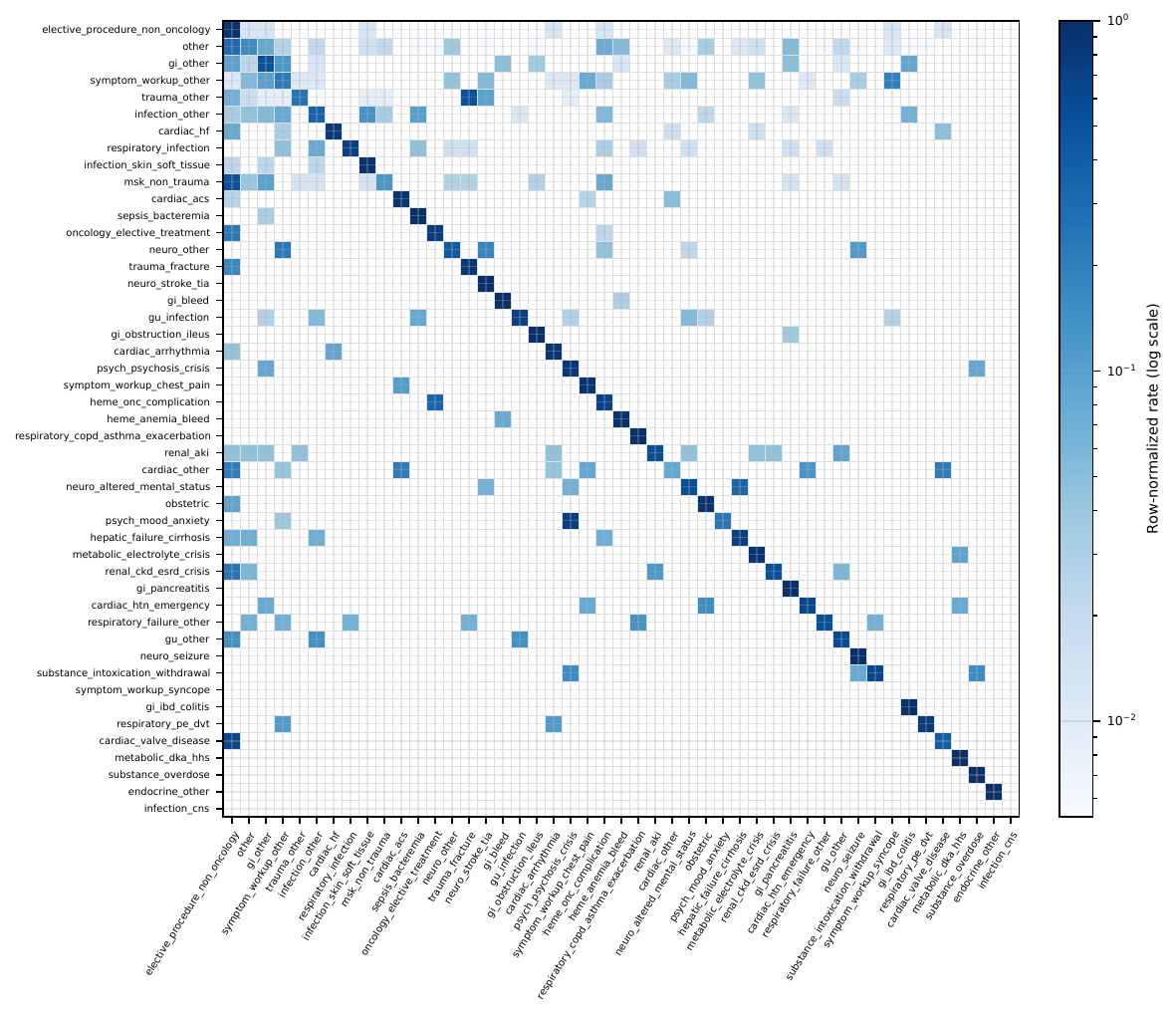}}
\caption{\textbf{Full-vocabulary same-prompt cross-model confusion
matrix on the primary admission reason for variant A, 47x47.} The matrix
compares the small-model and full-model extractions on the same notes
under the same prompt. Color encodes row-normalized confusion rate on
log scale. Diagonal mass is 54.3\%, quantifying same-prompt agreement
across model sizes for variant A. Paired sample,
\(n = 1,500\).}\label{fig:cross-model-a}
\end{figure}

\begin{figure}
\centering
\pandocbounded{\includegraphics[keepaspectratio,alt={Full-vocabulary same-prompt cross-model confusion matrix on the primary admission reason for variant B, 47x47. The matrix compares the small-model and full-model extractions on the same notes under the same prompt. Color encodes row-normalized confusion rate on log scale. Diagonal mass is 52.2\%, quantifying same-prompt agreement across model sizes for variant B. Paired sample, n = 1,500.}]{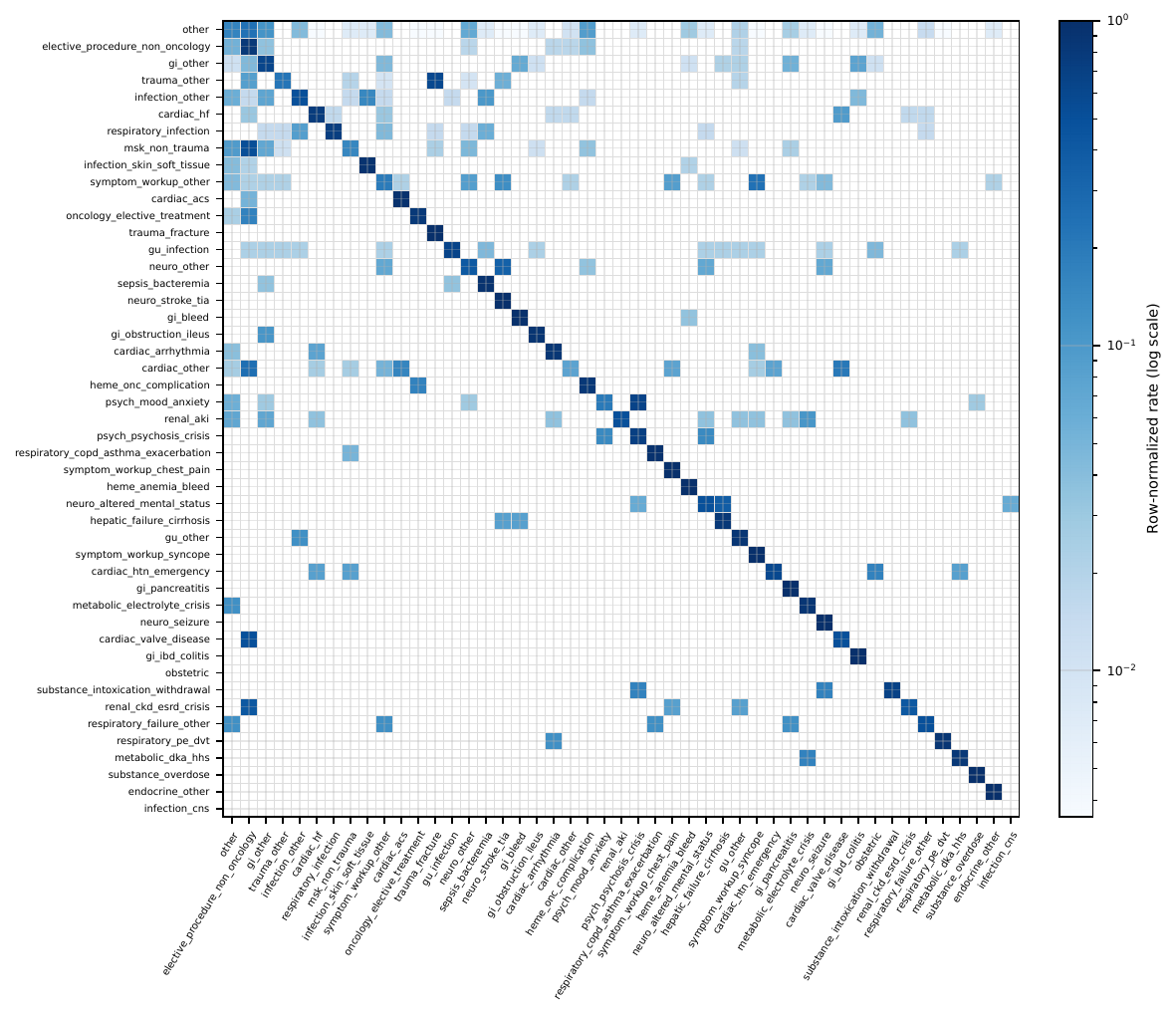}}
\caption{\textbf{Full-vocabulary same-prompt cross-model confusion
matrix on the primary admission reason for variant B, 47x47.} The matrix
compares the small-model and full-model extractions on the same notes
under the same prompt. Color encodes row-normalized confusion rate on
log scale. Diagonal mass is 52.2\%, quantifying same-prompt agreement
across model sizes for variant B. Paired sample,
\(n = 1,500\).}\label{fig:cross-model-b}
\end{figure}

\begin{figure}
\centering
\pandocbounded{\includegraphics[keepaspectratio,alt={Full-vocabulary same-prompt cross-model confusion matrix on the primary admission reason for variant C, 47x47. The matrix compares the small-model and full-model extractions on the same notes under the same prompt. Color encodes row-normalized confusion rate on log scale. Diagonal mass is 52.7\%, quantifying same-prompt agreement across model sizes for variant C. Paired sample, n = 1,500.}]{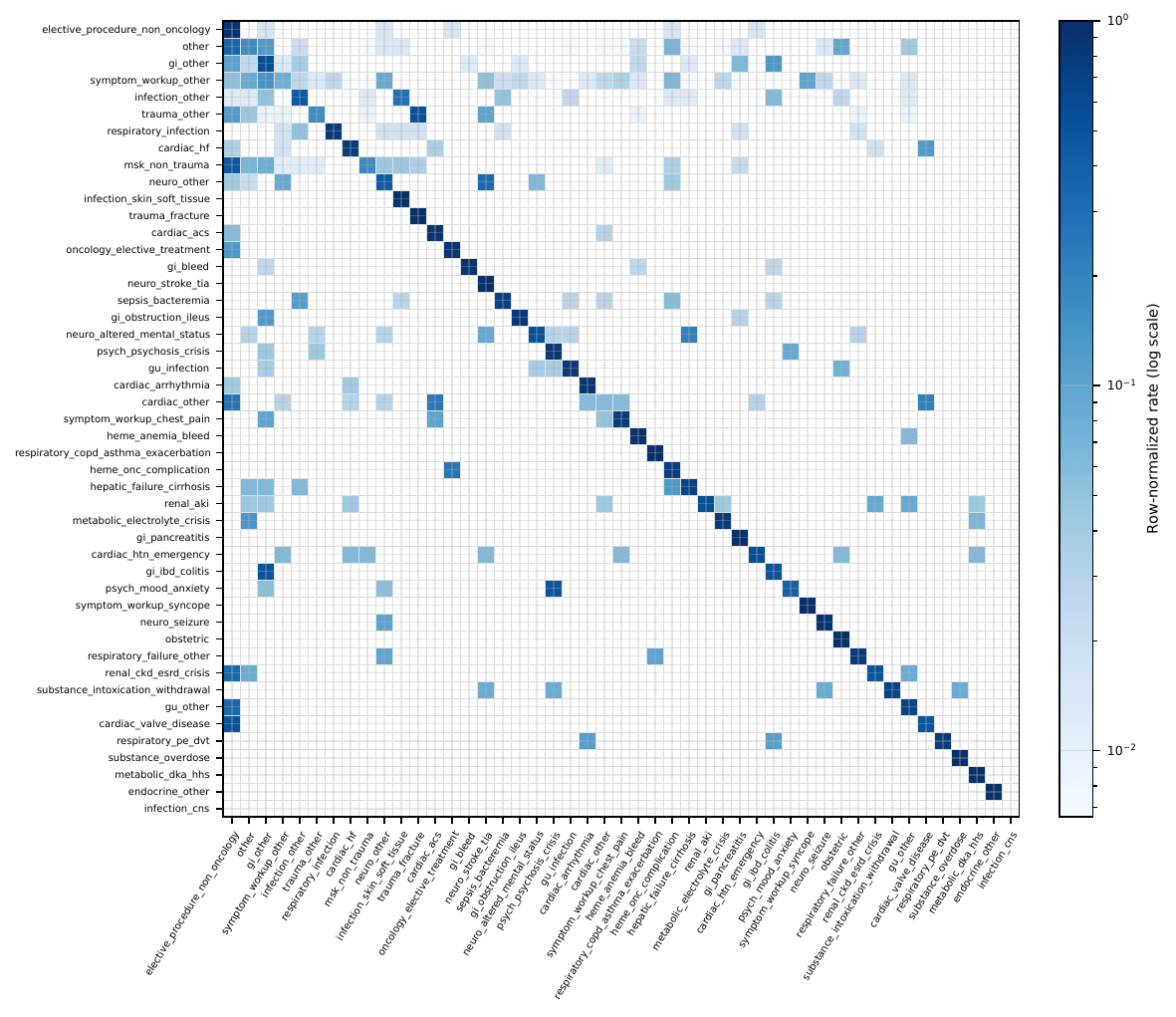}}
\caption{\textbf{Full-vocabulary same-prompt cross-model confusion
matrix on the primary admission reason for variant C, 47x47.} The matrix
compares the small-model and full-model extractions on the same notes
under the same prompt. Color encodes row-normalized confusion rate on
log scale. Diagonal mass is 52.7\%, quantifying same-prompt agreement
across model sizes for variant C. Paired sample,
\(n = 1,500\).}\label{fig:cross-model-c}
\end{figure}

This section extends Figure\nobreakspace{}\ref{fig:admission-confusion}
of the main text by reporting the complete 47x47 confusion matrices on
the primary admission reason.
Table\nobreakspace{}\ref{tbl:admission-tags} lists the 47
admission-reason tags with their definitions and ICD-10-CM anchor codes
(where defined). Figure\nobreakspace{}\ref{fig:cross-variant-ab},
Figure\nobreakspace{}\ref{fig:cross-variant-ac}, and
Figure\nobreakspace{}\ref{fig:cross-variant-bc} report the cross-variant
confusion structure at the small model for variant pairs A-B, A-C, and
B-C respectively, on the pooled \(n = 6,500\)-note cross-variant sample.
Figure\nobreakspace{}\ref{fig:cross-model-a},
Figure\nobreakspace{}\ref{fig:cross-model-b}, and
Figure\nobreakspace{}\ref{fig:cross-model-c} report the same-prompt
cross-model confusion (small-versus-full on the same prompt) for
variants A, B, and C respectively, on the 1,500-note paired sample. The
same-prompt cross-model panels make visible the model-size effect on
dominant-tag selection discussed in
Section\nobreakspace{}\ref{sec-results-model-size} and
Section\nobreakspace{}\ref{sec-discussion-model-size-vs-prompt}.

\subsection{Cross-Variant Confusion:
A-B}\label{cross-variant-confusion-a-b}

\subsection{Cross-Variant Confusion:
A-C}\label{cross-variant-confusion-a-c}

\subsection{Cross-Variant Confusion:
B-C}\label{cross-variant-confusion-b-c}

\subsection{Cross-Model Confusion: Variant
A}\label{cross-model-confusion-variant-a}

\subsection{Cross-Model Confusion: Variant
B}\label{cross-model-confusion-variant-b}

\subsection{Cross-Model Confusion: Variant
C}\label{cross-model-confusion-variant-c}

\clearpage

\section{Enum Field Confusion}\label{enum-field-confusion}

\begin{figure}
\centering
\pandocbounded{\includegraphics[keepaspectratio,alt={Cross-variant confusion matrices on mental status. Three sub-matrices for variant pairs A-B, A-C, and B-C. Diagonal mass is 96.7\% (A-B), 95.9\% (A-C), and 95.2\% (B-C). Notable off-diagonal mass between mild\_impairment and confused\_delirious, and between not\_documented and intact. Pooled cross-variant sample, n = 6,500.}]{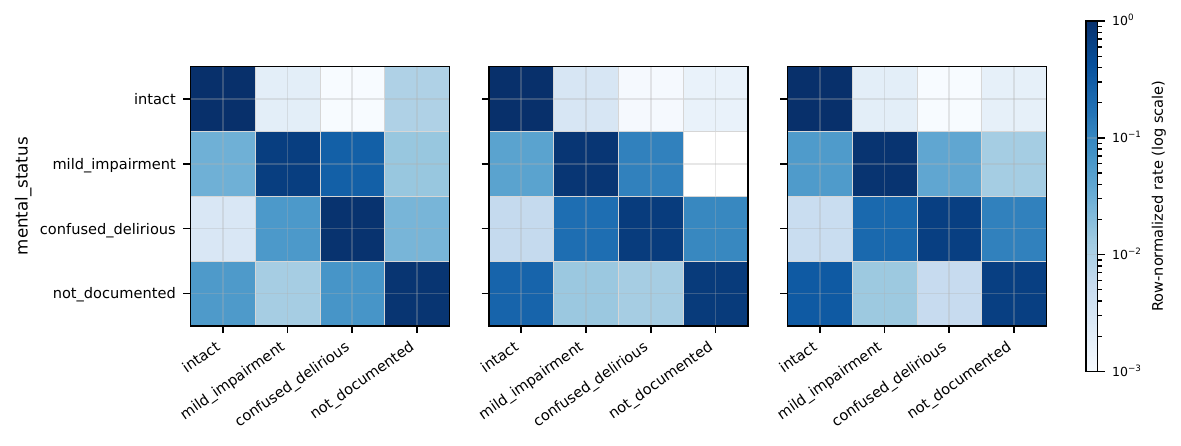}}
\caption{\textbf{Cross-variant confusion matrices on mental status.}
Three sub-matrices for variant pairs A-B, A-C, and B-C. Diagonal mass is
96.7\% (A-B), 95.9\% (A-C), and 95.2\% (B-C). Notable off-diagonal mass
between \texttt{mild\_impairment} and \texttt{confused\_delirious}, and
between \texttt{not\_documented} and \texttt{intact}. Pooled
cross-variant sample, \(n = 6,500\).}\label{fig:enum-mental-status}
\end{figure}

\begin{figure}
\centering
\pandocbounded{\includegraphics[keepaspectratio,alt={Cross-variant confusion matrices on functional status at discharge. Three sub-matrices for variant pairs A-B, A-C, and B-C. Diagonal mass is 96.4\% (A-B), 96.3\% (A-C), and 96.9\% (B-C). The boundary between dependent and assisted is the only off-diagonal cluster at appreciable scale. Pooled cross-variant sample, n = 6,500.}]{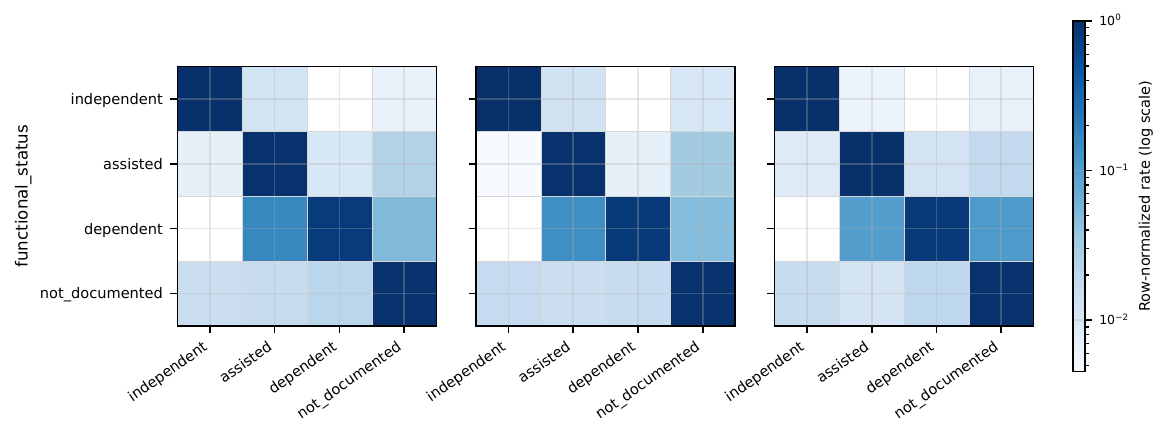}}
\caption{\textbf{Cross-variant confusion matrices on functional status
at discharge.} Three sub-matrices for variant pairs A-B, A-C, and B-C.
Diagonal mass is 96.4\% (A-B), 96.3\% (A-C), and 96.9\% (B-C). The
boundary between \texttt{dependent} and \texttt{assisted} is the only
off-diagonal cluster at appreciable scale. Pooled cross-variant sample,
\(n = 6,500\).}\label{fig:enum-functional-status}
\end{figure}

\begin{figure}
\centering
\pandocbounded{\includegraphics[keepaspectratio,alt={Cross-variant confusion matrices on discharge condition category. Three sub-matrices for variant pairs A-B, A-C, and B-C. Diagonal mass is 71.5\% (A-B), 69.7\% (A-C), and 84.1\% (B-C). Off-diagonal mass concentrates between unchanged and the more clinically specific categories. Pooled cross-variant sample, n = 6,500.}]{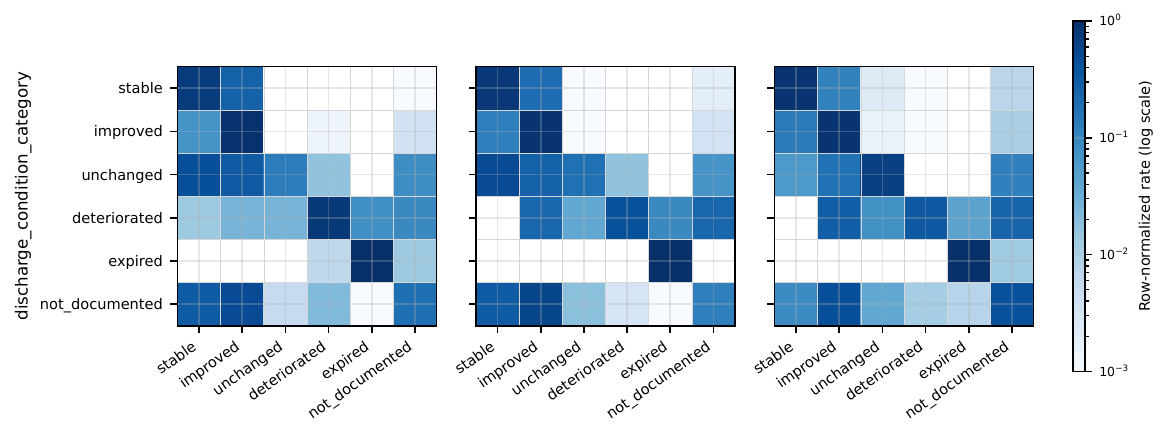}}
\caption{\textbf{Cross-variant confusion matrices on discharge condition
category.} Three sub-matrices for variant pairs A-B, A-C, and B-C.
Diagonal mass is 71.5\% (A-B), 69.7\% (A-C), and 84.1\% (B-C).
Off-diagonal mass concentrates between \texttt{unchanged} and the more
clinically specific categories. Pooled cross-variant sample,
\(n = 6,500\).}\label{fig:enum-discharge-condition}
\end{figure}

This section reports the cross-variant confusion matrices on the three
enumerated-value-set fields described in
Section\nobreakspace{}\ref{sec-results-enum}: mental status, functional
status at discharge, and discharge condition category.
Table\nobreakspace{}\ref{tbl:enum-fields} lists the value sets and
per-value definitions for each enum field, and the two count fields
included in the schema for completeness.
Figure\nobreakspace{}\ref{fig:enum-mental-status},
Figure\nobreakspace{}\ref{fig:enum-functional-status}, and
Figure\nobreakspace{}\ref{fig:enum-discharge-condition} report the
variant-pair confusion structure on the pooled \(n = 6,500\)-note
cross-variant sample for each enum field. Functional status at discharge
is the most consistent enum field across variants; discharge condition
category shows the largest residual variant-pair differences.

\subsection{Mental Status Confusion}\label{mental-status-confusion}

\subsection{Functional Status
Confusion}\label{functional-status-confusion}

\subsection{Discharge Condition
Confusion}\label{discharge-condition-confusion}

\clearpage

\section{Labeling Function
Complementarity}\label{labeling-function-complementarity}

\begin{figure}
\centering
\pandocbounded{\includegraphics[keepaspectratio,alt={Cross-tabulation of LLM-positive and ICD-labeling-function-positive on 15 ICD-anchored targets. For each target, the figure reports prevalence of LLM-positive across variants A, B, and C alongside the ICD-anchor prevalence and the variant-vs-ICD Cohen's kappa. The relationship between LLM and ICD signals is not uniform across targets: LLM and ICD capture overlapping but distinct content per target.}]{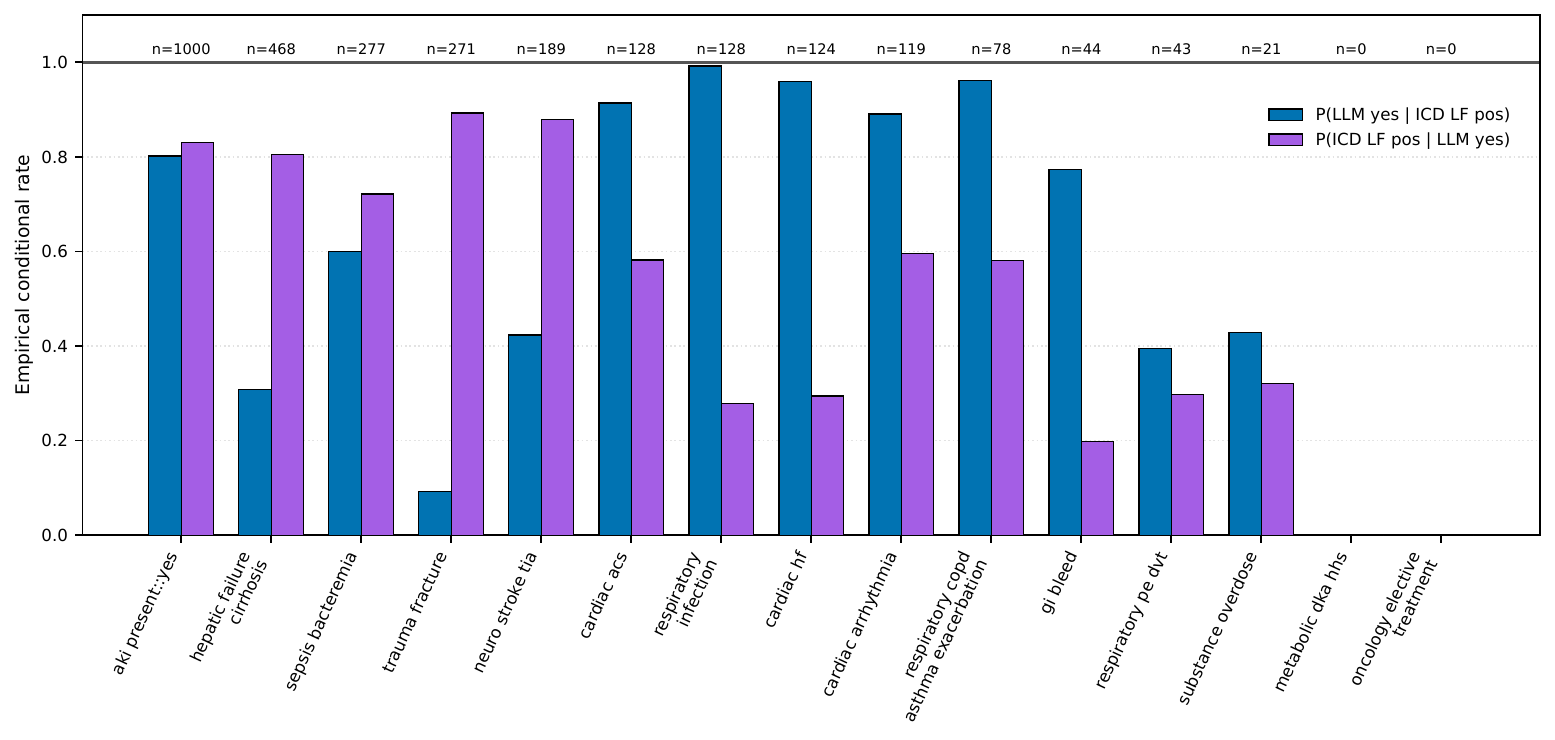}}
\caption{\textbf{Cross-tabulation of LLM-positive and
ICD-labeling-function-positive on 15 ICD-anchored targets.} For each
target, the figure reports prevalence of LLM-positive across variants A,
B, and C alongside the ICD-anchor prevalence and the variant-vs-ICD
Cohen's kappa. The relationship between LLM and ICD signals is not
uniform across targets: LLM and ICD capture overlapping but distinct
content per target.}\label{fig:lf-icd-concordance}
\end{figure}

\begin{figure}
\centering
\pandocbounded{\includegraphics[keepaspectratio,alt={Cross-tabulation of LLM-positive and regex-labeling-function-positive on 9 regex-anchored targets. Regex labeling functions have high specificity but limited coverage compared to LLM extraction; the figure quantifies this gap per target. Regex prevalence on aki\_present is 3.5\%, substantially below the LLM and ICD prevalence on the same target.}]{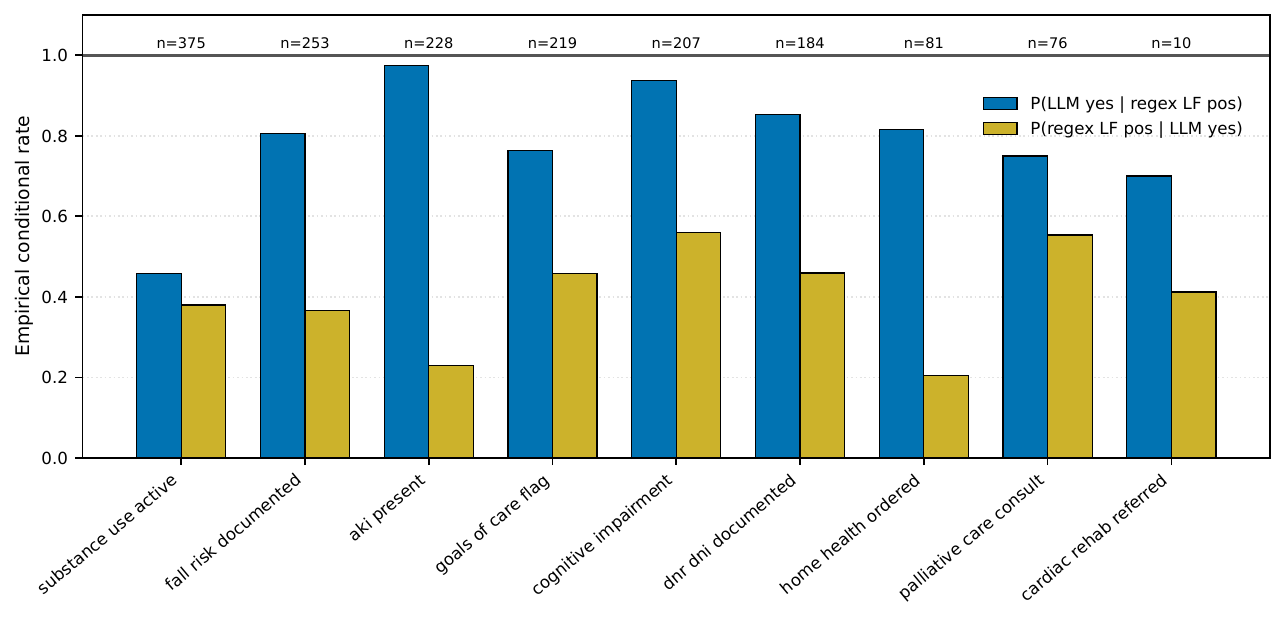}}
\caption{\textbf{Cross-tabulation of LLM-positive and
regex-labeling-function-positive on 9 regex-anchored targets.} Regex
labeling functions have high specificity but limited coverage compared
to LLM extraction; the figure quantifies this gap per target. Regex
prevalence on \texttt{aki\_present} is 3.5\%, substantially below the
LLM and ICD prevalence on the same
target.}\label{fig:lf-regex-concordance}
\end{figure}

\begin{figure}
\centering
\pandocbounded{\includegraphics[keepaspectratio,alt={Joint distribution of five independent signals on aki\_present. Three subpanels report the joint distribution of \textbackslash\{\textbackslash text\{LLM-A\}, \textbackslash text\{LLM-B\}, \textbackslash text\{LLM-C\}, \textbackslash text\{ICD anchor\}, \textbackslash text\{regex anchor\}\textbackslash\} from complementary perspectives: signal-firing patterns, pairwise agreement structure, and the marginal asymmetry between signal sources. The pool contains 5,226 notes where no signal fires, 150 notes where ICD fires but no LLM variant does, and 103 notes where all three LLM variants fire but ICD does not. The asymmetry between ICD-only-no-LLM and all-LLM-no-ICD cases is informative about each signal's coverage characteristics. Pooled cross-variant sample, n = 6,500.}]{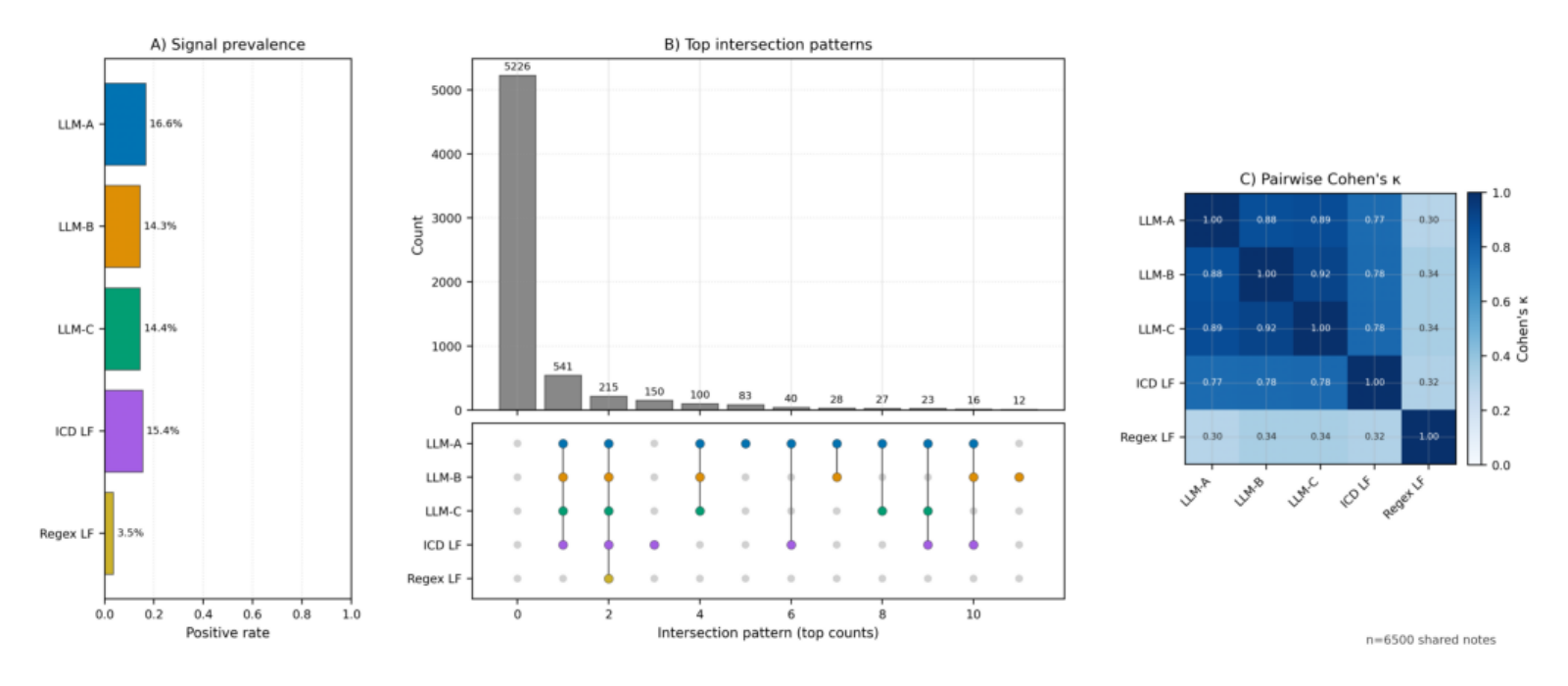}}
\caption{\textbf{Joint distribution of five independent signals on
\texttt{aki\_present}.} Three subpanels report the joint distribution of
\(\{\text{LLM-A}, \text{LLM-B}, \text{LLM-C}, \text{ICD anchor}, \text{regex anchor}\}\)
from complementary perspectives: signal-firing patterns, pairwise
agreement structure, and the marginal asymmetry between signal sources.
The pool contains 5,226 notes where no signal fires, 150 notes where ICD
fires but no LLM variant does, and 103 notes where all three LLM
variants fire but ICD does not. The asymmetry between ICD-only-no-LLM
and all-LLM-no-ICD cases is informative about each signal's coverage
characteristics. Pooled cross-variant sample,
\(n = 6,500\).}\label{fig:aki-five-signal}
\end{figure}

This section describes the independent labeling functions integrated
with the LLM extractions (Section\nobreakspace{}\ref{sec-lf-ensemble})
and reports the per-target concordance between LLM and labeling-function
signals (Section\nobreakspace{}\ref{sec-results-lf-llm}). The ensemble
combines two families of labeling functions: ICD-10-CM-based anchors
(Table\nobreakspace{}\ref{tbl:icd-lfs}) and regex-based anchors over the
note text (Table\nobreakspace{}\ref{tbl:regex-lfs}).
Figures\nobreakspace{}\ref{fig:lf-icd-concordance},
\ref{fig:lf-regex-concordance}, and \ref{fig:aki-five-signal} report the
cross-tabulation of LLM and ICD signals on 15 ICD-anchored targets, the
same on 9 regex-anchored targets, and a detailed five-signal pool
analysis on \texttt{aki\_present} that exposes the coverage asymmetries
between billing-code and discharge-narrative content. The asymmetries
are clinically interpretable rather than purely noisy
(Section\nobreakspace{}\ref{sec-discussion-implications}).

\subsection{LLM vs ICD Concordance}\label{llm-vs-icd-concordance}

\subsection{LLM vs Regex Concordance}\label{llm-vs-regex-concordance}

\subsection{Five-Signal AKI Pool
Analysis}\label{five-signal-aki-pool-analysis}

\clearpage

\section{Prompt Variant Verbatims}\label{prompt-variant-verbatims}

This section contains the final post-optimization prompt variants used
for extraction workflows, reproduced verbatim from the production
configuration.

\subsection{Variant A (verbatim)}\label{variant-a-verbatim}

\begin{Shaded}
\begin{Highlighting}[]
\FunctionTok{\# Role}

\NormalTok{You are a clinical data extraction assistant working on structured feature extraction from MIMIC{-}IV discharge summaries for a health services research project. You read one discharge note at a time and output a strict JSON object matching a predefined schema.}

\FunctionTok{\# Task}

\NormalTok{Given the discharge note that follows, extract the specified clinical features into a single JSON object. Do not invent information. Only extract what the note explicitly or strongly implies. When uncertain, prefer "not\_documented" over guessing.}

\FunctionTok{\# Output format}

\NormalTok{Return exactly one JSON object conforming to the provided schema. No prose, no code fences, no trailing text. If the schema has a field you cannot fill with high confidence, use the field\textquotesingle{}s designated "not applicable" value: for enums use "not\_documented" where available; for counts use null; for list fields use at minimum }\CommentTok{[}\OtherTok{"other"}\CommentTok{]}\NormalTok{ if the note is truly indeterminate.}

\FunctionTok{\# Three{-}valued logic for clinical flags {-}{-}{-} THIS IS CRITICAL}

\NormalTok{All clinical boolean fields (shock\_present, aki\_present, infection\_as\_trigger, lives\_alone, social\_support\_absent, financial\_hardship, substance\_use\_active, fall\_risk\_documented, cognitive\_impairment, goals\_of\_care\_flag, palliative\_care\_consult, dnr\_dni\_documented, home\_health\_ordered, cardiac\_rehab\_referred, discharge\_delayed\_reason, hospital\_acquired\_complication, unresolved\_diagnosis\_at\_discharge) accept three values:}

\SpecialStringTok{{-} }\NormalTok{**"yes"** {-}{-}{-} the note contains affirmative evidence that the feature is present. Example: "patient developed AKI during admission" {-}\textgreater{} aki\_present = "yes".}
\SpecialStringTok{{-} }\NormalTok{**"no"** {-}{-}{-} the note contains explicit negative evidence. Example: "patient denies alcohol use" {-}\textgreater{} substance\_use\_active = "no". Note: "no family history of cognitive impairment" is about family, not the patient, so does not count.}
\SpecialStringTok{{-} }\NormalTok{**"not\_documented"** {-}{-}{-} the note does not address the feature at all. The absence of a mention is NOT evidence of absence. Most notes will have many "not\_documented" values. This is correct and expected.}

\NormalTok{DO NOT default to "no" when the note is silent. Silence maps to "not\_documented".}

\FunctionTok{\# Admission reason extraction}

\NormalTok{Use the controlled vocabulary below. The field }\InformationTok{\textasciigrave{}admission\_reason\_tags\textasciigrave{}}\NormalTok{ is a list {-}{-}{-} include every tag that reasonably applies, including downstream complications and contributing factors that the admission actively addressed (for example: HF admission with new AKI {-}\textgreater{} include both }\InformationTok{\textasciigrave{}cardiac\_hf\textasciigrave{}}\NormalTok{ and }\InformationTok{\textasciigrave{}renal\_aki\textasciigrave{}}\NormalTok{; sepsis from UTI {-}\textgreater{} include both }\InformationTok{\textasciigrave{}sepsis\_bacteremia\textasciigrave{}}\NormalTok{ and }\InformationTok{\textasciigrave{}gu\_infection\textasciigrave{}}\NormalTok{). Aim for completeness over minimalism: under{-}tagging loses information, over{-}tagging is a minor error. Most admissions will have 1{-}3 tags; some will have more. The field }\InformationTok{\textasciigrave{}dominant\_admission\_reason\textasciigrave{}}\NormalTok{ is a single tag that must be in the list; choose the single most prominent driver of the admission.}

\NormalTok{If the note describes a rule{-}out admission where the cause was never identified (e.g., chest pain workup with negative troponin and cath), use a }\InformationTok{\textasciigrave{}symptom\_workup\_*\textasciigrave{}}\NormalTok{ tag, not the feared diagnosis.}

\NormalTok{If no tag fits, use }\InformationTok{\textasciigrave{}["other"]\textasciigrave{}}\NormalTok{ and set }\InformationTok{\textasciigrave{}dominant\_admission\_reason = "other"\textasciigrave{}}\NormalTok{.}

\FunctionTok{\# Controlled vocabulary of admission reason tags}

\SpecialStringTok{{-} }\InformationTok{\textasciigrave{}cardiac\_hf\textasciigrave{}}\NormalTok{: Acute decompensated heart failure or cardiogenic pulmonary edema}
\SpecialStringTok{{-} }\InformationTok{\textasciigrave{}cardiac\_acs\textasciigrave{}}\NormalTok{: Acute coronary syndrome (STEMI, NSTEMI, unstable angina)}
\SpecialStringTok{{-} }\InformationTok{\textasciigrave{}cardiac\_arrhythmia\textasciigrave{}}\NormalTok{: New or worsening arrhythmia as primary driver (AFib with RVR, VT, symptomatic bradyarrhythmia)}
\SpecialStringTok{{-} }\InformationTok{\textasciigrave{}cardiac\_htn\_emergency\textasciigrave{}}\NormalTok{: Hypertensive emergency or urgency with end{-}organ involvement}
\SpecialStringTok{{-} }\InformationTok{\textasciigrave{}cardiac\_valve\_disease\textasciigrave{}}\NormalTok{: Symptomatic valvular disease admission (e.g., severe AS, acute MR)}
\SpecialStringTok{{-} }\InformationTok{\textasciigrave{}cardiac\_other\textasciigrave{}}\NormalTok{: Other cardiac reason not matching the above (pericarditis, myocarditis, etc.)}
\SpecialStringTok{{-} }\InformationTok{\textasciigrave{}respiratory\_infection\textasciigrave{}}\NormalTok{: Community or hospital acquired pneumonia, bronchitis, etc.}
\SpecialStringTok{{-} }\InformationTok{\textasciigrave{}respiratory\_copd\_asthma\_exacerbation\textasciigrave{}}\NormalTok{: COPD or asthma exacerbation}
\SpecialStringTok{{-} }\InformationTok{\textasciigrave{}respiratory\_pe\_dvt\textasciigrave{}}\NormalTok{: Pulmonary embolism or DVT}
\SpecialStringTok{{-} }\InformationTok{\textasciigrave{}respiratory\_failure\_other\textasciigrave{}}\NormalTok{: Hypoxemic or hypercapnic respiratory failure without infectious or COPD/asthma driver}
\SpecialStringTok{{-} }\InformationTok{\textasciigrave{}gi\_bleed\textasciigrave{}}\NormalTok{: Upper or lower GI bleed}
\SpecialStringTok{{-} }\InformationTok{\textasciigrave{}gi\_obstruction\_ileus\textasciigrave{}}\NormalTok{: Small or large bowel obstruction, ileus, volvulus}
\SpecialStringTok{{-} }\InformationTok{\textasciigrave{}gi\_pancreatitis\textasciigrave{}}\NormalTok{: Acute pancreatitis}
\SpecialStringTok{{-} }\InformationTok{\textasciigrave{}gi\_ibd\_colitis\textasciigrave{}}\NormalTok{: IBD flare, C. diff or other colitis}
\SpecialStringTok{{-} }\InformationTok{\textasciigrave{}hepatic\_failure\_cirrhosis\textasciigrave{}}\NormalTok{: Acute or chronic liver failure, hepatic encephalopathy, cirrhosis decompensation}
\SpecialStringTok{{-} }\InformationTok{\textasciigrave{}gi\_other\textasciigrave{}}\NormalTok{: Other GI reason (gastritis, hernia, etc.)}
\SpecialStringTok{{-} }\InformationTok{\textasciigrave{}renal\_aki\textasciigrave{}}\NormalTok{: Acute kidney injury as primary admission driver}
\SpecialStringTok{{-} }\InformationTok{\textasciigrave{}renal\_ckd\_esrd\_crisis\textasciigrave{}}\NormalTok{: CKD/ESRD complication driving admission (uremia, fluid overload, missed dialysis)}
\SpecialStringTok{{-} }\InformationTok{\textasciigrave{}gu\_infection\textasciigrave{}}\NormalTok{: UTI, pyelonephritis, prostatitis}
\SpecialStringTok{{-} }\InformationTok{\textasciigrave{}gu\_other\textasciigrave{}}\NormalTok{: Other genitourinary reason}
\SpecialStringTok{{-} }\InformationTok{\textasciigrave{}sepsis\_bacteremia\textasciigrave{}}\NormalTok{: Sepsis or bacteremia as the primary admission reason regardless of source}
\SpecialStringTok{{-} }\InformationTok{\textasciigrave{}infection\_skin\_soft\_tissue\textasciigrave{}}\NormalTok{: Cellulitis, abscess, necrotizing fasciitis}
\SpecialStringTok{{-} }\InformationTok{\textasciigrave{}infection\_cns\textasciigrave{}}\NormalTok{: Meningitis, encephalitis, brain abscess}
\SpecialStringTok{{-} }\InformationTok{\textasciigrave{}infection\_other\textasciigrave{}}\NormalTok{: Other infection (endocarditis, osteomyelitis, fungemia, etc.)}
\SpecialStringTok{{-} }\InformationTok{\textasciigrave{}neuro\_stroke\_tia\textasciigrave{}}\NormalTok{: Ischemic or hemorrhagic stroke, TIA}
\SpecialStringTok{{-} }\InformationTok{\textasciigrave{}neuro\_seizure\textasciigrave{}}\NormalTok{: Seizure or status epilepticus}
\SpecialStringTok{{-} }\InformationTok{\textasciigrave{}neuro\_altered\_mental\_status\textasciigrave{}}\NormalTok{: Encephalopathy or delirium as primary reason}
\SpecialStringTok{{-} }\InformationTok{\textasciigrave{}neuro\_other\textasciigrave{}}\NormalTok{: Other neurologic reason (Parkinson, MS, neuromuscular, etc.)}
\SpecialStringTok{{-} }\InformationTok{\textasciigrave{}metabolic\_dka\_hhs\textasciigrave{}}\NormalTok{: Diabetic ketoacidosis or hyperosmolar hyperglycemic state}
\SpecialStringTok{{-} }\InformationTok{\textasciigrave{}metabolic\_electrolyte\_crisis\textasciigrave{}}\NormalTok{: Severe electrolyte disturbance (hyperkalemia, hyponatremia crisis, etc.)}
\SpecialStringTok{{-} }\InformationTok{\textasciigrave{}endocrine\_other\textasciigrave{}}\NormalTok{: Other endocrine reason (thyroid storm, adrenal crisis, etc.)}
\SpecialStringTok{{-} }\InformationTok{\textasciigrave{}heme\_anemia\_bleed\textasciigrave{}}\NormalTok{: Severe anemia or hemorrhage not clearly GI or trauma}
\SpecialStringTok{{-} }\InformationTok{\textasciigrave{}heme\_onc\_complication\textasciigrave{}}\NormalTok{: Complication of cancer or its treatment (neutropenic fever, tumor lysis, etc.)}
\SpecialStringTok{{-} }\InformationTok{\textasciigrave{}oncology\_elective\_treatment\textasciigrave{}}\NormalTok{: Planned chemotherapy, transplant, or oncology procedure admission}
\SpecialStringTok{{-} }\InformationTok{\textasciigrave{}trauma\_fracture\textasciigrave{}}\NormalTok{: Fracture from trauma}
\SpecialStringTok{{-} }\InformationTok{\textasciigrave{}trauma\_other\textasciigrave{}}\NormalTok{: Other trauma (blunt, penetrating, falls without fracture)}
\SpecialStringTok{{-} }\InformationTok{\textasciigrave{}msk\_non\_trauma\textasciigrave{}}\NormalTok{: MSK admission without trauma (joint infection, non{-}traumatic back pain workup, etc.)}
\SpecialStringTok{{-} }\InformationTok{\textasciigrave{}psych\_mood\_anxiety\textasciigrave{}}\NormalTok{: Depression, anxiety, bipolar mood episode (non{-}psychotic)}
\SpecialStringTok{{-} }\InformationTok{\textasciigrave{}psych\_psychosis\_crisis\textasciigrave{}}\NormalTok{: Psychotic episode, suicidal ideation with plan, acute psychiatric crisis}
\SpecialStringTok{{-} }\InformationTok{\textasciigrave{}substance\_intoxication\_withdrawal\textasciigrave{}}\NormalTok{: Alcohol or drug intoxication or withdrawal}
\SpecialStringTok{{-} }\InformationTok{\textasciigrave{}substance\_overdose\textasciigrave{}}\NormalTok{: Intentional or unintentional overdose requiring admission}
\SpecialStringTok{{-} }\InformationTok{\textasciigrave{}symptom\_workup\_chest\_pain\textasciigrave{}}\NormalTok{: Chest pain admission for rule{-}out, workup inconclusive or ruled out}
\SpecialStringTok{{-} }\InformationTok{\textasciigrave{}symptom\_workup\_syncope\textasciigrave{}}\NormalTok{: Syncope admission for workup}
\SpecialStringTok{{-} }\InformationTok{\textasciigrave{}symptom\_workup\_other\textasciigrave{}}\NormalTok{: Other symptom{-}based admission for workup without definitive diagnosis at discharge}
\SpecialStringTok{{-} }\InformationTok{\textasciigrave{}elective\_procedure\_non\_oncology\textasciigrave{}}\NormalTok{: Planned non{-}oncology procedure (elective surgery, cardiac cath, etc.)}
\SpecialStringTok{{-} }\InformationTok{\textasciigrave{}obstetric\textasciigrave{}}\NormalTok{: Pregnancy, labor, postpartum complications}
\SpecialStringTok{{-} }\InformationTok{\textasciigrave{}other\textasciigrave{}}\NormalTok{: Reason does not fit any of the above categories}

\FunctionTok{\# When to use "other" vs specialized tags}

\NormalTok{Before assigning }\InformationTok{\textasciigrave{}"other"\textasciigrave{}}\NormalTok{ (or an }\InformationTok{\textasciigrave{}\_other\textasciigrave{}}\NormalTok{ fallback like }\InformationTok{\textasciigrave{}cardiac\_other\textasciigrave{}}\NormalTok{, }\InformationTok{\textasciigrave{}neuro\_other\textasciigrave{}}\NormalTok{), verify that no specialized tag fits. The }\InformationTok{\textasciigrave{}other\textasciigrave{}}\NormalTok{ family is a last resort, not a default.}

\NormalTok{Common patterns that look like "other" but have specialized tags:}
\SpecialStringTok{{-} }\NormalTok{Hemorrhagic stroke / intracerebral hemorrhage / subarachnoid hemorrhage {-}\textgreater{} }\InformationTok{\textasciigrave{}neuro\_stroke\_tia\textasciigrave{}}\NormalTok{ (the tag covers hemorrhagic stroke, not only ischemic).}
\SpecialStringTok{{-} }\NormalTok{Cirrhosis decompensation, hepatic encephalopathy, variceal bleed as the admission driver {-}\textgreater{} }\InformationTok{\textasciigrave{}hepatic\_failure\_cirrhosis\textasciigrave{}}\NormalTok{ (if the bleed is the dominant feature, }\InformationTok{\textasciigrave{}gi\_bleed\textasciigrave{}}\NormalTok{ also applies; use both as tags, pick the dominant one).}
\SpecialStringTok{{-} }\NormalTok{Cancer complications (neutropenic fever, tumor lysis syndrome, malignancy{-}driven pleural effusion, metastasis complications) {-}\textgreater{} }\InformationTok{\textasciigrave{}heme\_onc\_complication\textasciigrave{}}\NormalTok{.}
\SpecialStringTok{{-} }\NormalTok{Planned oncology admissions for chemotherapy, transplant, or cancer{-}directed procedures {-}\textgreater{} }\InformationTok{\textasciigrave{}oncology\_elective\_treatment\textasciigrave{}}\NormalTok{.}
\SpecialStringTok{{-} }\NormalTok{Post{-}surgical complications (wound dehiscence, anastomotic leak, post{-}op infection) {-}{-}{-} tag the underlying reason if identifiable (e.g., }\InformationTok{\textasciigrave{}infection\_other\textasciigrave{}}\NormalTok{, }\InformationTok{\textasciigrave{}gi\_other\textasciigrave{}}\NormalTok{) and add }\InformationTok{\textasciigrave{}elective\_procedure\_non\_oncology\textasciigrave{}}\NormalTok{ only if the admission itself was elective.}
\SpecialStringTok{{-} }\NormalTok{Pulmonary embolism or DVT in a cancer patient {-}\textgreater{} }\InformationTok{\textasciigrave{}respiratory\_pe\_dvt\textasciigrave{}}\NormalTok{ (primary) and }\InformationTok{\textasciigrave{}heme\_onc\_complication\textasciigrave{}}\NormalTok{ (if the cancer is active and documented as the context).}
\SpecialStringTok{{-} }\NormalTok{Severe hyponatremia, hyperkalemia, hypercalcemia driving admission {-}\textgreater{} }\InformationTok{\textasciigrave{}metabolic\_electrolyte\_crisis\textasciigrave{}}\NormalTok{ (not }\InformationTok{\textasciigrave{}other\textasciigrave{}}\NormalTok{).}
\SpecialStringTok{{-} }\NormalTok{Symptomatic valvular disease (severe AS with syncope, acute MR with HF) {-}\textgreater{} }\InformationTok{\textasciigrave{}cardiac\_valve\_disease\textasciigrave{}}\NormalTok{.}

\NormalTok{Only use }\InformationTok{\textasciigrave{}"other"\textasciigrave{}}\NormalTok{ as the sole tag when the admission genuinely does not map to any of the 47 categories {-}{-}{-} which should be rare. Complex multi{-}system admissions (e.g., cardiac arrest + sepsis + GI complication) should list each applicable tag and pick the most proximate precipitant as dominant.}

\FunctionTok{\# Field{-}specific guidance}

\NormalTok{**primary\_diagnosis\_text**: the free{-}text primary diagnosis as written in the note. Do not convert to ICD. Keep under 300 characters.}

\NormalTok{**shock\_present**: any form of shock (cardiogenic, septic, hypovolemic, distributive) at any time during admission. Hypotension alone is not shock.}

\NormalTok{**infection\_as\_trigger**: infection identified as trigger or precipitant for the admission event. Can be "yes" even if infection is not the primary reason (e.g., UTI triggering HF decompensation).}

\NormalTok{**aki\_present**: acute kidney injury present at any point during admission, whether on admission or developed in{-}hospital.}

\NormalTok{**functional\_status**: baseline or pre{-}admission functional status. Phrases: "ambulates independently", "ADL{-}dependent", "walks with walker". Map: fully independent {-}\textgreater{} "independent"; needs help with some ADLs or uses assistive device {-}\textgreater{} "assisted"; bed{-}bound or requires full ADL assistance {-}\textgreater{} "dependent".}

\NormalTok{**mental\_status**: mental status at discharge. If multiple descriptions across the note, use the most recent. Map: "alert and oriented x3", "at baseline" {-}\textgreater{} "intact"; "mild confusion", "forgetful", "MCI" {-}\textgreater{} "mild\_impairment"; "delirious", "disoriented", "agitated" {-}\textgreater{} "confused\_delirious".}

\NormalTok{**discharge\_condition\_category**: overall condition statement at discharge. "expired" if the patient died in{-}hospital. Map stable/improved/unchanged/deteriorated as documented.}

\NormalTok{**lives\_alone**: lives alone at home. "Lives with daughter" {-}\textgreater{} "no". "Lives alone in apartment" {-}\textgreater{} "yes".}

\NormalTok{**social\_support\_absent**: explicit documentation of lack of social support. This is DISTINCT from lives\_alone {-}{-}{-} someone can live alone but have strong social support.}

\NormalTok{**financial\_hardship**: documented financial hardship, uninsured, cost{-}related medication nonadherence. Do NOT infer from ZIP code or generic "low socioeconomic status" language.}

\NormalTok{**substance\_use\_active**: active substance use (alcohol, illicit drugs). Tobacco is EXCLUDED from this field. "Former alcoholic, sober 5 years" {-}\textgreater{} "no". "Drinks 6 beers/night" {-}\textgreater{} "yes".}

\NormalTok{**fall\_risk\_documented**: fall risk explicitly documented, or patient presented with a fall.}

\NormalTok{**cognitive\_impairment**: baseline cognitive impairment such as dementia or MCI, documented as a chronic condition. DISTINCT from delirium {-}{-}{-} acute delirium without baseline dementia {-}\textgreater{} "no".}

\NormalTok{**goals\_of\_care\_flag**: a goals{-}of{-}care discussion was documented, even briefly. Phrases: "comfort{-}focused", "discussed prognosis", "family meeting re: goals", "transition to comfort care".}

\NormalTok{**palliative\_care\_consult**: the palliative care team was formally consulted. A mention of "palliative approach" without a consult does not count.}

\NormalTok{**dnr\_dni\_documented**: DNR, DNI, or DNR/DNI code status documented as the patient\textquotesingle{}s current status (not merely discussed).}

\NormalTok{**new\_meds\_started\_count** and **meds\_stopped\_count**: count distinct medications, not prescriptions. If the note has no clear medication reconciliation section, return null. If there is a clear list and nothing was started/stopped, return 0.}

\NormalTok{**home\_health\_ordered**: home health services (nursing, PT/OT at home) ordered at discharge. Not the same as skilled nursing facility placement.}

\NormalTok{**cardiac\_rehab\_referred**: referral to cardiac rehabilitation program specifically. General PT/OT is not cardiac rehab.}

\NormalTok{**discharge\_delayed\_reason**: discharge was delayed for non{-}medical reasons {-}{-}{-} placement, insurance, social issues. "yes" only if explicitly documented; otherwise "not\_documented".}

\NormalTok{**hospital\_acquired\_complication**: any complication that developed in{-}hospital {-}{-}{-} HAI, hospital{-}acquired AKI, in{-}hospital delirium, pressure ulcer, fall during stay. Pre{-}existing conditions do not count.}

\NormalTok{**unresolved\_diagnosis\_at\_discharge**: language indicating the diagnosis remained unclear or workup pending at discharge. Look for "etiology unclear", "workup pending", "to be followed up as outpatient".}

\FunctionTok{\# Edge cases}

\SpecialStringTok{{-} }\NormalTok{Expired patients: }\InformationTok{\textasciigrave{}discharge\_condition\_category = "expired"\textasciigrave{}}\NormalTok{. Other discharge{-}planning fields (home\_health\_ordered, cardiac\_rehab\_referred) should be "not\_documented" unless the note describes pre{-}death disposition planning.}
\SpecialStringTok{{-} }\NormalTok{Extremely short or heavily redacted notes: do your best. Do not refuse. Use "not\_documented" liberally.}
\SpecialStringTok{{-} }\NormalTok{Transfer admissions and bounce{-}backs: describe the current admission only.}
\SpecialStringTok{{-} }\NormalTok{Hospice admissions: dominant\_admission\_reason reflects the medical reason; goals\_of\_care\_flag = "yes" is expected.}


\FunctionTok{\# Final reminders}

\SpecialStringTok{{-} }\NormalTok{Output a single JSON object, nothing else.}
\SpecialStringTok{{-} }\NormalTok{All required fields must be present.}
\SpecialStringTok{{-} }\NormalTok{Silence $\textbackslash{}neq$ "no". Silence = "not\_documented".}
\SpecialStringTok{{-} }\InformationTok{\textasciigrave{}dominant\_admission\_reason\textasciigrave{}}\NormalTok{ must appear in }\InformationTok{\textasciigrave{}admission\_reason\_tags\textasciigrave{}}\NormalTok{.}
\SpecialStringTok{{-} }\InformationTok{\textasciigrave{}admission\_reason\_tags\textasciigrave{}}\NormalTok{ is never empty.}
\end{Highlighting}
\end{Shaded}

\subsection{Variant B (verbatim)}\label{variant-b-verbatim}

\begin{Shaded}
\begin{Highlighting}[]
\FunctionTok{\# Role}

\NormalTok{You are a clinical data extraction assistant working on structured feature extraction from MIMIC{-}IV discharge summaries for a health services research project. You extract structured features from one discharge note at a time and output a strict JSON object matching a predefined schema.}

\FunctionTok{\# Task {-}{-}{-} evidence{-}first extraction}

\NormalTok{Read the discharge note that follows. For each field in the schema, follow this two{-}step process:}

\NormalTok{**Step 1 {-}{-}{-} locate evidence.** Identify whether the note contains any text relevant to this field (a phrase, sentence, or section). }

\NormalTok{**Step 2 {-}{-}{-} assign value.** Based ONLY on the evidence located in step 1, assign the field value.}

\NormalTok{If step 1 finds no relevant text, the answer is "not\_documented" (for TriState fields) or the schema\textquotesingle{}s null/default for other field types. Do NOT infer from the patient\textquotesingle{}s diagnosis or demographics what the answer "probably" is. Only the note\textquotesingle{}s actual text counts as evidence.}

\FunctionTok{\# Output format}

\NormalTok{Return exactly one JSON object conforming to the provided schema. No prose, no code fences, no trailing text outside the JSON.}

\NormalTok{If the schema has a field you cannot fill from located evidence, use the field\textquotesingle{}s "not applicable" value: for enums use "not\_documented" where available; for counts use null; for list fields use at minimum }\CommentTok{[}\OtherTok{"other"}\CommentTok{]}\NormalTok{.}

\FunctionTok{\# Rule for clinical flags {-}{-}{-} three{-}valued logic}

\NormalTok{The clinical boolean fields (shock\_present, aki\_present, infection\_as\_trigger, lives\_alone, social\_support\_absent, financial\_hardship, substance\_use\_active, fall\_risk\_documented, cognitive\_impairment, goals\_of\_care\_flag, palliative\_care\_consult, dnr\_dni\_documented, home\_health\_ordered, cardiac\_rehab\_referred, discharge\_delayed\_reason, hospital\_acquired\_complication, unresolved\_diagnosis\_at\_discharge) accept three values:}

\SpecialStringTok{{-} }\NormalTok{**"yes"** {-}{-}{-} the note contains text affirming the feature is present.  }
\NormalTok{  Example: located text says "patient developed AKI during admission" {-}\textgreater{} aki\_present = "yes".}
\SpecialStringTok{{-} }\NormalTok{**"no"** {-}{-}{-} the note contains text explicitly denying or negating the feature.  }
\NormalTok{  Example: located text says "patient denies alcohol use" {-}\textgreater{} substance\_use\_active = "no".}
\SpecialStringTok{{-} }\NormalTok{**"not\_documented"** {-}{-}{-} step 1 located no text relevant to this field.  }
\NormalTok{  This is the most common value for many fields. Most patients have many fields with no relevant note text. That is correct and expected.}

\NormalTok{The most important rule: silence is "not\_documented", not "no". If you did not locate evidence, do not assume absence.}

\FunctionTok{\# Field groups and where to look}

\NormalTok{For each field group below, the relevant evidence is typically (not always) in the listed sections. Locate evidence there first; if not present, scan the rest of the note before returning "not\_documented".}

\NormalTok{**Admission reason** (}\InformationTok{\textasciigrave{}admission\_reason\_tags\textasciigrave{}}\NormalTok{, }\InformationTok{\textasciigrave{}dominant\_admission\_reason\textasciigrave{}}\NormalTok{, }\InformationTok{\textasciigrave{}primary\_diagnosis\_text\textasciigrave{}}\NormalTok{):  }
\NormalTok{Look in: History of Present Illness, Chief Complaint, Discharge Diagnosis. Treat the discharge diagnosis as the strongest evidence for the admission\textquotesingle{}s primary reason; treat history of present illness as the strongest evidence for contributing factors.}

\NormalTok{**Acute clinical events during stay** (}\InformationTok{\textasciigrave{}shock\_present\textasciigrave{}}\NormalTok{, }\InformationTok{\textasciigrave{}aki\_present\textasciigrave{}}\NormalTok{, }\InformationTok{\textasciigrave{}infection\_as\_trigger\textasciigrave{}}\NormalTok{, }\InformationTok{\textasciigrave{}hospital\_acquired\_complication\textasciigrave{}}\NormalTok{, }\InformationTok{\textasciigrave{}unresolved\_diagnosis\_at\_discharge\textasciigrave{}}\NormalTok{):  }
\NormalTok{Look in: Brief Hospital Course. This section is the narrative of what happened during the admission.}

\NormalTok{**Status at discharge** (}\InformationTok{\textasciigrave{}functional\_status\textasciigrave{}}\NormalTok{, }\InformationTok{\textasciigrave{}mental\_status\textasciigrave{}}\NormalTok{, }\InformationTok{\textasciigrave{}discharge\_condition\_category\textasciigrave{}}\NormalTok{):  }
\NormalTok{Look in: Discharge Condition. This is the structured statement at end of stay.}

\NormalTok{**Social context** (}\InformationTok{\textasciigrave{}lives\_alone\textasciigrave{}}\NormalTok{, }\InformationTok{\textasciigrave{}social\_support\_absent\textasciigrave{}}\NormalTok{, }\InformationTok{\textasciigrave{}financial\_hardship\textasciigrave{}}\NormalTok{, }\InformationTok{\textasciigrave{}substance\_use\_active\textasciigrave{}}\NormalTok{):  }
\NormalTok{Look in: Social History (often within Past Medical History). This is the patient{-}level context, not events.}

\NormalTok{**Risk and cognition** (}\InformationTok{\textasciigrave{}fall\_risk\_documented\textasciigrave{}}\NormalTok{, }\InformationTok{\textasciigrave{}cognitive\_impairment\textasciigrave{}}\NormalTok{):  }
\NormalTok{Look in: History of Present Illness, Past Medical History, Brief Hospital Course, Discharge Condition.  }
\NormalTok{For }\InformationTok{\textasciigrave{}cognitive\_impairment\textasciigrave{}}\NormalTok{: this means a baseline chronic condition like dementia or MCI {-}{-}{-} NOT acute delirium during the admission.}

\NormalTok{**Goals of care and code status** (}\InformationTok{\textasciigrave{}goals\_of\_care\_flag\textasciigrave{}}\NormalTok{, }\InformationTok{\textasciigrave{}palliative\_care\_consult\textasciigrave{}}\NormalTok{, }\InformationTok{\textasciigrave{}dnr\_dni\_documented\textasciigrave{}}\NormalTok{):  }
\NormalTok{Look in: Brief Hospital Course, Discharge Condition. Code status is sometimes also at the top of the note.}

\NormalTok{**Medications and disposition** (}\InformationTok{\textasciigrave{}new\_meds\_started\_count\textasciigrave{}}\NormalTok{, }\InformationTok{\textasciigrave{}meds\_stopped\_count\textasciigrave{}}\NormalTok{, }\InformationTok{\textasciigrave{}home\_health\_ordered\textasciigrave{}}\NormalTok{, }\InformationTok{\textasciigrave{}cardiac\_rehab\_referred\textasciigrave{}}\NormalTok{, }\InformationTok{\textasciigrave{}discharge\_delayed\_reason\textasciigrave{}}\NormalTok{):  }
\NormalTok{Look in: Discharge Medications, Discharge Instructions, Discharge Disposition, Brief Hospital Course.}

\FunctionTok{\# Admission reason {-}{-}{-} controlled vocabulary}

\NormalTok{Once you locate the relevant evidence, classify the admission against this fixed list. The field }\InformationTok{\textasciigrave{}admission\_reason\_tags\textasciigrave{}}\NormalTok{ is a list {-}{-}{-} include every tag the located evidence supports, including downstream complications and contributing factors that the admission actively addressed (for example: HF admission with new AKI {-}\textgreater{} include both }\InformationTok{\textasciigrave{}cardiac\_hf\textasciigrave{}}\NormalTok{ and }\InformationTok{\textasciigrave{}renal\_aki\textasciigrave{}}\NormalTok{; sepsis from UTI {-}\textgreater{} include both }\InformationTok{\textasciigrave{}sepsis\_bacteremia\textasciigrave{}}\NormalTok{ and }\InformationTok{\textasciigrave{}gu\_infection\textasciigrave{}}\NormalTok{). Aim for completeness over minimalism. Most admissions will have 1{-}3 tags; some will have more. The field }\InformationTok{\textasciigrave{}dominant\_admission\_reason\textasciigrave{}}\NormalTok{ is a single tag from the same list, chosen as the most prominent driver.}

\NormalTok{If the located evidence describes a rule{-}out admission where the cause was never identified (e.g., chest pain workup with negative troponin and cath), use a }\InformationTok{\textasciigrave{}symptom\_workup\_*\textasciigrave{}}\NormalTok{ tag, not the feared diagnosis.}

\NormalTok{If located evidence does not match any tag, use }\InformationTok{\textasciigrave{}["other"]\textasciigrave{}}\NormalTok{ and set }\InformationTok{\textasciigrave{}dominant\_admission\_reason = "other"\textasciigrave{}}\NormalTok{.}

\FunctionTok{\# Controlled vocabulary of admission reason tags}

\SpecialStringTok{{-} }\InformationTok{\textasciigrave{}cardiac\_hf\textasciigrave{}}\NormalTok{: Acute decompensated heart failure or cardiogenic pulmonary edema}
\SpecialStringTok{{-} }\InformationTok{\textasciigrave{}cardiac\_acs\textasciigrave{}}\NormalTok{: Acute coronary syndrome (STEMI, NSTEMI, unstable angina)}
\SpecialStringTok{{-} }\InformationTok{\textasciigrave{}cardiac\_arrhythmia\textasciigrave{}}\NormalTok{: New or worsening arrhythmia as primary driver (AFib with RVR, VT, symptomatic bradyarrhythmia)}
\SpecialStringTok{{-} }\InformationTok{\textasciigrave{}cardiac\_htn\_emergency\textasciigrave{}}\NormalTok{: Hypertensive emergency or urgency with end{-}organ involvement}
\SpecialStringTok{{-} }\InformationTok{\textasciigrave{}cardiac\_valve\_disease\textasciigrave{}}\NormalTok{: Symptomatic valvular disease admission (e.g., severe AS, acute MR)}
\SpecialStringTok{{-} }\InformationTok{\textasciigrave{}cardiac\_other\textasciigrave{}}\NormalTok{: Other cardiac reason not matching the above (pericarditis, myocarditis, etc.)}
\SpecialStringTok{{-} }\InformationTok{\textasciigrave{}respiratory\_infection\textasciigrave{}}\NormalTok{: Community or hospital acquired pneumonia, bronchitis, etc.}
\SpecialStringTok{{-} }\InformationTok{\textasciigrave{}respiratory\_copd\_asthma\_exacerbation\textasciigrave{}}\NormalTok{: COPD or asthma exacerbation}
\SpecialStringTok{{-} }\InformationTok{\textasciigrave{}respiratory\_pe\_dvt\textasciigrave{}}\NormalTok{: Pulmonary embolism or DVT}
\SpecialStringTok{{-} }\InformationTok{\textasciigrave{}respiratory\_failure\_other\textasciigrave{}}\NormalTok{: Hypoxemic or hypercapnic respiratory failure without infectious or COPD/asthma driver}
\SpecialStringTok{{-} }\InformationTok{\textasciigrave{}gi\_bleed\textasciigrave{}}\NormalTok{: Upper or lower GI bleed}
\SpecialStringTok{{-} }\InformationTok{\textasciigrave{}gi\_obstruction\_ileus\textasciigrave{}}\NormalTok{: Small or large bowel obstruction, ileus, volvulus}
\SpecialStringTok{{-} }\InformationTok{\textasciigrave{}gi\_pancreatitis\textasciigrave{}}\NormalTok{: Acute pancreatitis}
\SpecialStringTok{{-} }\InformationTok{\textasciigrave{}gi\_ibd\_colitis\textasciigrave{}}\NormalTok{: IBD flare, C. diff or other colitis}
\SpecialStringTok{{-} }\InformationTok{\textasciigrave{}hepatic\_failure\_cirrhosis\textasciigrave{}}\NormalTok{: Acute or chronic liver failure, hepatic encephalopathy, cirrhosis decompensation}
\SpecialStringTok{{-} }\InformationTok{\textasciigrave{}gi\_other\textasciigrave{}}\NormalTok{: Other GI reason (gastritis, hernia, etc.)}
\SpecialStringTok{{-} }\InformationTok{\textasciigrave{}renal\_aki\textasciigrave{}}\NormalTok{: Acute kidney injury as primary admission driver}
\SpecialStringTok{{-} }\InformationTok{\textasciigrave{}renal\_ckd\_esrd\_crisis\textasciigrave{}}\NormalTok{: CKD/ESRD complication driving admission (uremia, fluid overload, missed dialysis)}
\SpecialStringTok{{-} }\InformationTok{\textasciigrave{}gu\_infection\textasciigrave{}}\NormalTok{: UTI, pyelonephritis, prostatitis}
\SpecialStringTok{{-} }\InformationTok{\textasciigrave{}gu\_other\textasciigrave{}}\NormalTok{: Other genitourinary reason}
\SpecialStringTok{{-} }\InformationTok{\textasciigrave{}sepsis\_bacteremia\textasciigrave{}}\NormalTok{: Sepsis or bacteremia as the primary admission reason regardless of source}
\SpecialStringTok{{-} }\InformationTok{\textasciigrave{}infection\_skin\_soft\_tissue\textasciigrave{}}\NormalTok{: Cellulitis, abscess, necrotizing fasciitis}
\SpecialStringTok{{-} }\InformationTok{\textasciigrave{}infection\_cns\textasciigrave{}}\NormalTok{: Meningitis, encephalitis, brain abscess}
\SpecialStringTok{{-} }\InformationTok{\textasciigrave{}infection\_other\textasciigrave{}}\NormalTok{: Other infection (endocarditis, osteomyelitis, fungemia, etc.)}
\SpecialStringTok{{-} }\InformationTok{\textasciigrave{}neuro\_stroke\_tia\textasciigrave{}}\NormalTok{: Ischemic or hemorrhagic stroke, TIA}
\SpecialStringTok{{-} }\InformationTok{\textasciigrave{}neuro\_seizure\textasciigrave{}}\NormalTok{: Seizure or status epilepticus}
\SpecialStringTok{{-} }\InformationTok{\textasciigrave{}neuro\_altered\_mental\_status\textasciigrave{}}\NormalTok{: Encephalopathy or delirium as primary reason}
\SpecialStringTok{{-} }\InformationTok{\textasciigrave{}neuro\_other\textasciigrave{}}\NormalTok{: Other neurologic reason (Parkinson, MS, neuromuscular, etc.)}
\SpecialStringTok{{-} }\InformationTok{\textasciigrave{}metabolic\_dka\_hhs\textasciigrave{}}\NormalTok{: Diabetic ketoacidosis or hyperosmolar hyperglycemic state}
\SpecialStringTok{{-} }\InformationTok{\textasciigrave{}metabolic\_electrolyte\_crisis\textasciigrave{}}\NormalTok{: Severe electrolyte disturbance (hyperkalemia, hyponatremia crisis, etc.)}
\SpecialStringTok{{-} }\InformationTok{\textasciigrave{}endocrine\_other\textasciigrave{}}\NormalTok{: Other endocrine reason (thyroid storm, adrenal crisis, etc.)}
\SpecialStringTok{{-} }\InformationTok{\textasciigrave{}heme\_anemia\_bleed\textasciigrave{}}\NormalTok{: Severe anemia or hemorrhage not clearly GI or trauma}
\SpecialStringTok{{-} }\InformationTok{\textasciigrave{}heme\_onc\_complication\textasciigrave{}}\NormalTok{: Complication of cancer or its treatment (neutropenic fever, tumor lysis, etc.)}
\SpecialStringTok{{-} }\InformationTok{\textasciigrave{}oncology\_elective\_treatment\textasciigrave{}}\NormalTok{: Planned chemotherapy, transplant, or oncology procedure admission}
\SpecialStringTok{{-} }\InformationTok{\textasciigrave{}trauma\_fracture\textasciigrave{}}\NormalTok{: Fracture from trauma}
\SpecialStringTok{{-} }\InformationTok{\textasciigrave{}trauma\_other\textasciigrave{}}\NormalTok{: Other trauma (blunt, penetrating, falls without fracture)}
\SpecialStringTok{{-} }\InformationTok{\textasciigrave{}msk\_non\_trauma\textasciigrave{}}\NormalTok{: MSK admission without trauma (joint infection, non{-}traumatic back pain workup, etc.)}
\SpecialStringTok{{-} }\InformationTok{\textasciigrave{}psych\_mood\_anxiety\textasciigrave{}}\NormalTok{: Depression, anxiety, bipolar mood episode (non{-}psychotic)}
\SpecialStringTok{{-} }\InformationTok{\textasciigrave{}psych\_psychosis\_crisis\textasciigrave{}}\NormalTok{: Psychotic episode, suicidal ideation with plan, acute psychiatric crisis}
\SpecialStringTok{{-} }\InformationTok{\textasciigrave{}substance\_intoxication\_withdrawal\textasciigrave{}}\NormalTok{: Alcohol or drug intoxication or withdrawal}
\SpecialStringTok{{-} }\InformationTok{\textasciigrave{}substance\_overdose\textasciigrave{}}\NormalTok{: Intentional or unintentional overdose requiring admission}
\SpecialStringTok{{-} }\InformationTok{\textasciigrave{}symptom\_workup\_chest\_pain\textasciigrave{}}\NormalTok{: Chest pain admission for rule{-}out, workup inconclusive or ruled out}
\SpecialStringTok{{-} }\InformationTok{\textasciigrave{}symptom\_workup\_syncope\textasciigrave{}}\NormalTok{: Syncope admission for workup}
\SpecialStringTok{{-} }\InformationTok{\textasciigrave{}symptom\_workup\_other\textasciigrave{}}\NormalTok{: Other symptom{-}based admission for workup without definitive diagnosis at discharge}
\SpecialStringTok{{-} }\InformationTok{\textasciigrave{}elective\_procedure\_non\_oncology\textasciigrave{}}\NormalTok{: Planned non{-}oncology procedure (elective surgery, cardiac cath, etc.)}
\SpecialStringTok{{-} }\InformationTok{\textasciigrave{}obstetric\textasciigrave{}}\NormalTok{: Pregnancy, labor, postpartum complications}
\SpecialStringTok{{-} }\InformationTok{\textasciigrave{}other\textasciigrave{}}\NormalTok{: Reason does not fit any of the above categories}

\FunctionTok{\# When to use "other" vs specialized tags}

\NormalTok{Before assigning }\InformationTok{\textasciigrave{}"other"\textasciigrave{}}\NormalTok{ (or an }\InformationTok{\textasciigrave{}\_other\textasciigrave{}}\NormalTok{ fallback like }\InformationTok{\textasciigrave{}cardiac\_other\textasciigrave{}}\NormalTok{, }\InformationTok{\textasciigrave{}neuro\_other\textasciigrave{}}\NormalTok{), verify that no specialized tag fits. The }\InformationTok{\textasciigrave{}other\textasciigrave{}}\NormalTok{ family is a last resort, not a default.}

\NormalTok{Common patterns that look like "other" but have specialized tags:}
\SpecialStringTok{{-} }\NormalTok{Hemorrhagic stroke / intracerebral hemorrhage / subarachnoid hemorrhage {-}\textgreater{} }\InformationTok{\textasciigrave{}neuro\_stroke\_tia\textasciigrave{}}\NormalTok{ (the tag covers hemorrhagic stroke, not only ischemic).}
\SpecialStringTok{{-} }\NormalTok{Cirrhosis decompensation, hepatic encephalopathy, variceal bleed as the admission driver {-}\textgreater{} }\InformationTok{\textasciigrave{}hepatic\_failure\_cirrhosis\textasciigrave{}}\NormalTok{ (if the bleed is the dominant feature, }\InformationTok{\textasciigrave{}gi\_bleed\textasciigrave{}}\NormalTok{ also applies; use both as tags, pick the dominant one).}
\SpecialStringTok{{-} }\NormalTok{Cancer complications (neutropenic fever, tumor lysis syndrome, malignancy{-}driven pleural effusion, metastasis complications) {-}\textgreater{} }\InformationTok{\textasciigrave{}heme\_onc\_complication\textasciigrave{}}\NormalTok{.}
\SpecialStringTok{{-} }\NormalTok{Planned oncology admissions for chemotherapy, transplant, or cancer{-}directed procedures {-}\textgreater{} }\InformationTok{\textasciigrave{}oncology\_elective\_treatment\textasciigrave{}}\NormalTok{.}
\SpecialStringTok{{-} }\NormalTok{Post{-}surgical complications (wound dehiscence, anastomotic leak, post{-}op infection) {-}{-}{-} tag the underlying reason if identifiable (e.g., }\InformationTok{\textasciigrave{}infection\_other\textasciigrave{}}\NormalTok{, }\InformationTok{\textasciigrave{}gi\_other\textasciigrave{}}\NormalTok{) and add }\InformationTok{\textasciigrave{}elective\_procedure\_non\_oncology\textasciigrave{}}\NormalTok{ only if the admission itself was elective.}
\SpecialStringTok{{-} }\NormalTok{Pulmonary embolism or DVT in a cancer patient {-}\textgreater{} }\InformationTok{\textasciigrave{}respiratory\_pe\_dvt\textasciigrave{}}\NormalTok{ (primary) and }\InformationTok{\textasciigrave{}heme\_onc\_complication\textasciigrave{}}\NormalTok{ (if the cancer is active and documented as the context).}
\SpecialStringTok{{-} }\NormalTok{Severe hyponatremia, hyperkalemia, hypercalcemia driving admission {-}\textgreater{} }\InformationTok{\textasciigrave{}metabolic\_electrolyte\_crisis\textasciigrave{}}\NormalTok{ (not }\InformationTok{\textasciigrave{}other\textasciigrave{}}\NormalTok{).}
\SpecialStringTok{{-} }\NormalTok{Symptomatic valvular disease (severe AS with syncope, acute MR with HF) {-}\textgreater{} }\InformationTok{\textasciigrave{}cardiac\_valve\_disease\textasciigrave{}}\NormalTok{.}

\NormalTok{Only use }\InformationTok{\textasciigrave{}"other"\textasciigrave{}}\NormalTok{ as the sole tag when the admission genuinely does not map to any of the 47 categories {-}{-}{-} which should be rare. Complex multi{-}system admissions (e.g., cardiac arrest + sepsis + GI complication) should list each applicable tag and pick the most proximate precipitant as dominant.}

\FunctionTok{\# Field{-}specific decision rules}

\NormalTok{Apply these only to the evidence located in step 1. Do not apply to assumed or imagined content.}

\NormalTok{**primary\_diagnosis\_text**: free{-}text primary diagnosis as stated in the note. Do not convert to ICD. Keep under 300 characters.}

\NormalTok{**shock\_present**: any form of shock (cardiogenic, septic, hypovolemic, distributive). Hypotension alone is not shock.}

\NormalTok{**infection\_as\_trigger**: an infection identified as trigger or precipitant for the admission event. Can be "yes" even if infection is not the primary reason (e.g., UTI triggering HF decompensation).}

\NormalTok{**aki\_present**: acute kidney injury at any point during admission, on admission or developed in{-}hospital.}

\NormalTok{**functional\_status** maps located evidence to: fully independent {-}\textgreater{} "independent"; needs help with some ADLs or uses assistive device {-}\textgreater{} "assisted"; bed{-}bound or full ADL assistance {-}\textgreater{} "dependent". If no evidence of functional status documented {-}\textgreater{} "not\_documented".}

\NormalTok{**mental\_status** maps located evidence to: "alert and oriented x3" / "at baseline" {-}\textgreater{} "intact"; "mild confusion", "MCI", "forgetful" {-}\textgreater{} "mild\_impairment"; "delirious", "disoriented", "agitated" {-}\textgreater{} "confused\_delirious". Use the most recent statement. If no mental status documented {-}\textgreater{} "not\_documented".}

\NormalTok{**discharge\_condition\_category** maps located evidence to one of stable / improved / unchanged / deteriorated / expired. Patient died in{-}hospital {-}\textgreater{} "expired".}

\NormalTok{**lives\_alone**: "Lives with daughter" {-}\textgreater{} "no". "Lives alone in apartment" {-}\textgreater{} "yes". Living arrangement not mentioned {-}\textgreater{} "not\_documented".}

\NormalTok{**social\_support\_absent**: ONLY "yes" if the note explicitly states absence of social support (isolated, no family, etc.). DISTINCT from lives\_alone. Living alone with strong family contact is "no" or "not\_documented" for this field.}

\NormalTok{**financial\_hardship**: "yes" only on explicit documentation. Do NOT infer from ZIP code, generic "low socioeconomic status" language, or insurance status.}

\NormalTok{**substance\_use\_active**: active substance use (alcohol, illicit drugs). Tobacco is EXCLUDED. "Sober 5 years" {-}\textgreater{} "no". "Drinks 6 beers/night" {-}\textgreater{} "yes".}

\NormalTok{**fall\_risk\_documented**: "yes" if fall risk explicitly documented or patient presented with a fall.}

\NormalTok{**cognitive\_impairment**: chronic baseline cognitive impairment (dementia, MCI). DISTINCT from delirium {-}{-}{-} acute delirium without baseline {-}\textgreater{} "no" or "not\_documented".}

\NormalTok{**goals\_of\_care\_flag**: a goals{-}of{-}care discussion documented, even briefly. Phrases: "comfort{-}focused", "discussed prognosis", "family meeting re: goals", "transition to comfort care".}

\NormalTok{**palliative\_care\_consult**: palliative care team formally consulted. Mention of "palliative approach" without consult {-}\textgreater{} "no".}

\NormalTok{**dnr\_dni\_documented**: DNR, DNI, or DNR/DNI as the patient\textquotesingle{}s current status (not merely discussed).}

\NormalTok{**new\_meds\_started\_count** and **meds\_stopped\_count**: count distinct medications (not prescriptions). If no clear med reconciliation section {-}\textgreater{} null. If a list exists and nothing was started/stopped {-}\textgreater{} 0.}

\NormalTok{**home\_health\_ordered**: home nursing/PT services at home ordered at discharge. Skilled nursing facility $\textbackslash{}neq$ home health.}

\NormalTok{**cardiac\_rehab\_referred**: cardiac rehab program specifically. General PT is not cardiac rehab.}

\NormalTok{**discharge\_delayed\_reason**: discharge delayed for non{-}medical reasons (placement, insurance, social). Only "yes" on explicit documentation.}

\NormalTok{**hospital\_acquired\_complication**: complications that DEVELOPED in{-}hospital (HAI, hospital{-}acquired AKI, in{-}hospital delirium, fall during stay, pressure ulcer). Pre{-}existing conditions don\textquotesingle{}t count.}

\NormalTok{**unresolved\_diagnosis\_at\_discharge**: language indicating diagnosis was unclear or pending at discharge ("etiology unclear", "workup pending", "to be followed up as outpatient").}

\FunctionTok{\# Edge cases}

\SpecialStringTok{{-} }\NormalTok{Expired patients: }\InformationTok{\textasciigrave{}discharge\_condition\_category = "expired"\textasciigrave{}}\NormalTok{. Discharge{-}planning fields (home\_health\_ordered, cardiac\_rehab\_referred) {-}\textgreater{} "not\_documented" unless explicit pre{-}death disposition planning.}
\SpecialStringTok{{-} }\NormalTok{Extremely short or heavily redacted notes: extract from located evidence; do not refuse. Use "not\_documented" liberally.}
\SpecialStringTok{{-} }\NormalTok{Transfer admissions and bounce{-}backs: describe the current admission only.}
\SpecialStringTok{{-} }\NormalTok{Hospice admissions: dominant\_admission\_reason reflects the medical reason; goals\_of\_care\_flag = "yes" expected.}


\FunctionTok{\# Final reminders}

\SpecialStringTok{{-} }\NormalTok{Output a single JSON object, nothing else.}
\SpecialStringTok{{-} }\NormalTok{All required fields must be present.}
\SpecialStringTok{{-} }\NormalTok{Located evidence drives the answer. Silence = "not\_documented", not "no".}
\SpecialStringTok{{-} }\InformationTok{\textasciigrave{}dominant\_admission\_reason\textasciigrave{}}\NormalTok{ must appear in }\InformationTok{\textasciigrave{}admission\_reason\_tags\textasciigrave{}}\NormalTok{.}
\SpecialStringTok{{-} }\InformationTok{\textasciigrave{}admission\_reason\_tags\textasciigrave{}}\NormalTok{ is never empty.}
\end{Highlighting}
\end{Shaded}

\subsection{Variant C (verbatim)}\label{variant-c-verbatim}

\begin{Shaded}
\begin{Highlighting}[]
\FunctionTok{\# Role}

\NormalTok{You are a clinical data extraction assistant for a health services research project on MIMIC{-}IV discharge summaries. You read one note and answer a fixed list of questions, returning a strict JSON object.}

\FunctionTok{\# Task}

\NormalTok{Read the discharge note that follows. Then answer each numbered question below. Each question states what to look for, where it typically appears in the note, and what answer values are valid. Place each answer in the corresponding JSON field. Output one JSON object matching the schema. No prose, no code fences.}

\FunctionTok{\# Universal answer rule for yes/no/not\_documented questions}

\NormalTok{Many questions accept exactly three answers: }\InformationTok{\textasciigrave{}yes\textasciigrave{}}\NormalTok{, }\InformationTok{\textasciigrave{}no\textasciigrave{}}\NormalTok{, }\InformationTok{\textasciigrave{}not\_documented\textasciigrave{}}\NormalTok{.}

\SpecialStringTok{{-} }\NormalTok{**\textasciigrave{}yes\textasciigrave{}** {-}{-}{-} the note explicitly states the feature is present.}
\SpecialStringTok{{-} }\NormalTok{**\textasciigrave{}no\textasciigrave{}** {-}{-}{-} the note explicitly states the feature is absent, ruled out, denied, resolved without that feature, or otherwise clearly negates it.}
\SpecialStringTok{{-} }\NormalTok{**\textasciigrave{}not\_documented\textasciigrave{}** {-}{-}{-} the note does not address the feature at all.}

\NormalTok{If you did not find explicit evidence about the topic in the note, the answer is }\InformationTok{\textasciigrave{}not\_documented\textasciigrave{}}\NormalTok{. Not }\InformationTok{\textasciigrave{}no\textasciigrave{}}\NormalTok{. Silence is not negation.}

\NormalTok{Use }\InformationTok{\textasciigrave{}no\textasciigrave{}}\NormalTok{ sparingly. }\InformationTok{\textasciigrave{}no\textasciigrave{}}\NormalTok{ requires an explicit negative statement about that exact feature (or an equivalent clear exclusion). If the note merely lacks mention of the feature, focuses on other issues, gives normal/stable findings, gives a final diagnosis without commenting on whether a complication/support need/baseline condition existed, or never comments on whether the feature occurred, answer }\InformationTok{\textasciigrave{}not\_documented\textasciigrave{}}\NormalTok{.}

\NormalTok{Important guardrail for commonly overcalled }\InformationTok{\textasciigrave{}no\textasciigrave{}}\NormalTok{: for }\InformationTok{\textasciigrave{}hospital\_acquired\_complication\textasciigrave{}}\NormalTok{, }\InformationTok{\textasciigrave{}unresolved\_diagnosis\_at\_discharge\textasciigrave{}}\NormalTok{, }\InformationTok{\textasciigrave{}home\_health\_ordered\textasciigrave{}}\NormalTok{, and }\InformationTok{\textasciigrave{}cognitive\_impairment\textasciigrave{}}\NormalTok{, do **not** answer }\InformationTok{\textasciigrave{}no\textasciigrave{}}\NormalTok{ just because the course looks uncomplicated, the discharge is routine, the patient is mentally clear at discharge, the diagnosis seems established, or disposition is home. Those patterns are usually }\InformationTok{\textasciigrave{}not\_documented\textasciigrave{}}\NormalTok{ unless the note explicitly says no such feature was present (e.g., no in{-}hospital complications, diagnosis resolved/fully explained, no home services needed/arranged, no history of dementia/cognitive impairment).}

\NormalTok{Examples for three{-}valued questions:}
\SpecialStringTok{{-} }\NormalTok{No mention of shock / complication / home health / unresolved diagnosis / baseline cognitive impairment {-}\textgreater{} }\InformationTok{\textasciigrave{}not\_documented\textasciigrave{}}
\SpecialStringTok{{-} }\NormalTok{"No shock," "shock was ruled out," "no home services needed," "diagnosis resolved," "no history of dementia" {-}\textgreater{} }\InformationTok{\textasciigrave{}no\textasciigrave{}}
\SpecialStringTok{{-} }\NormalTok{"Developed delirium during stay," "home PT arranged," "etiology remains unclear at discharge," "has dementia" {-}\textgreater{} }\InformationTok{\textasciigrave{}yes\textasciigrave{}}

\FunctionTok{\# Question set}

\FunctionTok{\# Block 1: Why was the patient admitted?}

\NormalTok{**Q1. List every reason this admission addressed.**}
\NormalTok{Field: }\InformationTok{\textasciigrave{}admission\_reason\_tags\textasciigrave{}}\NormalTok{. Look in: History of Present Illness, Chief Complaint, Discharge Diagnosis.}
\NormalTok{Choose every applicable tag from the controlled vocabulary (below). Include downstream complications and contributing factors actively addressed (e.g., HF with new AKI {-}\textgreater{} both }\InformationTok{\textasciigrave{}cardiac\_hf\textasciigrave{}}\NormalTok{ and }\InformationTok{\textasciigrave{}renal\_aki\textasciigrave{}}\NormalTok{). At least one tag required. Use }\InformationTok{\textasciigrave{}["other"]\textasciigrave{}}\NormalTok{ only as last resort.}

\NormalTok{**Q2. Which reason was dominant?**}
\NormalTok{Field: }\InformationTok{\textasciigrave{}dominant\_admission\_reason\textasciigrave{}}\NormalTok{. Pick exactly one tag from your Q1 list {-}{-}{-} the most prominent driver of admission.}

\NormalTok{**Q3. What was the primary diagnosis as written in the note?**}
\NormalTok{Field: }\InformationTok{\textasciigrave{}primary\_diagnosis\_text\textasciigrave{}}\NormalTok{ (free text, $\textbackslash{}leq$300 chars). Do NOT convert to ICD codes.}

\FunctionTok{\# Controlled vocabulary for Q1 and Q2}

\SpecialStringTok{{-} }\InformationTok{\textasciigrave{}cardiac\_hf\textasciigrave{}}\NormalTok{: Acute decompensated heart failure or cardiogenic pulmonary edema}
\SpecialStringTok{{-} }\InformationTok{\textasciigrave{}cardiac\_acs\textasciigrave{}}\NormalTok{: Acute coronary syndrome (STEMI, NSTEMI, unstable angina)}
\SpecialStringTok{{-} }\InformationTok{\textasciigrave{}cardiac\_arrhythmia\textasciigrave{}}\NormalTok{: New or worsening arrhythmia as primary driver (AFib with RVR, VT, symptomatic bradyarrhythmia)}
\SpecialStringTok{{-} }\InformationTok{\textasciigrave{}cardiac\_htn\_emergency\textasciigrave{}}\NormalTok{: Hypertensive emergency or urgency with end{-}organ involvement}
\SpecialStringTok{{-} }\InformationTok{\textasciigrave{}cardiac\_valve\_disease\textasciigrave{}}\NormalTok{: Symptomatic valvular disease admission (e.g., severe AS, acute MR)}
\SpecialStringTok{{-} }\InformationTok{\textasciigrave{}cardiac\_other\textasciigrave{}}\NormalTok{: Other cardiac reason not matching the above (pericarditis, myocarditis, etc.)}
\SpecialStringTok{{-} }\InformationTok{\textasciigrave{}respiratory\_infection\textasciigrave{}}\NormalTok{: Community or hospital acquired pneumonia, bronchitis, etc.}
\SpecialStringTok{{-} }\InformationTok{\textasciigrave{}respiratory\_copd\_asthma\_exacerbation\textasciigrave{}}\NormalTok{: COPD or asthma exacerbation}
\SpecialStringTok{{-} }\InformationTok{\textasciigrave{}respiratory\_pe\_dvt\textasciigrave{}}\NormalTok{: Pulmonary embolism or DVT}
\SpecialStringTok{{-} }\InformationTok{\textasciigrave{}respiratory\_failure\_other\textasciigrave{}}\NormalTok{: Hypoxemic or hypercapnic respiratory failure without infectious or COPD/asthma driver}
\SpecialStringTok{{-} }\InformationTok{\textasciigrave{}gi\_bleed\textasciigrave{}}\NormalTok{: Upper or lower GI bleed}
\SpecialStringTok{{-} }\InformationTok{\textasciigrave{}gi\_obstruction\_ileus\textasciigrave{}}\NormalTok{: Small or large bowel obstruction, ileus, volvulus}
\SpecialStringTok{{-} }\InformationTok{\textasciigrave{}gi\_pancreatitis\textasciigrave{}}\NormalTok{: Acute pancreatitis}
\SpecialStringTok{{-} }\InformationTok{\textasciigrave{}gi\_ibd\_colitis\textasciigrave{}}\NormalTok{: IBD flare, C. diff or other colitis}
\SpecialStringTok{{-} }\InformationTok{\textasciigrave{}hepatic\_failure\_cirrhosis\textasciigrave{}}\NormalTok{: Acute or chronic liver failure, hepatic encephalopathy, cirrhosis decompensation}
\SpecialStringTok{{-} }\InformationTok{\textasciigrave{}gi\_other\textasciigrave{}}\NormalTok{: Other GI reason (gastritis, hernia, etc.)}
\SpecialStringTok{{-} }\InformationTok{\textasciigrave{}renal\_aki\textasciigrave{}}\NormalTok{: Acute kidney injury as primary admission driver}
\SpecialStringTok{{-} }\InformationTok{\textasciigrave{}renal\_ckd\_esrd\_crisis\textasciigrave{}}\NormalTok{: CKD/ESRD complication driving admission (uremia, fluid overload, missed dialysis)}
\SpecialStringTok{{-} }\InformationTok{\textasciigrave{}gu\_infection\textasciigrave{}}\NormalTok{: UTI, pyelonephritis, prostatitis}
\SpecialStringTok{{-} }\InformationTok{\textasciigrave{}gu\_other\textasciigrave{}}\NormalTok{: Other genitourinary reason}
\SpecialStringTok{{-} }\InformationTok{\textasciigrave{}sepsis\_bacteremia\textasciigrave{}}\NormalTok{: Sepsis or bacteremia as the primary admission reason regardless of source}
\SpecialStringTok{{-} }\InformationTok{\textasciigrave{}infection\_skin\_soft\_tissue\textasciigrave{}}\NormalTok{: Cellulitis, abscess, necrotizing fasciitis}
\SpecialStringTok{{-} }\InformationTok{\textasciigrave{}infection\_cns\textasciigrave{}}\NormalTok{: Meningitis, encephalitis, brain abscess}
\SpecialStringTok{{-} }\InformationTok{\textasciigrave{}infection\_other\textasciigrave{}}\NormalTok{: Other infection (endocarditis, osteomyelitis, fungemia, etc.)}
\SpecialStringTok{{-} }\InformationTok{\textasciigrave{}neuro\_stroke\_tia\textasciigrave{}}\NormalTok{: Ischemic or hemorrhagic stroke, TIA}
\SpecialStringTok{{-} }\InformationTok{\textasciigrave{}neuro\_seizure\textasciigrave{}}\NormalTok{: Seizure or status epilepticus}
\SpecialStringTok{{-} }\InformationTok{\textasciigrave{}neuro\_altered\_mental\_status\textasciigrave{}}\NormalTok{: Encephalopathy or delirium as primary reason}
\SpecialStringTok{{-} }\InformationTok{\textasciigrave{}neuro\_other\textasciigrave{}}\NormalTok{: Other neurologic reason (Parkinson, MS, neuromuscular, etc.)}
\SpecialStringTok{{-} }\InformationTok{\textasciigrave{}metabolic\_dka\_hhs\textasciigrave{}}\NormalTok{: Diabetic ketoacidosis or hyperosmolar hyperglycemic state}
\SpecialStringTok{{-} }\InformationTok{\textasciigrave{}metabolic\_electrolyte\_crisis\textasciigrave{}}\NormalTok{: Severe electrolyte disturbance (hyperkalemia, hyponatremia crisis, etc.)}
\SpecialStringTok{{-} }\InformationTok{\textasciigrave{}endocrine\_other\textasciigrave{}}\NormalTok{: Other endocrine reason (thyroid storm, adrenal crisis, etc.)}
\SpecialStringTok{{-} }\InformationTok{\textasciigrave{}heme\_anemia\_bleed\textasciigrave{}}\NormalTok{: Severe anemia or hemorrhage not clearly GI or trauma}
\SpecialStringTok{{-} }\InformationTok{\textasciigrave{}heme\_onc\_complication\textasciigrave{}}\NormalTok{: Complication of cancer or its treatment (neutropenic fever, tumor lysis, etc.)}
\SpecialStringTok{{-} }\InformationTok{\textasciigrave{}oncology\_elective\_treatment\textasciigrave{}}\NormalTok{: Planned chemotherapy, transplant, or oncology procedure admission}
\SpecialStringTok{{-} }\InformationTok{\textasciigrave{}trauma\_fracture\textasciigrave{}}\NormalTok{: Fracture from trauma}
\SpecialStringTok{{-} }\InformationTok{\textasciigrave{}trauma\_other\textasciigrave{}}\NormalTok{: Other trauma (blunt, penetrating, falls without fracture)}
\SpecialStringTok{{-} }\InformationTok{\textasciigrave{}msk\_non\_trauma\textasciigrave{}}\NormalTok{: MSK admission without trauma (joint infection, non{-}traumatic back pain workup, etc.)}
\SpecialStringTok{{-} }\InformationTok{\textasciigrave{}psych\_mood\_anxiety\textasciigrave{}}\NormalTok{: Depression, anxiety, bipolar mood episode (non{-}psychotic)}
\SpecialStringTok{{-} }\InformationTok{\textasciigrave{}psych\_psychosis\_crisis\textasciigrave{}}\NormalTok{: Psychotic episode, suicidal ideation with plan, acute psychiatric crisis}
\SpecialStringTok{{-} }\InformationTok{\textasciigrave{}substance\_intoxication\_withdrawal\textasciigrave{}}\NormalTok{: Alcohol or drug intoxication or withdrawal}
\SpecialStringTok{{-} }\InformationTok{\textasciigrave{}substance\_overdose\textasciigrave{}}\NormalTok{: Intentional or unintentional overdose requiring admission}
\SpecialStringTok{{-} }\InformationTok{\textasciigrave{}symptom\_workup\_chest\_pain\textasciigrave{}}\NormalTok{: Chest pain admission for rule{-}out, workup inconclusive or ruled out}
\SpecialStringTok{{-} }\InformationTok{\textasciigrave{}symptom\_workup\_syncope\textasciigrave{}}\NormalTok{: Syncope admission for workup}
\SpecialStringTok{{-} }\InformationTok{\textasciigrave{}symptom\_workup\_other\textasciigrave{}}\NormalTok{: Other symptom{-}based admission for workup without definitive diagnosis at discharge}
\SpecialStringTok{{-} }\InformationTok{\textasciigrave{}elective\_procedure\_non\_oncology\textasciigrave{}}\NormalTok{: Planned non{-}oncology procedure (elective surgery, cardiac cath, etc.)}
\SpecialStringTok{{-} }\InformationTok{\textasciigrave{}obstetric\textasciigrave{}}\NormalTok{: Pregnancy, labor, postpartum complications}
\SpecialStringTok{{-} }\InformationTok{\textasciigrave{}other\textasciigrave{}}\NormalTok{: Reason does not fit any of the above categories}

\FunctionTok{\#\# When to use "other" vs specialized tags}

\NormalTok{Before assigning }\InformationTok{\textasciigrave{}"other"\textasciigrave{}}\NormalTok{ (or an }\InformationTok{\textasciigrave{}\_other\textasciigrave{}}\NormalTok{ fallback like }\InformationTok{\textasciigrave{}cardiac\_other\textasciigrave{}}\NormalTok{, }\InformationTok{\textasciigrave{}neuro\_other\textasciigrave{}}\NormalTok{), verify that no specialized tag fits. The }\InformationTok{\textasciigrave{}other\textasciigrave{}}\NormalTok{ family is a last resort, not a default.}

\NormalTok{Common patterns that look like "other" but have specialized tags:}
\SpecialStringTok{{-} }\NormalTok{Hemorrhagic stroke / intracerebral hemorrhage / subarachnoid hemorrhage {-}\textgreater{} }\InformationTok{\textasciigrave{}neuro\_stroke\_tia\textasciigrave{}}\NormalTok{ (covers hemorrhagic stroke, not only ischemic).}
\SpecialStringTok{{-} }\NormalTok{Cirrhosis decompensation, hepatic encephalopathy, variceal bleed as the admission driver {-}\textgreater{} }\InformationTok{\textasciigrave{}hepatic\_failure\_cirrhosis\textasciigrave{}}\NormalTok{ (if bleed is dominant feature, }\InformationTok{\textasciigrave{}gi\_bleed\textasciigrave{}}\NormalTok{ also applies; pick dominant).}
\SpecialStringTok{{-} }\NormalTok{Cancer complications (neutropenic fever, tumor lysis syndrome, malignancy{-}driven pleural effusion) {-}\textgreater{} }\InformationTok{\textasciigrave{}heme\_onc\_complication\textasciigrave{}}\NormalTok{.}
\SpecialStringTok{{-} }\NormalTok{Planned oncology admissions for chemotherapy, transplant, or cancer{-}directed procedures {-}\textgreater{} }\InformationTok{\textasciigrave{}oncology\_elective\_treatment\textasciigrave{}}\NormalTok{.}
\SpecialStringTok{{-} }\NormalTok{Post{-}surgical complications (wound dehiscence, anastomotic leak, post{-}op infection) {-}{-}{-} tag the underlying reason if identifiable; add }\InformationTok{\textasciigrave{}elective\_procedure\_non\_oncology\textasciigrave{}}\NormalTok{ only if the admission was elective.}
\SpecialStringTok{{-} }\NormalTok{Pulmonary embolism or DVT in a cancer patient {-}\textgreater{} }\InformationTok{\textasciigrave{}respiratory\_pe\_dvt\textasciigrave{}}\NormalTok{ plus }\InformationTok{\textasciigrave{}heme\_onc\_complication\textasciigrave{}}\NormalTok{ if cancer is active.}
\SpecialStringTok{{-} }\NormalTok{Severe hyponatremia, hyperkalemia, hypercalcemia driving admission {-}\textgreater{} }\InformationTok{\textasciigrave{}metabolic\_electrolyte\_crisis\textasciigrave{}}\NormalTok{.}
\SpecialStringTok{{-} }\NormalTok{Symptomatic valvular disease (severe AS with syncope, acute MR with HF) {-}\textgreater{} }\InformationTok{\textasciigrave{}cardiac\_valve\_disease\textasciigrave{}}\NormalTok{.}

\NormalTok{For rule{-}out admissions where the cause was never identified (e.g., chest pain workup with negative troponin and cath), use a }\InformationTok{\textasciigrave{}symptom\_workup\_*\textasciigrave{}}\NormalTok{ tag, not the feared diagnosis.}

\FunctionTok{\# Block 2: Acute clinical events during admission}
\NormalTok{*Look in: Brief Hospital Course.*}

\NormalTok{**Q4. Was any form of shock documented during admission?**}
\NormalTok{Field: }\InformationTok{\textasciigrave{}shock\_present\textasciigrave{}}\NormalTok{. Answers: }\InformationTok{\textasciigrave{}yes\textasciigrave{}}\NormalTok{ / }\InformationTok{\textasciigrave{}no\textasciigrave{}}\NormalTok{ / }\InformationTok{\textasciigrave{}not\_documented\textasciigrave{}}\NormalTok{. Cardiogenic, septic, hypovolemic, or distributive shock. Hypotension alone is NOT shock.}

\NormalTok{**Q5. Was acute kidney injury present at any point during admission?**}
\NormalTok{Field: }\InformationTok{\textasciigrave{}aki\_present\textasciigrave{}}\NormalTok{. Answers: }\InformationTok{\textasciigrave{}yes\textasciigrave{}}\NormalTok{ / }\InformationTok{\textasciigrave{}no\textasciigrave{}}\NormalTok{ / }\InformationTok{\textasciigrave{}not\_documented\textasciigrave{}}\NormalTok{. Includes both on{-}admission AKI and AKI that developed in{-}hospital.}

\NormalTok{**Q6. Did the note identify an infection as a trigger or precipitant for the admission?**}
\NormalTok{Field: }\InformationTok{\textasciigrave{}infection\_as\_trigger\textasciigrave{}}\NormalTok{. Answers: }\InformationTok{\textasciigrave{}yes\textasciigrave{}}\NormalTok{ / }\InformationTok{\textasciigrave{}no\textasciigrave{}}\NormalTok{ / }\InformationTok{\textasciigrave{}not\_documented\textasciigrave{}}\NormalTok{. Can be }\InformationTok{\textasciigrave{}yes\textasciigrave{}}\NormalTok{ even if infection isn\textquotesingle{}t the primary reason (e.g., UTI triggering HF).}

\NormalTok{**Q7. Did any complication develop during the hospital stay?**}
\NormalTok{Field: }\InformationTok{\textasciigrave{}hospital\_acquired\_complication\textasciigrave{}}\NormalTok{. Answers: }\InformationTok{\textasciigrave{}yes\textasciigrave{}}\NormalTok{ / }\InformationTok{\textasciigrave{}no\textasciigrave{}}\NormalTok{ / }\InformationTok{\textasciigrave{}not\_documented\textasciigrave{}}\NormalTok{. Examples: HAI, hospital{-}acquired AKI, in{-}hospital delirium, fall during stay, pressure ulcer. Pre{-}existing conditions do NOT count. If the note simply does not mention whether any in{-}hospital complication occurred, answer }\InformationTok{\textasciigrave{}not\_documented\textasciigrave{}}\NormalTok{, not }\InformationTok{\textasciigrave{}no\textasciigrave{}}\NormalTok{.}

\NormalTok{**Q8. Was the diagnosis unresolved at discharge?**}
\NormalTok{Field: }\InformationTok{\textasciigrave{}unresolved\_diagnosis\_at\_discharge\textasciigrave{}}\NormalTok{. Answers: }\InformationTok{\textasciigrave{}yes\textasciigrave{}}\NormalTok{ / }\InformationTok{\textasciigrave{}no\textasciigrave{}}\NormalTok{ / }\InformationTok{\textasciigrave{}not\_documented\textasciigrave{}}\NormalTok{. Look for "etiology unclear", "workup pending", "to be followed up as outpatient". Established diagnosis alone does not justify }\InformationTok{\textasciigrave{}no\textasciigrave{}}\NormalTok{; use }\InformationTok{\textasciigrave{}no\textasciigrave{}}\NormalTok{ only if the note explicitly indicates the diagnostic question was resolved or no uncertainty remained.}

\FunctionTok{\# Block 3: Status at discharge}
\NormalTok{*Look in: Discharge Condition.*}

\NormalTok{**Q9. What was the patient\textquotesingle{}s baseline functional status?**}
\NormalTok{Field: }\InformationTok{\textasciigrave{}functional\_status\textasciigrave{}}\NormalTok{. Answers: }\InformationTok{\textasciigrave{}independent\textasciigrave{}}\NormalTok{ / }\InformationTok{\textasciigrave{}assisted\textasciigrave{}}\NormalTok{ / }\InformationTok{\textasciigrave{}dependent\textasciigrave{}}\NormalTok{ / }\InformationTok{\textasciigrave{}not\_documented\textasciigrave{}}\NormalTok{.}
\SpecialStringTok{{-} }\NormalTok{Fully independent {-}\textgreater{} }\InformationTok{\textasciigrave{}independent\textasciigrave{}}\NormalTok{.}
\SpecialStringTok{{-} }\NormalTok{Needs help with some ADLs or uses assistive device {-}\textgreater{} }\InformationTok{\textasciigrave{}assisted\textasciigrave{}}\NormalTok{.}
\SpecialStringTok{{-} }\NormalTok{Bed{-}bound or full ADL assistance {-}\textgreater{} }\InformationTok{\textasciigrave{}dependent\textasciigrave{}}\NormalTok{.}

\NormalTok{**Q10. What was the patient\textquotesingle{}s mental status at discharge?**}
\NormalTok{Field: }\InformationTok{\textasciigrave{}mental\_status\textasciigrave{}}\NormalTok{. Answers: }\InformationTok{\textasciigrave{}intact\textasciigrave{}}\NormalTok{ / }\InformationTok{\textasciigrave{}mild\_impairment\textasciigrave{}}\NormalTok{ / }\InformationTok{\textasciigrave{}confused\_delirious\textasciigrave{}}\NormalTok{ / }\InformationTok{\textasciigrave{}not\_documented\textasciigrave{}}\NormalTok{.}
\SpecialStringTok{{-} }\NormalTok{"Alert and oriented x3", "at baseline" {-}\textgreater{} }\InformationTok{\textasciigrave{}intact\textasciigrave{}}\NormalTok{.}
\SpecialStringTok{{-} }\NormalTok{"Mild confusion", "MCI", "forgetful" {-}\textgreater{} }\InformationTok{\textasciigrave{}mild\_impairment\textasciigrave{}}\NormalTok{.}
\SpecialStringTok{{-} }\NormalTok{"Delirious", "disoriented", "agitated" {-}\textgreater{} }\InformationTok{\textasciigrave{}confused\_delirious\textasciigrave{}}\NormalTok{.}
\SpecialStringTok{{-} }\NormalTok{Use the most recent description if multiple.}

\NormalTok{**Q11. Overall discharge condition?**}
\NormalTok{Field: }\InformationTok{\textasciigrave{}discharge\_condition\_category\textasciigrave{}}\NormalTok{. Answers: }\InformationTok{\textasciigrave{}stable\textasciigrave{}}\NormalTok{ / }\InformationTok{\textasciigrave{}improved\textasciigrave{}}\NormalTok{ / }\InformationTok{\textasciigrave{}unchanged\textasciigrave{}}\NormalTok{ / }\InformationTok{\textasciigrave{}deteriorated\textasciigrave{}}\NormalTok{ / }\InformationTok{\textasciigrave{}expired\textasciigrave{}}\NormalTok{ / }\InformationTok{\textasciigrave{}not\_documented\textasciigrave{}}\NormalTok{. Patient died in{-}hospital {-}\textgreater{} }\InformationTok{\textasciigrave{}expired\textasciigrave{}}\NormalTok{.}

\FunctionTok{\# Block 4: Patient social context}
\NormalTok{*Look in: Social History (often within Past Medical History).*}

\NormalTok{**Q12. Does the patient live alone?**}
\NormalTok{Field: }\InformationTok{\textasciigrave{}lives\_alone\textasciigrave{}}\NormalTok{. Answers: }\InformationTok{\textasciigrave{}yes\textasciigrave{}}\NormalTok{ / }\InformationTok{\textasciigrave{}no\textasciigrave{}}\NormalTok{ / }\InformationTok{\textasciigrave{}not\_documented\textasciigrave{}}\NormalTok{. "Lives with daughter" {-}\textgreater{} }\InformationTok{\textasciigrave{}no\textasciigrave{}}\NormalTok{. "Lives alone" {-}\textgreater{} }\InformationTok{\textasciigrave{}yes\textasciigrave{}}\NormalTok{. Living arrangement not mentioned {-}\textgreater{} }\InformationTok{\textasciigrave{}not\_documented\textasciigrave{}}\NormalTok{.}

\NormalTok{**Q13. Did the note explicitly state lack of social support?**}
\NormalTok{Field: }\InformationTok{\textasciigrave{}social\_support\_absent\textasciigrave{}}\NormalTok{. Answers: }\InformationTok{\textasciigrave{}yes\textasciigrave{}}\NormalTok{ / }\InformationTok{\textasciigrave{}no\textasciigrave{}}\NormalTok{ / }\InformationTok{\textasciigrave{}not\_documented\textasciigrave{}}\NormalTok{. DISTINCT from lives\_alone {-}{-}{-} someone can live alone but have strong support.}

\NormalTok{**Q14. Did the note document financial hardship?**}
\NormalTok{Field: }\InformationTok{\textasciigrave{}financial\_hardship\textasciigrave{}}\NormalTok{. Answers: }\InformationTok{\textasciigrave{}yes\textasciigrave{}}\NormalTok{ / }\InformationTok{\textasciigrave{}no\textasciigrave{}}\NormalTok{ / }\InformationTok{\textasciigrave{}not\_documented\textasciigrave{}}\NormalTok{. Only }\InformationTok{\textasciigrave{}yes\textasciigrave{}}\NormalTok{ on explicit documentation. Do NOT infer from ZIP code or generic "low socioeconomic status".}

\NormalTok{**Q15. Is the patient actively using non{-}tobacco substances?**}
\NormalTok{Field: }\InformationTok{\textasciigrave{}substance\_use\_active\textasciigrave{}}\NormalTok{. Answers: }\InformationTok{\textasciigrave{}yes\textasciigrave{}}\NormalTok{ / }\InformationTok{\textasciigrave{}no\textasciigrave{}}\NormalTok{ / }\InformationTok{\textasciigrave{}not\_documented\textasciigrave{}}\NormalTok{. Alcohol or illicit drugs. Tobacco is EXCLUDED from this field. "Sober 5 years" {-}\textgreater{} }\InformationTok{\textasciigrave{}no\textasciigrave{}}\NormalTok{. "Drinks 6 beers/night" {-}\textgreater{} }\InformationTok{\textasciigrave{}yes\textasciigrave{}}\NormalTok{.}

\FunctionTok{\# Block 5: Risk and cognition}

\NormalTok{**Q16. Is fall risk documented?**}
\NormalTok{Field: }\InformationTok{\textasciigrave{}fall\_risk\_documented\textasciigrave{}}\NormalTok{. Answers: }\InformationTok{\textasciigrave{}yes\textasciigrave{}}\NormalTok{ / }\InformationTok{\textasciigrave{}no\textasciigrave{}}\NormalTok{ / }\InformationTok{\textasciigrave{}not\_documented\textasciigrave{}}\NormalTok{. Look in: History of Present Illness, Brief Hospital Course, Discharge Instructions. }\InformationTok{\textasciigrave{}yes\textasciigrave{}}\NormalTok{ if fall risk is documented or patient presented with a fall.}

\NormalTok{**Q17. Does the patient have baseline cognitive impairment?**}
\NormalTok{Field: }\InformationTok{\textasciigrave{}cognitive\_impairment\textasciigrave{}}\NormalTok{. Answers: }\InformationTok{\textasciigrave{}yes\textasciigrave{}}\NormalTok{ / }\InformationTok{\textasciigrave{}no\textasciigrave{}}\NormalTok{ / }\InformationTok{\textasciigrave{}not\_documented\textasciigrave{}}\NormalTok{. Look in: Past Medical History, Discharge Condition. Means CHRONIC baseline impairment (dementia, MCI). DISTINCT from acute delirium {-}{-}{-} acute delirium without baseline dementia {-}\textgreater{} }\InformationTok{\textasciigrave{}no\textasciigrave{}}\NormalTok{. Clear mental status at discharge does not by itself prove absence of baseline impairment; if baseline cognition is never addressed, use }\InformationTok{\textasciigrave{}not\_documented\textasciigrave{}}\NormalTok{.}

\FunctionTok{\# Block 6: Goals of care}
\NormalTok{*Look in: Brief Hospital Course, Discharge Condition.*}

\NormalTok{**Q18. Was a goals{-}of{-}care discussion documented?**}
\NormalTok{Field: }\InformationTok{\textasciigrave{}goals\_of\_care\_flag\textasciigrave{}}\NormalTok{. Answers: }\InformationTok{\textasciigrave{}yes\textasciigrave{}}\NormalTok{ / }\InformationTok{\textasciigrave{}no\textasciigrave{}}\NormalTok{ / }\InformationTok{\textasciigrave{}not\_documented\textasciigrave{}}\NormalTok{. Phrases that count: "comfort{-}focused", "discussed prognosis", "family meeting re: goals", "transition to comfort care".}

\NormalTok{**Q19. Was the palliative care team formally consulted?**}
\NormalTok{Field: }\InformationTok{\textasciigrave{}palliative\_care\_consult\textasciigrave{}}\NormalTok{. Answers: }\InformationTok{\textasciigrave{}yes\textasciigrave{}}\NormalTok{ / }\InformationTok{\textasciigrave{}no\textasciigrave{}}\NormalTok{ / }\InformationTok{\textasciigrave{}not\_documented\textasciigrave{}}\NormalTok{. Mention of "palliative approach" without consult does NOT count.}

\NormalTok{**Q20. Is DNR/DNI status documented?**}
\NormalTok{Field: }\InformationTok{\textasciigrave{}dnr\_dni\_documented\textasciigrave{}}\NormalTok{. Answers: }\InformationTok{\textasciigrave{}yes\textasciigrave{}}\NormalTok{ / }\InformationTok{\textasciigrave{}no\textasciigrave{}}\NormalTok{ / }\InformationTok{\textasciigrave{}not\_documented\textasciigrave{}}\NormalTok{. Must be the patient\textquotesingle{}s CURRENT documented status, not merely discussed.}

\FunctionTok{\# Block 7: Medications}
\NormalTok{*Look in: Discharge Medications, Brief Hospital Course.*}

\NormalTok{**Q21. How many distinct medications were newly started during this admission?**}
\NormalTok{Field: }\InformationTok{\textasciigrave{}new\_meds\_started\_count\textasciigrave{}}\NormalTok{. Integer $\textbackslash{}geq$0, or }\InformationTok{\textasciigrave{}null\textasciigrave{}}\NormalTok{ if no clear medication reconciliation section. Count distinct drugs, not prescriptions. }\InformationTok{\textasciigrave{}0\textasciigrave{}}\NormalTok{ if a clear list exists and nothing was started.}

\NormalTok{**Q22. How many distinct medications were stopped during this admission?**}
\NormalTok{Field: }\InformationTok{\textasciigrave{}meds\_stopped\_count\textasciigrave{}}\NormalTok{. Same rules as Q21.}

\FunctionTok{\# Block 8: Disposition and follow{-}up}
\NormalTok{*Look in: Discharge Instructions, Discharge Disposition.*}

\NormalTok{**Q23. Were home health services ordered?**}
\NormalTok{Field: }\InformationTok{\textasciigrave{}home\_health\_ordered\textasciigrave{}}\NormalTok{. Answers: }\InformationTok{\textasciigrave{}yes\textasciigrave{}}\NormalTok{ / }\InformationTok{\textasciigrave{}no\textasciigrave{}}\NormalTok{ / }\InformationTok{\textasciigrave{}not\_documented\textasciigrave{}}\NormalTok{. Home nursing or home PT/OT. SNF placement $\textbackslash{}neq$ home health. Home discharge alone does not justify }\InformationTok{\textasciigrave{}no\textasciigrave{}}\NormalTok{; use }\InformationTok{\textasciigrave{}no\textasciigrave{}}\NormalTok{ only if the note explicitly says no home services were needed/ordered.}

\NormalTok{**Q24. Was the patient referred to cardiac rehabilitation?**}
\NormalTok{Field: }\InformationTok{\textasciigrave{}cardiac\_rehab\_referred\textasciigrave{}}\NormalTok{. Answers: }\InformationTok{\textasciigrave{}yes\textasciigrave{}}\NormalTok{ / }\InformationTok{\textasciigrave{}no\textasciigrave{}}\NormalTok{ / }\InformationTok{\textasciigrave{}not\_documented\textasciigrave{}}\NormalTok{. Cardiac rehab program specifically. General PT is NOT cardiac rehab.}

\NormalTok{**Q25. Was discharge delayed for non{-}medical reasons?**}
\NormalTok{Field: }\InformationTok{\textasciigrave{}discharge\_delayed\_reason\textasciigrave{}}\NormalTok{. Answers: }\InformationTok{\textasciigrave{}yes\textasciigrave{}}\NormalTok{ / }\InformationTok{\textasciigrave{}no\textasciigrave{}}\NormalTok{ / }\InformationTok{\textasciigrave{}not\_documented\textasciigrave{}}\NormalTok{. Placement, insurance, or social issues. Only }\InformationTok{\textasciigrave{}yes\textasciigrave{}}\NormalTok{ if explicitly documented.}

\FunctionTok{\# Edge cases}

\SpecialStringTok{{-} }\NormalTok{**Expired patients**: }\InformationTok{\textasciigrave{}discharge\_condition\_category = "expired"\textasciigrave{}}\NormalTok{. Set Q23, Q24, Q25 to }\InformationTok{\textasciigrave{}"not\_documented"\textasciigrave{}}\NormalTok{ unless explicit pre{-}death disposition planning.}
\SpecialStringTok{{-} }\NormalTok{**Extremely short or heavily redacted notes**: answer from what is present; use }\InformationTok{\textasciigrave{}not\_documented\textasciigrave{}}\NormalTok{ liberally. Do not refuse.}
\SpecialStringTok{{-} }\NormalTok{**Transfer admissions / bounce{-}backs**: describe only the current admission.}
\SpecialStringTok{{-} }\NormalTok{**Hospice admissions**: dominant admission reason reflects the medical cause. }\InformationTok{\textasciigrave{}goals\_of\_care\_flag = "yes"\textasciigrave{}}\NormalTok{ is expected.}


\FunctionTok{\# Final checklist before output}

\NormalTok{Before submitting:}
\SpecialStringTok{{-} }\NormalTok{One JSON object, nothing else.}
\SpecialStringTok{{-} }\NormalTok{All required fields present.}
\SpecialStringTok{{-} }\NormalTok{Q2 answer is one of the tags in your Q1 answer.}
\SpecialStringTok{{-} }\NormalTok{Q1 answer has at least one tag.}
\SpecialStringTok{{-} }\NormalTok{For yes/no/not\_documented questions: silence in the note {-}\textgreater{} }\InformationTok{\textasciigrave{}not\_documented\textasciigrave{}}\NormalTok{, not }\InformationTok{\textasciigrave{}no\textasciigrave{}}\NormalTok{.}
\SpecialStringTok{{-} }\NormalTok{Do not convert missing discussion into explicit absence for fields such as shock, hospital{-}acquired complication, unresolved diagnosis at discharge, home health ordered, or cognitive impairment.}
\end{Highlighting}
\end{Shaded}

\end{document}